\PassOptionsToPackage{table}{xcolor}

\documentclass[Afour,sageh,times]{sagej}

\usepackage[ruled, vlined, titlenotnumbered, linesnumbered]{algorithm2e}
\usepackage{multirow}
\usepackage{tabularx}
\usepackage{longtable}
\usepackage{array}
\usepackage{ragged2e}
\usepackage{colortbl}
\usepackage{makecell}
\usepackage{multicol}
\usepackage{float}
\usepackage{tablefootnote}
\usepackage{pifont}
\usepackage{wrapfig}
\usepackage{subcaption}
\usepackage{soul}
\usepackage[normalem]{ulem}
\usepackage{tikz}
\usetikzlibrary{positioning, fit, arrows.meta, shapes.geometric, calc}

\usepackage{bm}

\newcolumntype{Y}{>{\raggedright\arraybackslash}X}

\usepackage[colorlinks,bookmarksopen,bookmarksnumbered,citecolor=green,urlcolor=blue,linkcolor=blue]{hyperref}
\def\volumeyear{2026}
\def\journalname{The International Journal of Robotics Research}
\def\volumenumber{0}
\def\issuenumber{0}

\begin{document}

\runninghead{Zheng \textit{et~al.}}

\title{From Video to Control: A Survey of Learning Manipulation Interfaces from Temporal Visual Data}

\author{Linfang Zheng\affilnum{1}, Zikai Ouyang\affilnum{2,3}, Chen Wang\affilnum{1,2},  Jia Pan\affilnum{1} and Wei Zhang\affilnum{2,4}}

\affiliation{\affilnum{1}The University of Hong Kong, Hong Kong, China\\
\affilnum{2}Southern University of Science and Technology, Shenzhen, China\\
\affilnum{3}Peng Cheng Laboratory, Shenzhen, China\\
\affilnum{4}LimX Dynamics}

\corrauth{Wei Zhang, School of Automation and Intelligent Manufacturing (AiM), and Guangdong Provincial Key Laboratory of Fully Actuated System Control Theory and Technology, Southern University of Science and Technology, Shenzhen 518055, China.}

\email{zhangw3@sustech.edu.cn}

\begin{abstract}
Video is a scalable observation of physical dynamics: it captures how objects move, how contact unfolds, and how scenes evolve under interaction---all without requiring robot action labels.
Yet translating this temporal structure into reliable robotic control remains an open challenge, because video lacks action supervision and differs from robot experience in embodiment, viewpoint, and physical constraints.
This survey reviews methods that exploit non-action-annotated temporal video to learn control interfaces for robotic manipulation.
We introduce an \emph{interface-centric taxonomy} organized by where the video-to-control interface is constructed and what control properties it enables, identifying three families: direct video--action policies, which keep the interface implicit; latent-action methods, which route temporal structure through a compact learned intermediate; and explicit visual interfaces, which predict interpretable targets for downstream control.
For each family, we analyze control-integration properties---how the loop is closed, what can be verified before execution, and where failures enter.
A cross-family synthesis reveals that the most pressing open challenges center on the \emph{robotics integration layer}---the mechanisms that connect video-derived predictions to dependable robot behavior---and we outline research directions toward closing this gap.
\end{abstract}


\maketitle

\section{Introduction}
\label{sec:intro}

Robotic manipulation is a cornerstone capability for embodied intelligence, underpinning applications ranging from household assistance~\citep{household_robots_fiorini2000cleaning, household_robots_soni2024advancing} and logistics~\citep{Robotics_logistics_echelmeyer2008robotics} to industrial automation~\citep{robot_industrial_automation_faheem2024ai} and human--robot collaboration~\citep{human_robot_collection_review_baratta2023human}.
Recent advances in large-scale learning-based policies have demonstrated impressive progress toward generalist manipulation, with systems trained on hundreds of thousands of robot demonstrations exhibiting robustness across tasks, objects, and environments~\citep{RT1_Brohan_2022, RT2_Brohan_2023, pi0_Black_2024, SpatialVLA_Qu_2025}.
Despite these successes, such approaches remain fundamentally constrained by data: collecting robot trajectories with synchronized actions, proprioception, and rewards is expensive, time-consuming, and difficult to scale, even with coordinated multi-robot efforts such as Open X-Embodiment (OXE)~\citep{OXE_ICRA_2024}.

\begin{figure}[t]
    \centering
    \includegraphics[width=1\linewidth, trim= 150 120 150 120, clip]{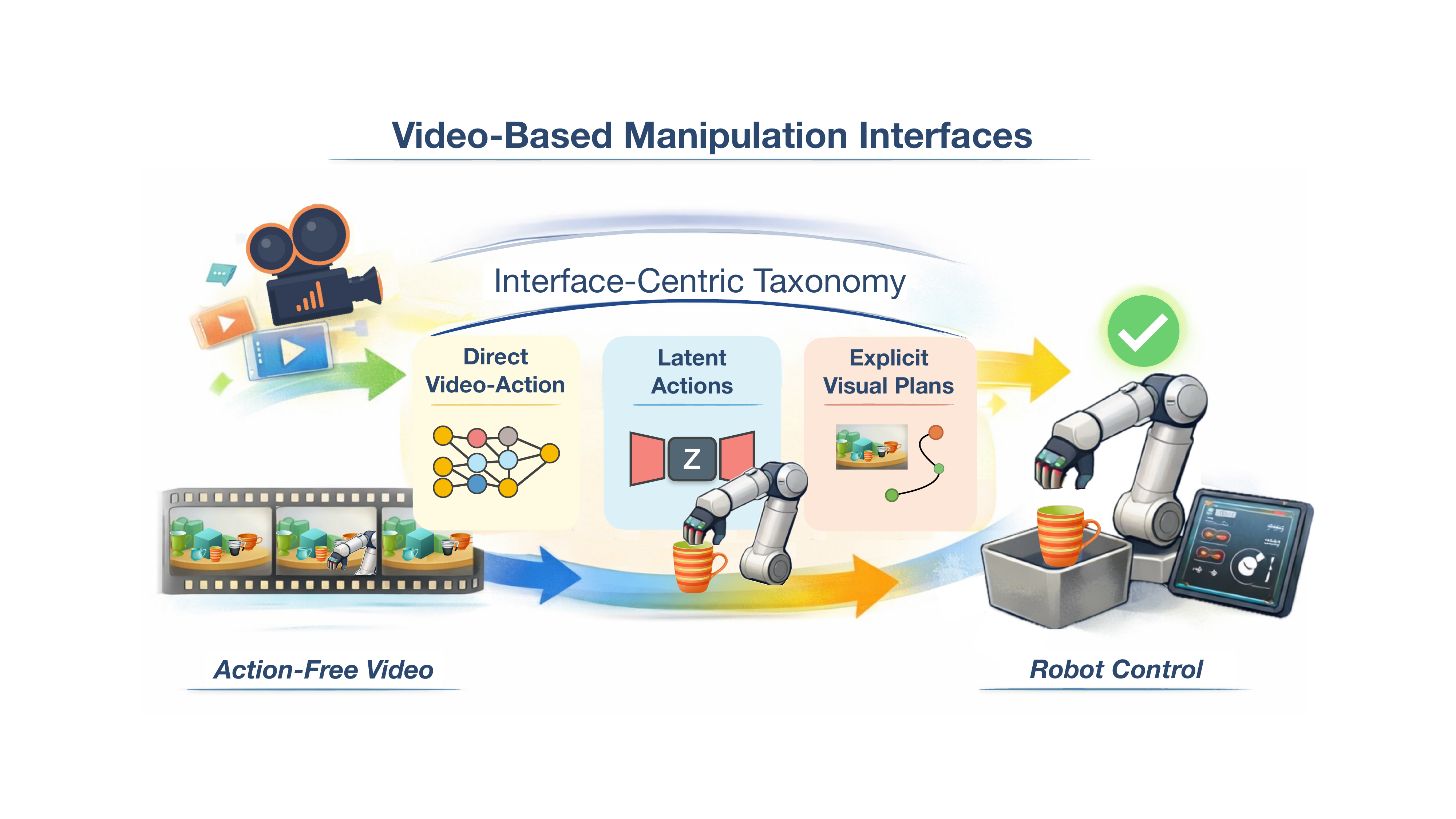}
    \vspace{-5mm}
    \caption{\textbf{Video-based manipulation interfaces.}
    This survey organizes the literature by how video-derived temporal structure is connected to robot control through three recurring interface families: direct video--action policies, latent-action intermediates, and explicit visual interfaces (e.g., subgoal images, trajectories, or poses).
    These families differ in how explicitly that structure is exposed to the robot's control loop, providing the conceptual basis for the taxonomy and comparative analyses that follow.}
    \label{fig:teaser}
    \vspace{-8mm}
\end{figure}

\begin{figure*}
    \centering
    \includegraphics[width=0.8\linewidth, trim= 280 280 300 280, clip]{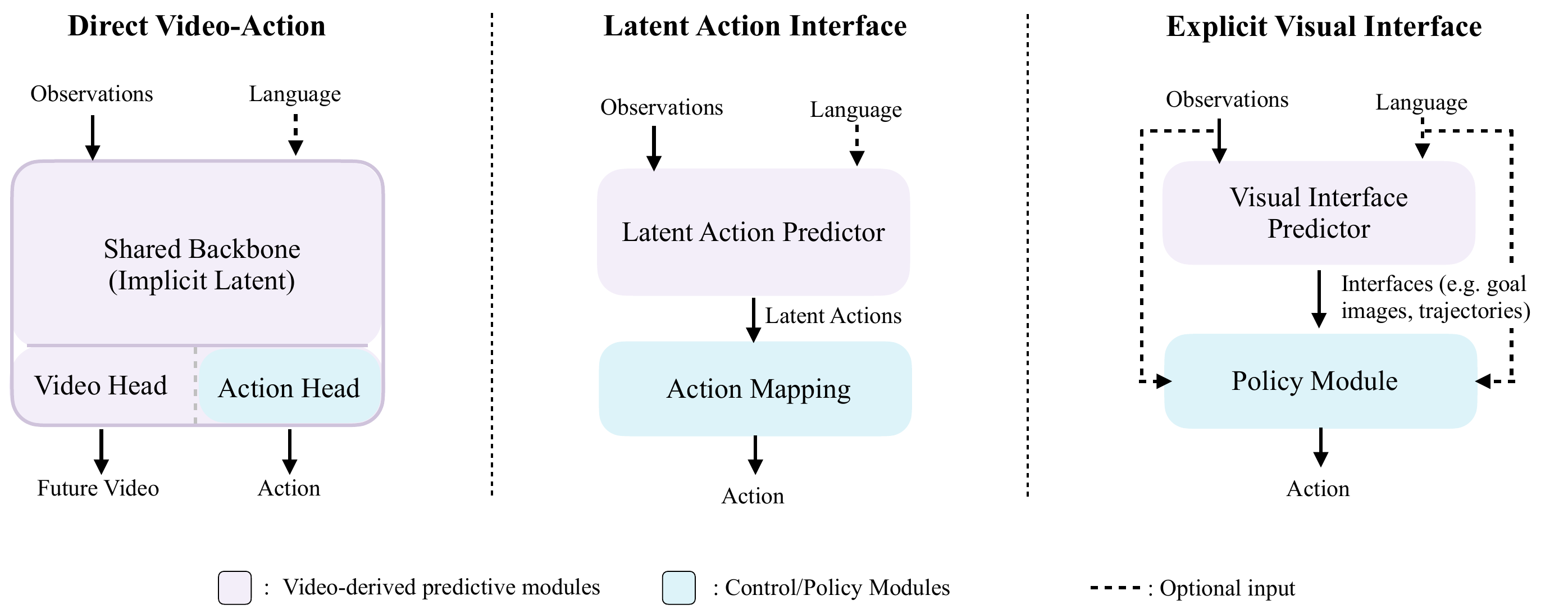}
    \caption{\textbf{Three families of video-based manipulation interfaces.}
    Each column shows how video-derived temporal structure is connected to robot actions through predictive modules and control modules.
    Direct video--action policies (left) keep this interface implicit within a shared backbone; latent-action methods (center) route it through a learned action-like intermediate; and explicit visual interfaces (right) predict interpretable targets that a downstream controller or policy module tracks.
    From left to right, the interface becomes more explicit and the stages of prediction and control become more cleanly separated.}
    \label{fig:three_category_structure}
\end{figure*}
In contrast, the web and personal devices now host vast quantities of video depicting physical interaction.
Egocentric datasets (e.g., Ego4D~\citep{Ego4d_CVPR_2022}, EPIC-Kitchens~\citep{EPIC-KITCHENS-100_IJCV_2022}) capture rich hand--object interactions in diverse environments, while online platforms contain countless videos of robots performing manipulation tasks.
These videos encode critical information about objects, affordances, contact, and temporal structure---\emph{how the world changes over time as a result of interaction}.
However, they lack synchronized robot action labels and are often recorded from different embodiments, viewpoints, and sensing modalities than a target robot.
This creates a central tension:
\emph{the most abundant source of experience is action-free video, while the most directly useful supervision---robot actions---is scarce.}
Bridging this gap is not merely a data problem; the temporal structure extracted from video must ultimately close a control loop, respect physical constraints, and function within the embodiment limits of a specific robot.

Rather than organizing the literature by model class or generative technique, we adopt an \emph{interface-centric} view: the central question is where and how video-derived temporal structure enters the robot's control stack, and what control properties that interface affords.
Accordingly, we analyze not only how methods are trained, but also how they close the control loop, what can be inspected or verified before execution, and where physical inconsistencies or grounding failures may arise.
The resulting families align with familiar robotics patterns, including end-to-end visuomotor control and hierarchical planner--controller decompositions.
We focus on approaches that use \emph{non-action-annotated video} as a primary supervision signal; we do not cover visual pretraining based solely on static images or contrastive objectives without temporal prediction.

\paragraph{The unifying question of this survey is:}
\begin{quote}
\emph{How can large-scale, non-action-annotated video---viewed as a scalable observation of world dynamics---be used to learn control interfaces that support reliable robotic manipulation?}
\end{quote}

While specific techniques vary, most methods share a common structure: video shapes an intermediate representation---explicit or implicit---that captures how scenes evolve over time, and a smaller amount of robot-specific data grounds this representation to executable actions.
What differs fundamentally is where this interface is constructed, how explicit or interpretable it is, and how it is integrated into the control loop---choices that determine what can be executed, inspected, and transferred across embodiments.

\paragraph{Three families of video-based manipulation methods.}
We organize the literature around this interface-centric perspective into three families (Figure~\ref{fig:three_category_structure}).

\textbf{Direct video--action policies} keep the video-to-control interface implicit: temporal prediction shapes internal representations from which actions are decoded directly, without exposing an intermediate target for inspection or downstream planning.
This simplifies deployment, but makes the video-to-action link opaque and harder to verify, interpret, or transfer across embodiments.

\textbf{Latent-action methods} introduce compact action-like intermediates learned from observed transitions, which may be retained as planning/control interfaces or used as action-centric supervision before native-action fine-tuning.
This can reduce the need for robot-action supervision and make temporal abstraction more modular, but the learned codes may entangle controllable and exogenous change, and grounding or transfer may be brittle under distribution shift.

\textbf{Explicit visual interfaces} produce structured, interpretable targets---subgoal images, video plans, trajectories, or pose sequences---that a downstream controller explicitly tracks.
This improves transparency and can ease cross-embodiment transfer, but introduces perception and grounding stages whose errors can dominate execution.

Across all three families, action-free video provides the temporal prior, while robot data or interaction grounds the learned interface to executable behavior; what changes is how explicitly that interface is exposed and what kind of control-loop integration it requires.
In recent terminology, tightly coupled direct video--action policies are often described as \emph{world--action models}, since they jointly model future world evolution and the actions associated with it; our taxonomy places these models within a broader interface-centric view that also includes latent-action and explicit visual-interface approaches.

Crucially, methods built on similar video backbones can fall into different control regimes, and a visually accurate prediction need not be executable on a given robot.
An architecture- or generative-technique-centric grouping can therefore obscure the control distinctions this survey targets; our interface-centric taxonomy is designed to make those distinctions visible.

\paragraph{This survey makes three 
contributions:}
\begin{itemize}
    \item An \emph{interface-centric taxonomy} that organizes the literature by where and how video-derived temporal structure enters the robot's control stack, structured along two design axes (interface explicitness and distance from robot actions).
    \item A \emph{per-family control-integration analysis} that compares how interface design shapes control-loop closure, what can be verified before execution, and where failures enter---properties that cut across architecture choices.
    \item A \emph{cross-family synthesis} yielding a \emph{robotics integration layer} thesis: the most pressing unresolved gaps lie in grounding, loop closure, physical feasibility, and verification, with future research consolidated into four diagnosis-driven themes.
\end{itemize}
\section{Related Surveys and Selection Criteria}
\label{sec:related}

This section positions our survey relative to existing reviews and clarifies the scope of papers we include.
Broadly, prior related surveys on robotic manipulation fall into two overlapping streams:
(i) surveys centered on \emph{foundation models and vision--language--action (VLA) systems}, where action-labeled robot trajectories and/or robot interaction are the dominant grounding signal; and
(ii) surveys on \emph{visual perception, dynamics modeling, and world models}, which are typically grounded in robot interaction data or classical sensing--control pipelines.
In contrast, our survey centers \emph{non-action-annotated temporal video} as a scalable supervision source and organizes methods by \emph{where the interface between video-derived dynamics and robot actions is constructed}.

\subsection{Vision--Language--Action Models and Foundation Models for Manipulation}

Recent surveys increasingly frame robotic manipulation through foundation models and vision--language--action (VLA) systems.
These reviews typically organize the literature around model families, system modules, training pipelines, or action-conditioned policies.
This landscape includes broad treatments of foundation models for embodied AI~\citep{xu2024surveyroboticsfoundationmodels}, surveys of foundation-model families and their roles across the manipulation pipeline~\citep{li2024foundation_manipulation_survey}, and VLA surveys organized by architectural paradigms, training pipelines, datasets, and evaluation protocols~\citep{shao2025large_VLA_survey,din2025vision_VLA_survey,motoda2025_VLA_survey}. More focused reviews examine reinforcement-learning-based post-training~\citep{deng2025_RL_VLA_manipulation_survey} or action-tokenization and action-representation choices~\citep{zhong2025surveyvisionlanguageactionmodelsaction}.

Within these foundation-model surveys, the closest point of contact with our survey is their treatment of video.
For example, Li et al.~\citep{li2024foundation_manipulation_survey} discuss how internet-scale video can provide manipulation-relevant signals for robotic foundation models, including trajectories, poses, affordances, and task semantics.
Their discussion is primarily organized around an information-and-utilization question: what useful signals can be extracted from video, often with foundation or off-the-shelf models, and how these signals can support broader robotic foundation-model pipelines.

By contrast, our survey asks a control-interface question: how temporally learned video structure is coupled to robot control, where the video-to-control interface is constructed, how it is grounded into executable actions, and how this coupling affects loop closure, pre-execution inspection, embodiment transfer, and failure modes.
This leads to a distinct taxonomy of direct video--action policies, latent-action methods, and explicit visual interfaces, organized by the mechanism that connects video-derived dynamics to control rather than by foundation-model type or manipulation-system module.

\subsection{Video-Based and Dynamics-Centered Surveys for Robotic Manipulation}
Earlier surveys on vision-based manipulation are largely organized around sensing and control pipelines.
These include classical treatments of visual servoing and closed-loop visual--motor control~\citep{kragic2002survey_visual_servoing_for_manipulation}, active vision and next-best-view planning under uncertainty~\citep{chen2011active_perception_survey}, affordance-centric perspectives on manipulation~\citep{yamanobe2017_affordance_manipulation_survey}, and surveys of 3D perception pipelines spanning sensing, pose estimation, grasping, and motion planning~\citep{cong2021_3D_perception_manipulation_survey}.
Broader reviews catalog practical components and challenges in vision-based manipulation systems~\citep{shahria2022_vision_manipulation_survey,wang2025robot_embodied_visual_perception_survey}.

More recent surveys emphasize predictive modeling and world-model viewpoints that integrate perception, prediction, and control~\citep{zhang2025stepworldmodelssurvey}, as well as design trade-offs in learned dynamics models (e.g., state representations, model architectures, and planning integration)~\citep{survey_learning_based_dynamics_2025}.
A separate recent survey reviews imitation learning for contact-rich manipulation, with an emphasis on force- and contact-aware policy learning~\citep{Tsuji2026_immitation_learning_survey}.
However, these works are typically grounded in robot interaction data or action-labeled demonstrations and do not center large-scale action-free video as a primary supervision source.

Closest to our scope are learning-from-video (LfV) surveys, which explicitly discuss drawing on large-scale internet or in-the-wild video despite the absence of action labels and substantial domain shift~\citep{mccarthy2025_video_manipulation_survye,eze2025learningwatchingreviewvideobased,Feng_human_video_survey_TechRxiv_2026}.
These surveys provide valuable coverage of datasets, challenges, and high-level strategies for incorporating video into robot learning, with Feng et al.\ focusing specifically on human-centric video as a data channel.
Our survey complements them with a \emph{structural taxonomy} (developed in \S\ref{sec:taxonomy}) that disentangles \emph{how} video-derived temporal structure is connected to robot control, surfacing recurring control-integration trade-offs that are easy to miss when treating ``learning from video'' as a single category.

\subsection{Paper Selection Criteria}
\label{subsec:selection}

This survey targets video-driven manipulation methods that learn \emph{temporally predictive} representations or interfaces from large corpora of \emph{non-action-annotated} video and subsequently \emph{ground} those representations to robot control.
Our goal is to synthesize a fast-growing literature whose defining technical ingredient is the use of temporal continuity—such as forecasting, planning, or tracking through time—as the primary supervision signal for learning an interface between video dynamics and robot actions.

\paragraph{Definition (non-action-annotated video).}
We use \emph{non-action-annotated video} to mean video data for which aligned robot actions are \emph{unavailable or unused} in the learning objective.
This covers both human/in-the-wild video and robot video where actions exist in the dataset but are not provided to the model during the video-learning phase.
Non-action-annotated video may still include auxiliary annotations such as captions, masks, bounding boxes, depth, keypoints, or point tracks.

\paragraph{Inclusion criteria}
We include a method if it satisfies all of the following:
(i) \textbf{Temporal video supervision:} it exploits \emph{temporal continuity} in video as a core training signal (e.g., video prediction, goal or subgoal frame prediction, point or flow trajectory forecasting, or learning temporally predictive latent variables from frame transitions), rather than relying solely on static, single-frame objectives;
(ii) \textbf{Interface learned from non-action video:} the key video-to-control interface is learned or pretrained using non-action-annotated video, potentially at large scale (from human videos, robot videos without actions, or mixed sources);
(iii) \textbf{Grounding to manipulation:} the learned interface is connected to robotic manipulation through robot data and/or is used within a policy, planner, or control loop.
Robot data may include action labels and is allowed to be limited in quantity (e.g., for imitation learning, action decoding, inverse dynamics, reinforcement learning fine-tuning, or embodiment calibration).

\paragraph{Treatment of preprint literature.}
The video-based manipulation literature is evolving rapidly, with substantial arXiv-only activity alongside peer-reviewed work.
We apply the inclusion criteria above (criteria (i)--(iii)) uniformly across published and preprint papers, while requiring preprints to provide enough methodological and experimental detail to assess the proposed video-to-control interface.
We cite preprints selectively when they introduce a distinct interface design or clarify an emerging design direction; we do not aim to exhaustively catalog contemporaneous arXiv variants of already-represented designs.
To keep the survey's conclusions on archival footing, we mark each method's publication status in Table~\ref{tab:method_taxonomy}.
We treat quantitative results from preprints as indicative of emerging directions rather than as a basis for comparative ranking; comparative conclusions in the synthesis rest primarily on peer-reviewed findings, with preprint evidence used only to illustrate design trends and flagged as such.

\paragraph{Search and screening protocol.}
We identified candidate papers through Google Scholar, arXiv, and Semantic Scholar, supplemented by conference proceedings and reference chasing from recent learning-from-video, vision--language--action (VLA), and robot-learning surveys.
Search terms combined keywords such as ``learning from video,'' ``imitation learning from video,'' ``video prediction,'' ``action-free video,'' ``latent action,'' ``video-to-action,'' ``world-action model,'' ``vision--language--action,'' ``subgoal image,'' ``visual planning,'' ``point/flow tracking,'' and ``trajectory prediction'' with ``robot manipulation'' or ``robot learning.''
We considered papers available up to 25~May 2026, and use this date as the reference point when describing the publication status of recent works.
Candidate papers were screened first for temporal video supervision, then for whether the learned video-derived signal serves as a control interface for manipulation, and finally for whether the method provides enough methodological and experimental detail to assess its grounding into robot control.
When multiple contemporaneous papers represented similar designs, we selected representative examples that clarify distinct interface choices rather than aiming for exhaustive coverage.

\paragraph{How included papers map to our taxonomy}
Included works fall into three families depending on where the interface between video dynamics and robot actions is constructed:
(1) \emph{direct video--action policies}, which keep the interface implicit;
(2) \emph{latent-action methods}, which route transitions through a compact learned intermediate; and
(3) \emph{explicit visual interfaces}, which predict interpretable targets for a downstream controller to track.

\paragraph{Exclusion criteria and boundary cases.}
To keep the survey focused on temporal video-derived supervision, we treat as out of scope:
(i) methods that rely only on static (single-image) affordances, keypoints, or segmentation cues without learning a temporal predictive model (e.g., MOKA~\citep{moka_RSS2024}, FlowBot3D~\citep{flowbot3d_RSS_2022}, KETO~\citep{keto_ICRA2020}, ReKep~\citep{ReKep_Huang_2024});
(ii) works that use temporal or self-supervised objectives primarily to learn visual state representations, while learning policies mainly from robot interaction or action-labeled robot data (e.g., ManipulateBySeeing~\citep{ManipulateBySeeing_ICCV_2023}, GENIMA~\citep{GENIMA_CoRL_2024}, DynaMo~\citep{DynaMo_cui2024});
(iii) approaches that learn rewards or values from video without learning predictive models or temporal control interfaces (e.g., VIP~\citep{VIP_ICLR_2023}, LIV~\citep{LIV_ma_2023});
(iv) general video world-model or ``universal simulator'' efforts whose primary goal is action-conditioned simulation for reinforcement learning or MPC, rather than transferring action-free video priors into manipulation interfaces (e.g., UniSim~\citep{UniSim_ICRA_2024}, PointWorld~\citep{pointworld_huang2026});
and (v) large-scale vision--language--action models trained primarily on action-labeled robot demonstrations (e.g., RT-1~\citep{RT1_Brohan_2022}, RT-2~\citep{RT2_Brohan_2023}, $\pi_0$~\citep{pi0_Black_2024}), which we cite as motivating context (\S\ref{sec:intro}) but exclude because their dominant supervision signal is robot action data rather than action-free video.

We also treat several nearby directions as boundary cases. Single-demonstration imitation systems that retarget motion from a specific human video (e.g., RSRD~\citep{RSRD_kerr2024}) focus on per-instance reconstruction rather than learning transferable interfaces from large-scale action-free video. A second boundary concerns latent-action methods whose interface is closely tied to action or robot-side supervision during latent-action learning: CLAM~\citep{CLAM_liang2025} and LAOM~\citep{LAOM_2025latent} introduce action supervision into latent-action learning, while villa-X~\citep{villaX_chen2026} incorporates robot proprioceptive states and actions when learning or grounding latent actions. These works fall outside the core scope because action labels, proprioception, or other embodiment-specific signals participate in learning the latent-action interface itself, rather than being used only after an action-free interface has been learned. While excluded from the core taxonomy, we occasionally reference such directions when they provide useful contrasts or help clarify the scope and design choices of in-scope methods.
\section{Taxonomy: The Video-to-Control Interface Spectrum}
\label{sec:taxonomy}

This survey organizes video-based robotic manipulation methods by \emph{where and how video-derived temporal structure is connected to robot control}.
Rather than categorizing work solely by model class (e.g., transformers vs.\ diffusion) or supervision type, we adopt an \emph{interface-centric} view: each method learns an intermediate representation or signal primarily from \emph{non-action-annotated} video and then \emph{grounds} it into a robot's action space.
Framing the literature through this interface highlights recurring design patterns and makes the main trade-offs explicit---data requirements for grounding, interpretability and debuggability, planning capability, and cross-embodiment transfer.
Interface location directly determines how a method enters the robot's control loop, what can be inspected or verified before execution, and where physical inconsistencies or grounding failures can arise.

\subsection{Design Axes and Design Space}
\label{subsec:design_space}
\begin{figure}
    \centering
    \includegraphics[width=1\linewidth, trim= 6 8 6 6, clip]{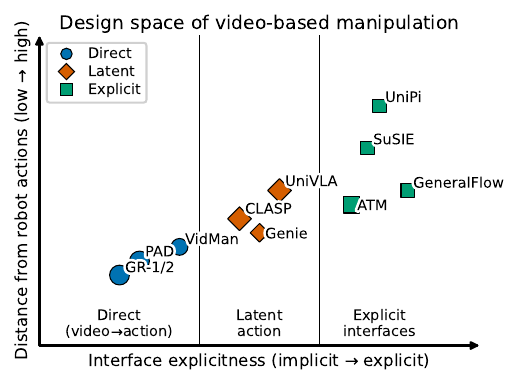}
    \vspace{-6mm}
    \caption{\textbf{Design space of video-based manipulation methods.}
    Methods are positioned by \textit{interface explicitness} (x-axis) and \textit{{distance from robot actions}} (y-axis; higher indicates that the control interface is farther from low-level actions, e.g., subgoals, trajectories, or poses, rather than direct action prediction).
    Marker shape indicates family (direct video$\rightarrow$action, latent-action, explicit interfaces).
    Marker size qualitatively reflects the typical amount of action-labeled robot data used for grounding/training (low/medium/high).
    \textit{Plotted methods are representative anchors, not exhaustive.}
}
    \label{fig:design_space}
    \vspace{-3mm}
\end{figure}

Figure~\ref{fig:design_space} situates representative methods (anchors) in a two-axis design space.
The horizontal axis measures \emph{interface explicitness} (how directly the method exposes the video$\rightarrow$control linkage), and the vertical axis measures \emph{distance from robot actions} (how far the interface lies from low-level motor commands).
Marker shape indicates the dominant family (direct video--action, latent-action, explicit interfaces), and marker size qualitatively reflects the typical amount of \emph{action-labeled} robot data used for grounding or training (low/medium/high).

\paragraph{Interface explicitness (x-axis).}
The x-axis captures how explicitly a method externalizes the connection between visual change and control.
At the left extreme, the video-to-action linkage is \emph{implicit} inside a shared representation and action head.
Moving rightward, methods introduce an explicit intermediate variable that summarizes transition structure.
At the far right, methods expose \emph{interpretable} outputs---such as subgoals, plans, trajectories, or pose sequences---that are explicitly consumed by downstream control.

\paragraph{Distance from robot actions (y-axis).}
The y-axis captures the degree of control abstraction as the separation between what is predicted/conditioned on and what the robot executes.
Lower positions correspond to interfaces that are close to motor commands (direct action prediction/decoding).
Higher positions correspond to interfaces that specify more abstract targets (e.g., subgoals, object motion, pose plans), requiring additional control modules to translate these targets into executable actions.
Higher abstraction does \emph{not} necessarily imply longer horizon; it primarily indicates that execution depends on a nontrivial translation from the interface to low-level control.

\begin{table*}[t]
  \caption{\textbf{Method taxonomy.}
  We group methods by where the video-to-control interface is defined:
  (i) \textbf{direct video--action} methods keep the interface implicit in shared features,
  (ii) \textbf{latent-action} methods expose abstract action codes, and
  (iii) \textbf{explicit visual interfaces} expose human-interpretable subgoals, plans, or trajectories. Within each subgroup, \textbf{Main} methods are discussed in detail and appear in the detailed comparison tables, while \textbf{Other} methods are covered through shorter block-style mentions. Boundary cases are marked explicitly.}
  \label{tab:method_taxonomy}
  \centering
  \footnotesize
  \setlength{\tabcolsep}{6pt}
  \renewcommand{\arraystretch}{1.30}

  \begin{tabularx}{\textwidth}{
      >{\raggedright\arraybackslash}p{0.24\textwidth}
      Y
    }
    \toprule
    \textbf{Subcategory} & \textbf{Methods} \\
    \midrule

    \rowcolor{gray!12}
    \multicolumn{2}{l}{\sagesf\bfseries Family\,I:\enspace Direct Video--Action} \\
    \addlinespace[3pt]

    Joint video--action generators &
    \textbf{Main:}\enspace
    GR-1~\citep{GR1_Wu_2023},\
    GR-2\textsuperscript{\ddag}~\citep{cheang2024gr2generativevideolanguageactionmodel},\
    PAD~\citep{PAD_guo2024_NIPS},\
    UWM~\citep{UWM_zhu2025_RSS},\
    UVA~\citep{UVA_ShuranSong_2025},\
    Cosmos Policy~\citep{kim2026cosmos},\
    Fast-WAM\textsuperscript{\ddag}~\citep{yuan2026fast_WAM}
    \newline
    \textbf{Other:}\enspace
    LDA-1B~\citep{lyu2026lda},\
    DreamZero\textsuperscript{\dag}~\citep{ye2026dreamzero},\
    X-WAM\textsuperscript{\ddag}~\citep{guo2026X_WAM} \\
    \addlinespace[2pt]

    Frozen predictive-video features &
    VidMan~\citep{VidMan_2024_NIPS},\
    VPP~\citep{VPP_Hu_2024} \\
    \addlinespace[2pt]

    Latent-state world models (\textit{boundary}) &
    APV~\citep{APV_2022},\
    ContextWM~\citep{ContextWM_NIPS_2023} \\

    \addlinespace[4pt]
    \rowcolor{gray!12}
    \multicolumn{2}{l}{\sagesf\bfseries Family\,II:\enspace Latent-Action} \\
    \addlinespace[3pt]

    Latent actions as standalone control interfaces &
    CLASP~\citep{CLASP_ICLR2019} \\
    \addlinespace[2pt]

    Latent actions in instruction-conditioned policy learning &
    \textbf{Main:}\enspace
    LAPA~\citep{LAPA_ye2025},\
    Moto~\citep{Moto_chen2025},\
    UniVLA~\citep{UniVLA_RSS_25},\
    ConLA\textsuperscript{\ddag}~\citep{ConLA_dai2026},\
    HiLAM\textsuperscript{\dag}~\citep{HiLAM_kim2026},\
    RotVLA\textsuperscript{\ddag}~\citep{RotVLA_li2026},\
    ALAM\textsuperscript{\ddag}~\citep{ALAM_tang2026}
    \newline
    \textbf{Other:}\enspace
    GO-1~\citep{GO1_Bu_IROS_2025},\ 
    CoMo\textsuperscript{\ddag}~\citep{CoMo_yang2026} \\

    \addlinespace[4pt]
    \rowcolor{gray!12}
    \multicolumn{2}{l}{\sagesf\bfseries Family\,III:\enspace Explicit Visual Interfaces} \\
    \addlinespace[3pt]

    \multirow{2}{=}{Frame-based interfaces} &
    \textit{Dense video plans:}\enspace UniPi~\citep{UniPi_Du_NIPS_2023},\
    Gen2Act~\citep{Gen2Act_bharadhwaj2024},\
    AVDC~\citep{AVDC_2023},\
    RIGVid~\citep{RIGVid_patel2025},\
    Dreamitate~\citep{Dreamitate_Liang_CoRL_2024},\
    GVF-TAPE~\citep{GVFTAPE_CoRL_2025},\
    Dream2Flow\textsuperscript{\dag}~\citep{dream2flow_dharmarajan2025} \\
    &
    \textit{Subgoal-image interfaces:}\enspace SuSIE~\citep{susie_2023},\
    V2A~\citep{V2A_2025},\
    CLOVER~\citep{CLOVER_2024} \\
    \addlinespace[4pt]

    \multirow{3}{=}{Trajectory-based interfaces} &
    \textit{Affordance trajectories:}\enspace VRB~\citep{VRB_2023},\
    SWIM~\citep{SWIM_2023} (\textit{boundary}) \\
    &
    \textit{2D pixel / object flow:}\enspace ATM~\citep{ATM_Wen_RSS_2024},\
    Im2Flow2Act~\citep{Im2Flow2Act_Xu_CoRL_2024},\
    Track2Act~\citep{track2act_ECCV_2024}
    \newline
    \textbf{Other:}\enspace Tra-MoE~\citep{tra_moe_CVPR2025} \\
    &
    \textit{3D/6D structured:}\enspace GeneralFlow~\citep{GeneralFlow_CoRL_2024},\
    SKIL-H~\citep{SKIL_RSS_2025},\
    MimicPlay~\citep{MimicPlay_CoRL_2023},\
    ZeroMimic~\citep{ZeroMimic_ICRA_2025} \\

    \bottomrule
  \end{tabularx}

  \vspace{2pt}
  \raggedright
  \scriptsize
  Superscripts denote publication status as of our 25~May 2026 literature cutoff: \textsuperscript{\dag}~workshop/non-archival release; \textsuperscript{\ddag}~preprint or arXiv-only release. Unmarked methods have a peer-reviewed archival version.
\end{table*}

\begin{figure*}[t]
    \centering
    \includegraphics[width=1\linewidth, trim= 18 20 15 15, clip]{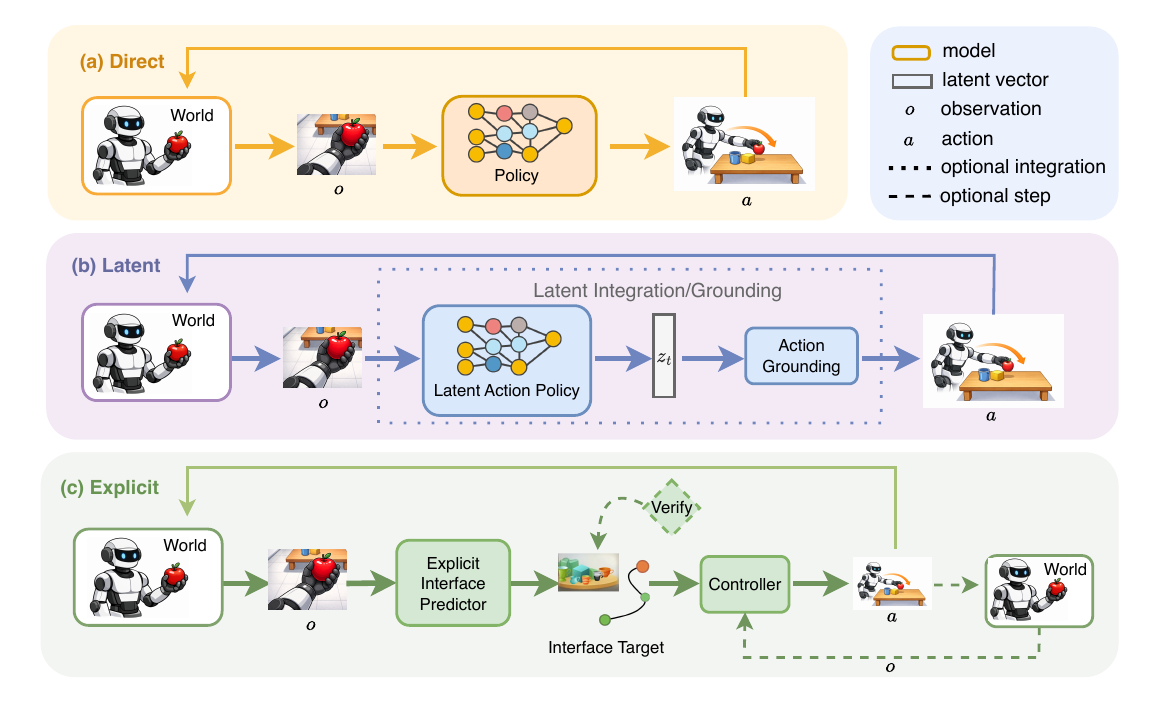}
    \caption{\textbf{Control-loop abstractions for video-based manipulation.}
    (a) \textbf{Direct} methods map observations directly to executable actions through a policy, leaving the connection between visual change and control implicit.
    (b) \textbf{Latent-action} methods introduce an intermediate latent variable $z_t$ between observations and robot commands. In standalone interfaces, $z_t$ is selected by a latent policy or planner and grounded to actions; in instruction-conditioned variants, $z_t$ may instead be explicit, internal, or training-only. The dotted box denotes an integrated latent-control interface whose components may be separate, fused, or co-trained depending on the method.
    (c) \textbf{Explicit-interface} methods predict a human-interpretable intermediate target, such as a goal image, trajectory, waypoint, flow, or keypoint, which can be verified or revised before a controller converts it into actions.}
    \label{fig:control_loop} 
\end{figure*}

\paragraph{How to interpret the figure.}
Figure~\ref{fig:design_space} is schematic: placements are approximate and intended to convey relative trends rather than precise measurements.
Family boundaries are not strict---many systems are hybrids---and we place each method by its \emph{dominant} control interface.
More explicit and/or more abstract interfaces can improve inspectability and transfer, but they may introduce additional failure modes (e.g., interface prediction errors, perception/transfer errors, or tracking and replanning instability).
Marker size is qualitative and reflects typical dependence on action-labeled robot data for grounding; it does not capture other resources such as action-free video scale, compute, or simulator interaction.
Together, these axes emphasize that video-based manipulation methods occupy regions of a continuous design space rather than perfectly discrete categories.

\subsection{Three Families of Interface Designs}
\label{subsec:three_families}

While the design space is continuous, most approaches cluster into three recurring \emph{interface design patterns}, illustrated schematically in Figure~\ref{fig:three_category_structure}. Table~\ref{tab:method_taxonomy} operationalizes this grouping for the remainder of the survey and marks boundary cases explicitly.
We define these families by the dominant interface through which video-derived temporal structure enters the robot's control loop---capturing both how training is factorized and what signal is exposed at deployment---rather than by backbone architecture alone.
Accordingly, when a method contains hybrid ingredients, we group it by the interface that carries the primary burden of grounding video-derived structure into executable control.

\paragraph{Direct video--action policies.}
Direct video--action methods offer the most immediate route from visual observation to executable control: the deployed system outputs low-level robot actions directly, without exposing an intermediate target for a downstream planner or controller.
Temporal prediction is used to shape the policy's internal representation, and grounding to the action space is typically achieved through end-to-end or interleaved training with action-labeled robot trajectories.
This keeps the deployment stack minimal, but it also makes the video-to-action linkage harder to inspect, verify, or transfer, because the operative interface lives in hidden features and embodiment-specific action heads.

\paragraph{Latent-action methods.}
Latent-action methods introduce a compact intermediate interface that can support planning, search, or policy learning before being grounded to a specific robot's control space.
They typically learn this action-like variable from observation transitions under a capacity bottleneck, and then use limited action-labeled data to ground, fine-tune, or integrate it into a deployable policy.
This factorization decouples transition understanding from embodiment-specific actuation, but it raises structural challenges---latent identifiability, controllability, and robustness of the grounding or policy-integration mechanism---especially under confounding and one-to-many futures.

\paragraph{Explicit visual interfaces.}
Explicit-interface methods expose the video-to-control interface as an interpretable target---such as a subgoal image, video plan, trajectory, or pose sequence---that a downstream controller explicitly tracks or follows.
Training is therefore often modular: a video-pretrained predictor produces the interface, sometimes with an additional transfer step (e.g., lifting 2D tracks to 6D poses), and a separate controller maps this interface plus robot state to low-level actions.
This design improves transparency and can ease cross-embodiment transfer because the interface is defined in visual or geometric space, but it also shifts the reliability burden to interface prediction and perception/transfer pipelines, often requiring execution-time tracking and replanning to control compounding errors.

\paragraph{Assignment rule and boundary cases.}
To keep family membership reproducible rather than impressionistic, we assign each method by the signal that carries the primary control burden at the final grounding stage---the point at which video-derived structure is converted into executable robot actions.
If the deployed system emits robot actions without an inspectable intermediate target consumed by a separate controller, we classify it as \textbf{direct video--action}, even when video prediction is used during training.
If the transferable object is a compact action-like code learned from transitions, we classify it as \textbf{latent-action}.
If the method exposes a visual or geometric target that a downstream controller tracks, we classify it as an \textbf{explicit visual interface}.

This rule resolves the boundary cases marked in Table~\ref{tab:method_taxonomy}.
APV and ContextWM are treated as latent-state world-model boundary cases rather than latent-action methods, because their latent variables form the predictive \emph{state} used for policy learning rather than an action-like interface.
UVA and Fast-WAM remain direct video--action methods because test-time control consumes hidden representations\,/\,action heads rather than explicit visual targets; in Fast-WAM, the video-generation branch is not required at deployment.
SWIM is treated as an explicit-interface boundary case because the exposed control object is affordance- or trajectory-like and is consumed by downstream planning and control.

\paragraph{Control-loop perspective.}
Beyond structural design, each family implies a distinct integration pattern with the robot's control loop.
Figure~\ref{fig:control_loop} illustrates these differences. Direct models collapse perception and action into a single inference call whose execution mode---stepwise, chunked, or receding-horizon---determines the length of open-loop intervals between re-observations; latent-action methods insert an action-like variable that may be planned over, decoded, used as policy supervision, retained as a token, or co-generated with robot actions, so control quality depends on both latent identifiability and grounding alignment; and explicit interfaces impose a two-level hierarchy where a controller tracks predicted targets, creating opportunities for pre-execution verification but introducing tracking error and perception-pipeline fragility.
We analyze these control-integration properties in detail within each family section (\S\ref{subsec:direct-control-integration}, \S\ref{subsec:latent-control-integration}, \S\ref{subsec:explicit-control-integration}) and synthesize them comparatively in \S\ref{subsec:control-comparison}.

\paragraph{Parallels to established robotics control patterns.}
The three families also align with familiar patterns in the robotics control literature.
Direct video--action policies are closest to end-to-end visuomotor control or learned visual servoing: perception is mapped to action in a single policy, with no explicit intermediate target available for planning or verification.
Latent-action methods occupy an intermediate position: standalone variants resemble learned model-predictive control over a compressed action-like space, while instruction-conditioned variants resemble representation- or skill-conditioned policy learning.
Explicit visual interfaces follow a hierarchical planner--controller decomposition familiar from visual servoing and task-space control: the video model proposes a visual or geometric target, and a downstream controller tracks it in robot space.
These parallels clarify both the appeal and the limitations of the three families: learned methods inherit flexibility and scalability from video pretraining, but they still typically lack the stability guarantees, explicit constraint enforcement, and reusable compositional structure of classical robotics pipelines.

\begin{figure*}
    \centering
    \includegraphics[width=1\linewidth, trim=70 270 70 270, clip]{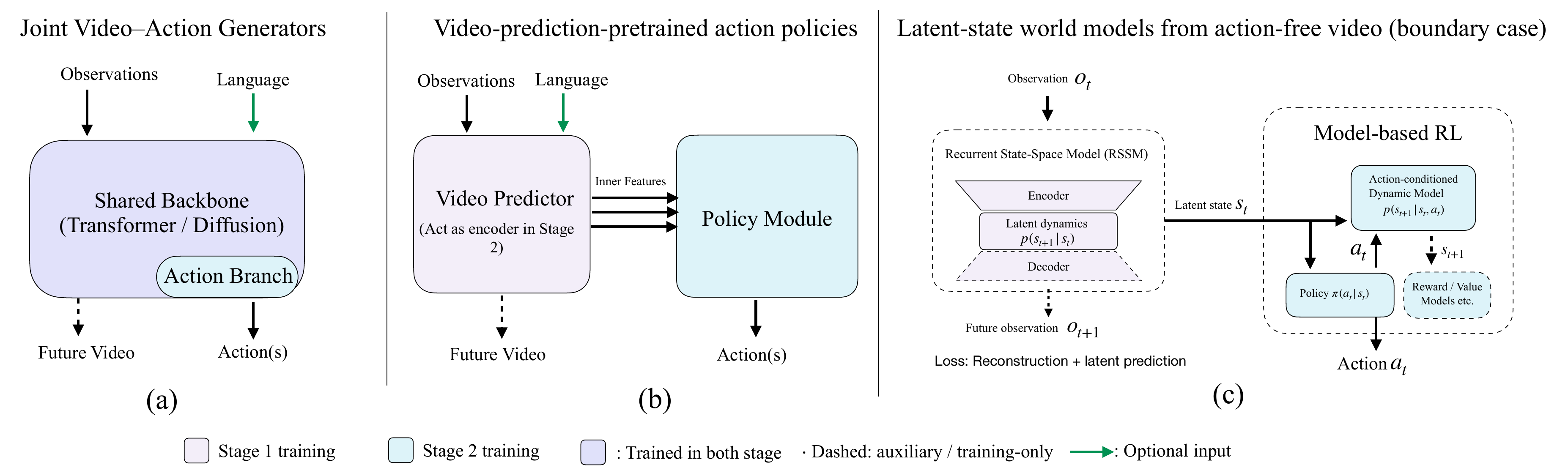}
    \vspace{-5mm}
    \caption{\textbf{Taxonomy of direct video--action modeling approaches.} \textbf{Left: }Joint video--action generators jointly predict future video and actions using a shared backbone, typically trained end-to-end on paired observation--action data. \textbf{Middle:} Video-prediction-pretrained action policies first learn a video prediction model for representation learning (stage 1), then train a policy on top of the learned features (stage 2), with video prediction serving as auxiliary supervision. \textbf{Right:} Latent-state world models from action-free video (boundary case) pretrain a latent world model from passive video via reconstruction and latent prediction objectives (stage 1), and subsequently learn action-conditioned dynamics and policies using model-based reinforcement learning in latent space with robot interaction data (stage 2). Unlike direct video--action generators, actions are not predicted directly from video but are produced via latent-space control, motivating their inclusion as a boundary case. Note: Some joint generators in (a) are trained in a single stage, jointly learning video and action prediction.}
    \label{fig:direct_video_action_pipeline}
\end{figure*}

\paragraph{Use throughout the survey.}
We use this interface-centric taxonomy to structure the remainder of the paper: Sections~\ref{sec:direct-video-action}--\ref{sec:visual-interfaces} analyze each family in detail, including both interface design and control-integration properties; Section~\ref{sec:datasets} catalogs the datasets and benchmarks used across families; and Section~\ref{sec:discussion_future} synthesizes cross-family design axes, deployment challenges, and future research directions.

\section{Direct Video--Action Policies}
\label{sec:direct-video-action}

Direct video--action policies represent the most tightly coupled way of integrating video-derived supervision into robot control. They map observations directly to actions in the robot's native control space, while using temporal video prediction to shape the internal visual representation. Manipulation is thus modeled as a visuomotor sequence prediction problem in which perception, dynamics modeling, and action generation are learned within systems that ultimately produce executable commands. This family also connects to the emerging notion of \emph{world--action models}, particularly its joint video--action generators that couple video/world prediction with action generation within a unified learned system.

A central assumption in this family is that predicting how a scene evolves over time encourages the model to encode dynamics-relevant structure—such as object motion, contact transitions, and interaction outcomes—within its internal representation. When robot demonstrations are introduced, these representations can then be grounded to actions without introducing an explicit intermediate interface such as latent actions, trajectories, or subgoal states.

Formally, let $o_t \in \mathcal{O}$ denote observations, $a_t \in \mathcal{A}$ robot actions, and $l$ an optional language instruction. Training often combines two complementary data sources: large corpora of action-free videos that supervise temporal prediction, and smaller robot datasets that provide action labels or interaction feedback. The policy thus learns visual--temporal structure from video while relying on robot demonstrations to anchor these representations to executable control signals.

Predicting actions directly from the learned representation simplifies inference because it avoids an additional planning module or explicit intermediate control interface. The same tight coupling, however, leaves the relationship between predicted visual changes and executable robot commands implicit in the policy parameters, which makes the resulting control strategy harder to interpret, debug, or modify through explicit planning.

We reserve the term \emph{direct video--action} for approaches where video-derived temporal structure is grounded in the robot's native action space without exposing a dedicated inspectable visual interface or a learned latent-action variable as the deployment-time control signal. Action-free world-model pretraining methods (APV, ContextWM) are included as boundary cases: they introduce internal latent-state planning rather than direct action decoding, but the final control output remains in native action space and no separate latent-action or explicit visual target mediates the control loop.

\paragraph{Organization.}
We organize direct video--action methods by how tightly temporal prediction is coupled to action generation at deployment. We first cover \emph{joint video--action generators}, in which a single backbone models both future visual observations and future actions. We then discuss \emph{frozen-feature policies} that freeze a video predictor and train a lightweight action module on top for closed-loop control. Finally, we include \emph{latent-state world models} pretrained from action-free video as boundary cases, which ground actions through model-based RL and latent-space rollouts rather than directly training an action predictor.

The overall pipeline of the three method types is summarized in Figure~\ref{fig:direct_video_action_pipeline}: joint generators (Fig.~\ref{fig:direct_video_action_pipeline}a) couple video prediction and action generation in a shared backbone; frozen-feature policies (Fig.~\ref{fig:direct_video_action_pipeline}b) freeze a video predictor and learn a lightweight policy on its predictive features; and boundary latent-state world models (Fig.~\ref{fig:direct_video_action_pipeline}c) use action-free video to pretrain latent dynamics and ground actions through model-based RL. Table~\ref{tab:direct-video-action-comparison} provides a structural overview (including the deployment \emph{execution} mode), Table~\ref{tab:direct-video-action-capabilities} summarizes training sources and transfer mechanisms (including real-robot evaluation), and Table~\ref{tab:direct_video_action_quant} gives a non-leaderboard snapshot of reported results where comparable numbers are available.

\subsection{Joint Video--Action Generators}

Methods in this subsection correspond to Fig.~\ref{fig:direct_video_action_pipeline}a: a single backbone is trained so that temporal visual prediction shapes the same internal representation used for action generation. The main design variation lies in how video-derived temporal structure is coupled to action prediction---for example, shared autoregressive token modeling (GR-1/2), joint diffusion over future images and actions (PAD), modality-specific diffusion timesteps enabling different predictive queries (UWM), or decoupled video/action heads on a shared latent representation (UVA) (Table~\ref{tab:direct-video-action-comparison}). These coupling choices also shape the most natural deployment loop: autoregressive models readily support stepwise or chunked decoding, diffusion policies are often run in receding-horizon form, and decoupled designs make action-only inference easier by bypassing explicit video generation at test time (Table~\ref{tab:direct-video-action-comparison}).

These methods also connect to the emerging terminology of \emph{world--action models}: learned systems that/ couple world or video prediction with action generation. The label fits most directly for UWM, PAD, and UVA, while GR-1/2 can be viewed as earlier instances of the same broad joint video--action direction.

\subsubsection{GR-1: GPT-style joint video and action prediction.}
\label{subsec:gr1}

GR-1~\citep{GR1_Wu_2023} establishes a basic template for direct video--action policies: use future-video prediction as an auxiliary objective inside the same autoregressive policy that generates robot actions.
Rather than exposing an intermediate plan, the method couples visual forecasting and control in a shared token space, so the temporal structure learned from video is grounded to actions through the internal representation of a single backbone.

Concretely, GR-1 conditions on language, past observations, and robot state, and predicts future frames and robot actions.
The video objective encourages the shared representation to encode dynamics-relevant cues such as object motion, contact transitions, and interaction outcomes, while robot demonstrations anchor that representation to executable commands.
Training follows this logic in two stages: large-scale egocentric video pretraining (e.g., Ego4D~\citep{Ego4d_CVPR_2022}) first teaches language-conditioned visual prediction, and robot fine-tuning then adds action supervision while keeping future-frame prediction in the objective.

Within this family, GR-1 is important as an early instance of shared-token coupling between visual forecasting and action generation.
Its strength is that video-pretrained temporal structure can improve generalization once grounded with demonstrations; its limitation is that the video-to-action link remains implicit, making the policy difficult to inspect or intervene on at deployment.

\begin{table*}[t]
  \caption{\textbf{Structural comparison of direct video--action methods.} Methods are grouped by architectural paradigm: \emph{joint} methods generate video and actions in a shared model, and \emph{frozen-feature} methods pre-train a video predictor and train a separate action head on its frozen predictive features. ``Video at Inference'' indicates whether explicit video generation is required during deployment: \emph{Optional} means the method can bypass video generation and decode actions directly.}
  \label{tab:direct-video-action-comparison}
  \centering
  \footnotesize
  \setlength{\tabcolsep}{4pt}
  \renewcommand{\arraystretch}{1.25}

  \begin{tabularx}{\textwidth}{
    @{}
    >{\raggedright\arraybackslash}p{3.2cm}
    Y
    >{\raggedright\arraybackslash}p{4.0cm} 
    >{\centering\arraybackslash}p{1.3cm}
    >{\raggedright\arraybackslash}p{3.7cm}
    >{\centering\arraybackslash}p{1.5cm}
    @{}
  }
    \toprule
    \textbf{Method} & \textbf{Architecture} & \textbf{Video--Action Coupling} & \textbf{Training} & \textbf{Video at Inference} & \textbf{Execution} \\
    \midrule
    \rowcolor{gray!8}
    \multicolumn{6}{l}{\textit{Joint video--action generators}} \\
    \addlinespace[2pt]
    GR-1~\citep{GR1_Wu_2023}           & AR Transformer  & Joint (shared GPT)        & Two-stage  & Optional (action-only decoding)  & Stepwise \\
    GR-2~\citep{cheang2024gr2generativevideolanguageactionmodel} & AR Transformer  & Joint (shared GPT)        & Two-stage  & Optional (action-only decoding)  & Chunked  \\
    PAD~\citep{PAD_guo2024_NIPS}        & Diff.\ Transformer & Joint denoising, masked modalities & Two-stage  & Optional (action-only denoising)  & Receding \\
    UWM~\citep{UWM_zhu2025_RSS}         & Diff.\ Transformer & Coupled diff.\ via timesteps    & End-to-end & Optional  & Chunked  \\
    UVA~\citep{UVA_ShuranSong_2025}     & Latent + diff.\ heads & Joint latent, decoupled decoding & End-to-end & No (action head only)        & Chunked  \\
    Cosmos Policy~\citep{kim2026cosmos} & Diff.\ Transformer & Latent-frame interface (non-image quantities as video-sequence frames) & Two-stage + RL rollouts & Optional (direct or planner mode) & Chunked / value-guided planner \\
    Fast-WAM~\citep{yuan2026fast_WAM}   & Diff.\ Transformer (MoT) & Co-trained, attention-masked & Two-stage & No (action-only)        & Chunked \\
    \addlinespace[4pt]
    \midrule
    \rowcolor{gray!8}
    \multicolumn{6}{l}{\textit{Frozen predictive-video features (video pre-training $\rightarrow$ action head)}} \\
    \addlinespace[2pt]
    VidMan~\citep{VidMan_2024_NIPS}     & Diff.\ Transformer & Adapter over video features    & Two-stage  & No        & Feat.-cond. \\
    VPP~\citep{VPP_Hu_2024}            & Video diff.\ + policy & Video features $\rightarrow$ action head & Two-stage  & No (representations only)        & Feat.-cond. \\
    \addlinespace[4pt]
    \midrule
    \rowcolor{gray!8}
    \multicolumn{6}{l}{\textit{Action-free pre-training for world-model RL}} \\
    \addlinespace[2pt]
    APV~\citep{APV_2022}               & RSSM             & Stacked (AF $\rightarrow$ AC + RL)  & Two-stage  & No        & Latent rollouts \\
    ContextWM~\citep{ContextWM_NIPS_2023} & RSSM (+ context) & Stacked with context           & Two-stage  & No        & Latent rollouts \\
    \bottomrule
  \end{tabularx}

  \vspace{2pt}
  \raggedright
  \scriptsize
  \textit{AR} = Autoregressive; \textit{Diff.} = Diffusion; \textit{MoT} = Mixture-of-Transformer; \textit{RSSM} = Recurrent State-Space Model; \textit{AF} = Action-free; \textit{AC} = Action-conditional; \textit{Feat.-cond.} = feature-conditioned action head.
\end{table*}

\subsubsection{GR-2: Scaling joint video--action models with web-scale data.}
\label{subsec:gr2}

GR-2~\citep{cheang2024gr2generativevideolanguageactionmodel} tests the scalability of the GR-1 coupling pattern: whether a single autoregressive backbone can continue to benefit from joint video forecasting and action prediction when both action-free video pretraining and robot grounding data are expanded substantially.
Like GR-1, it treats future video and robot actions as output modalities of the same transformer, but shifts the emphasis from establishing the formulation to scaling its data regime and execution horizon.

The video-pretraining stage is broadened from egocentric video to a mixture of large Internet video corpora and robot video data, with frames discretized by an image tokenizer and predicted autoregressively under language and observation context.
Robot fine-tuning then grounds the shared video--action representation to executable commands, while retaining future-frame prediction as part of the objective so that temporal visual structure continues to regularize action learning.

A second change is the move from single-step action prediction to short action chunks.
This does not introduce a new interface---actions are still generated directly in the robot's native action space---but changes the deployment behavior by improving temporal consistency and reducing the need for per-step replanning, at the cost of longer open-loop segments between observations.

GR-2 suggests that the GR-1 template can scale with broader video and robot data, while action-chunk supervision makes the same joint video--action formulation more stable during execution.

\subsubsection{PAD: joint denoising of future images and actions.}
\label{subsec:pad}
PAD~\citep{PAD_guo2024_NIPS} probes the role of generative architecture in tightly coupled video--action policies: keeping the joint video--action coupling of GR-1/2, it replaces autoregressive token prediction with joint diffusion over future images and actions.
Instead of predicting visual and action tokens step by step, PAD denoises future observations and action sequences together within a shared transformer, so video prediction and control remain tied inside one generative process.

This formulation makes action-free video useful through masked co-training.
For video-only samples, the action branch is masked and supervision is applied only to future image prediction; for robot demonstrations, both future images and actions are denoised and supervised.
The result is the same basic grounding logic as GR-style models---large-scale video shapes temporal visual representations, while robot trajectories anchor them to executable actions---but implemented through a diffusion objective that can represent multimodal futures.

At execution time, PAD uses a receding-horizon loop: it predicts a short horizon of future images and actions, executes the first action, and replans from the next observation.
PAD thereby changes how uncertainty is represented: rather than generating one future token trajectory per autoregressive rollout, it treats images and actions as a jointly denoised future, providing a natural route to multimodal prediction but making each closed-loop update computationally heavier.

\subsubsection{UWM: queryable video--action diffusion.}
\label{subsec:uwm}

UWM~\citep{UWM_zhu2025_RSS} extends PAD-style joint diffusion by turning joint video--action generation into a queryable predictive model.
Rather than denoising video and actions jointly for every use case, UWM treats the same shared model as a conditional predictor whose generated and conditioned modalities determine whether it functions as a policy, video predictor, forward dynamics model, or inverse-dynamics-style model.
This shifts the direct video--action formulation toward a more world-model-like role while preserving direct action generation in the robot's native control space.

The key mechanism is modality-specific diffusion within a shared transformer: future video latents and future action sequences share parameters, but their diffusion processes are controlled separately through different timestep schedules and selective masking.
When both modalities are available, UWM trains the coupled video--action model; when only video is available, the action branch is masked through its diffusion timestep while the visual branch remains supervised.

Compared with PAD, UWM makes the same video--action coupling available across more predictive queries, but this flexibility stays internal to the network: it exposes different functions of one learned model, not an inspectable plan or constraint-aware controller for checking feasibility, contact, or safety before execution.

\begin{table*}[t]
  \caption{\textbf{Training sources and transfer mechanisms of direct video--action methods.} 
  ``Video Pre-training Data'' describes the source of action-free video for learning visual dynamics.
  ``Action Supervision'' specifies how action labels are obtained.
  ``Video$\rightarrow$Action Transfer'' summarizes how video representations contribute to action prediction.
  ``Task Scope'' summarizes the broad task family reported in the original papers.
  ``Real-Robot Scope'' summarizes the real-robot experiments reported (or their absence).}
  \label{tab:direct-video-action-capabilities}
  \centering
  \footnotesize
  \setlength{\tabcolsep}{6pt}
  \renewcommand{\arraystretch}{1.15}

  \begin{tabularx}{\textwidth}{@{\extracolsep{\fill}}
    >{\raggedright\arraybackslash}p{1.8cm}
    >{\raggedright\arraybackslash}p{2.7cm}
    >{\raggedright\arraybackslash}p{1.7cm}
    >{\raggedright\arraybackslash}X
    >{\raggedright\arraybackslash}p{2.1cm}
    >{\centering\arraybackslash}p{2.4cm}}
    \toprule
    \textbf{Method} & 
    \textbf{Video Pre-training Data} & 
    \textbf{Action Supervision} & 
    \textbf{Video$\rightarrow$Action Transfer} & 
    \textbf{Task Scope} &
    \textbf{Real-Robot Scope} \\
    \midrule

    \rowcolor{gray!8}
    \multicolumn{6}{l}{\textit{Joint video--action generators}} \\
    \addlinespace[2pt]

    GR-1~\citep{GR1_Wu_2023} 
      & Ego4D (egocentric human video) 
      & Robot demos
      & Shared token space; video features $\rightarrow$ action tokens
      & Tabletop and articulated manip.
      & Object transport; articulated manip. \\
    \addlinespace[2pt]

    GR-2~\citep{cheang2024gr2generativevideolanguageactionmodel} 
      & Web-scale video + robot video 
      & Multi-robot demos 
      & Shared transformer; video--language features $\rightarrow$ action tokens
      & General tabletop manip.
      & 105 tabletop tasks + bin picking \\

    PAD~\citep{PAD_guo2024_NIPS} 
      & Robot demos + action-free video 
      & Robot demos
      & Joint denoising over video+action tokens (masked on video-only data)
      & Tabletop manip.
      & 4 Panda task groups 
      \\
    \addlinespace[2pt]

    UWM~\citep{UWM_zhu2025_RSS} 
      & Internet video + robot video 
      & Robot demos
      & Coupled diffusion; modality-specific timesteps
      & Tabletop manip.
      & 5 Franka DROID tasks \\
    \addlinespace[2pt]

    UVA~\citep{UVA_ShuranSong_2025} 
      & Robot video (action-free acceptable) 
      & Robot demos
      & Shared encoder; separate video/action heads
      & General manip.
      & 3 ARX X5 UMI tasks \\
    \addlinespace[2pt]

    Cosmos Policy~\citep{kim2026cosmos}
      & Web-scale video
      & Robot demos + on-policy rollouts
      & Latent-frame interface; actions/states/value as extra frames
      & Bimanual and multi-suite manip.
      & 4 ALOHA bimanual tasks \\
    \addlinespace[2pt]

    Fast-WAM~\citep{yuan2026fast_WAM}
      & Web-scale video + cross-embodiment video
      & Robot demos
      & Co-trained action expert (masked future-video access); action-only inference
      & Long-horizon manip.
      & Galaxea R1 towel folding \\

    \addlinespace[4pt]
    \midrule
    \rowcolor{gray!8}
    \multicolumn{6}{l}{\textit{Frozen predictive-video features (video pre-training $\rightarrow$ action head)}} \\
    \addlinespace[2pt]

    VidMan~\citep{VidMan_2024_NIPS} 
      & OXE robot video (action-free) 
      & Robot demos
      & Frozen diffusion features $\rightarrow$ action adapter (inverse dynamics)
      & General manip.
      & \textit{--} \\
    \addlinespace[2pt]

    VPP~\citep{VPP_Hu_2024} 
      & Human and robot video 
      & Robot demos
      & Video foundation-model features condition action diffusion (implicit inverse dynamics)
      & General + dexterous manip.
      & 30+ Panda + 100+ dexterous-hand tasks \\

    \addlinespace[4pt]
    \midrule
    \rowcolor{gray!8}
    \multicolumn{6}{l}{\textit{Action-free pre-training for world-model RL (boundary cases)}} \\
    \addlinespace[2pt]

    APV~\citep{APV_2022} 
      & RLBench (simulated video) 
      & RL interaction
      & Action-free world-model pretraining $\rightarrow$ Dreamer-style latent state
      & Simulated manip. and loco.
      & \textit{--} \\
    \addlinespace[2pt]

    ContextWM~\citep{ContextWM_NIPS_2023} 
      & Something-Something V2 (in-the-wild) 
      & RL interaction
      & Factorized context/dynamics $\rightarrow$ world-model RL
      & Simulated manip., loco., and driving
      & \textit{--} \\

    \bottomrule
  \end{tabularx}

  \vspace{3pt}
  \raggedright
  \footnotesize
  \textit{manip.} = manipulation; \textit{loco.} = locomotion; \textit{demos} = demonstrations; \textit{RL} = reinforcement learning; \textit{--} = no real-robot execution evaluation reported.
\end{table*}

\subsubsection{UVA: shared representation with separate video and action heads.}
\label{subsec:uva}

UVA~\citep{UVA_ShuranSong_2025} shares a similar world-model-style motivation with UWM---one learned system that can support policy, video prediction, and dynamics modeling queries---but separates video and action earlier in the architecture.
This design targets the deployment cost of fully coupled video--action generation while keeping video prediction close enough to shape the representation used for control.

In UVA, video and action processing share a common temporal representation, after which separate lightweight heads predict future video and actions.
During training, video forecasting, including from action-free video, shapes the shared backbone, while robot demonstrations ground the action head to controls in the robot's native action space.
At deployment, the policy can invoke only the action head, avoiding iterative generation of future video and making inference lighter than designs that keep video and action denoising coupled at test time.

Compared with UWM, UVA's factorization makes action-only inference cheaper and separates video prediction from action decoding more explicitly.
The likely trade-off is that visual prediction affects control mainly through the shared representation; if that representation drops control-relevant dynamics, the separated action head cannot recover them from an explicit predicted future at deployment.

\subsubsection{Cosmos Policy: latent-frame interfaces for policy and planning.}
\label{subsec:cosmos_policy}

Cosmos Policy~\citep{kim2026cosmos} extends the joint-generator template by expanding what the shared video diffusion backbone is asked to generate.
Instead of attaching a conventional action head to a video model, it preserves the pretrained model's latent-frame sequence format and inserts actions, future proprioception, future observations, and value estimates as additional latent frames, denoised jointly with the visual latents.
This turns the shared backbone into a common interface for policy learning, world modeling, and value-guided planning, rather than treating video prediction as a separate auxiliary branch or frozen feature provider.

The control pipeline is therefore organized around a common latent-frame interface.
Robot demonstrations supervise the action-producing part of the sequence, while on-policy rollouts further train the world model and value estimates on states that demonstrations may not cover.
At deployment, the model can either execute a denoised action chunk directly or evaluate multiple imagined candidates and choose the one with the highest predicted value.

Cosmos Policy's benefit is a unified interface for policy learning and planning on top of a pretrained video backbone; the trade-off is that the coupling still remains internal to the learned latent sequence, so feasibility, contact, and safety constraints are not exposed as separate control objects unless additional checking or planning structure is added.

\subsubsection{Fast-WAM: video prediction as training-time supervision.}
\label{subsec:fast_wam}

Fast-WAM~\citep{yuan2026fast_WAM} asks whether the benefit of joint video--action coupling comes from training-time video supervision or from generating future videos at test time. The system keeps video prediction during training but disables it at test time, running only the action branch. Architecturally, Fast-WAM separates a video diffusion backbone from an action expert and uses structured attention masking to prevent the action branch from relying on privileged future-video tokens during training.

Fast-WAM reports that removing video co-training during \emph{training} degrades downstream control performance more than removing future-video generation at \emph{test time}, suggesting that the joint video objective primarily shapes the learned action representation rather than providing useful test-time predictions.

This result supports the interpretation that, for joint video--action policies, temporal prediction functions primarily as a representation-shaping mechanism rather than as an explicit plan consumed at execution time.
Within this family, UVA arrives at action-only inference architecturally; Fast-WAM supports the same choice empirically.

\paragraph{Other extensions.}
Several recent methods retain the joint video--action template while exploring extensions around structured visual latents, larger pretrained video backbones, and richer spatial outputs.
LDA-1B~\citep{lyu2026lda} preserves UWM's multi-query world--action formulation but moves future prediction from pixel or VAE latents to structured DINO features, exploring feature-space prediction as a scalable interface for heterogeneous robot, human, and action-free video data.
DreamZero~\citep{ye2026dreamzero} scales WAM-style joint video--action generation with a large pretrained video diffusion backbone and system-level optimizations for real-time closed-loop control, providing a workshop example of treating tightly coupled video--action generators as feedback policies rather than only offline predictors.
X-WAM~\citep{guo2026X_WAM}, a recent preprint, extends WAM-style joint generation from RGB video to multi-view RGB-D futures, adding spatial reconstruction and asynchronous denoising so that actions can be decoded faster than high-fidelity visual outputs.
Together, these methods illustrate emerging design directions rather than serving as primary evidence for the section's comparative conclusions.

\paragraph{Takeaways for joint generators.}
Across joint video--action generators, the recurring pattern is not that generated future video is itself a reliable executable plan, but that future-prediction objectives help shape the representation used for action decoding.
The methods differ in how they implement this coupling---for example, through autoregressive token prediction, joint or modality-specific diffusion, latent-frame packaging, feature-space prediction, or action-only inference paths---yet many converge on chunked, action-only, or otherwise accelerated deployment rather than full future-video generation at every control step.
This interpretation is supported more by mechanism-level evidence than by aggregate benchmark scores: UVA reaches action-only deployment by architecturally separating its video and action heads, while Fast-WAM's ablations---reported in a recent preprint---similarly isolate training-time video co-training from test-time generation.
Accordingly, Table~\ref{tab:direct_video_action_quant} summarizes reported evidence and evaluation settings rather than ranking methods or isolating the effect of the video-prediction interface, because these numbers also vary with robot embodiment, task suite, data budget, and pretraining corpus.
Taken together, these results suggest that temporal video prediction is better supported as a mechanism for improving visuomotor \emph{representations} than as a source of explicit planning or guaranteed physical feasibility.
Because no inspectable intermediate is exposed, reachability, contact consistency, and safety remain the responsibility of the surrounding control stack rather than of the video-to-action interface itself.

\subsection{Action policies on frozen predictive-video features}
\label{subsec:vidman_vpp}

This subsection corresponds to Fig.~\ref{fig:direct_video_action_pipeline}b: video prediction is used primarily to learn dynamics-aware representations, and a separate policy module is trained on top for control. The key difference from joint generators is where the coupling occurs: joint generators co-train temporal prediction and action generation within a shared backbone, whereas these methods freeze a pretrained video predictor and learn a lightweight action head that consumes its internal predictive features. At deployment, both families may bypass explicit video generation (Table~\ref{tab:direct-video-action-comparison}); however, the frozen-feature design explicitly separates the (frozen) visual dynamics model from the control module. VidMan uses frozen diffusion features with lightweight adapters, while VPP conditions an action diffusion policy on predictive representations from a video foundation model. Table~\ref{tab:direct-video-action-comparison} and Table~\ref{tab:direct-video-action-capabilities} summarize this factorization and transfer mechanism. 

Compared with explicit-interface methods that expose inspectable intermediate targets (e.g., goal images, trajectories, or other structured plans), these frozen-feature approaches keep the video-to-control connection implicit: the policy consumes internal representations from the video model (hidden states / predictive embeddings) rather than an intermediate signal meant to be interpreted or edited.

\subsubsection{VidMan: video diffusion for robot manipulation.}
\label{subsec:vidman}

VidMan~\citep{VidMan_2024_NIPS} instantiates this frozen-feature design by using a pretrained video diffusion model as a fixed, dynamics-aware encoder for control. The key idea is to reuse the temporal structure learned by video prediction---how scenes evolve under interaction---as input features for an action predictor, without requiring video generation at deployment. VidMan motivates this factorization using a dual-process analogy: a ``slow'' video predictor learns temporally predictive dynamics, and a ``fast'' action head reuses intermediate features for real-time control.

In the first stage, VidMan trains an Open-Sora-style video diffusion transformer~\citep{OpenSora_zheng2024} on robot videos (OXE~\citep{OXE_ICRA_2024}) to predict future visual trajectories. In the second stage, the video model is frozen and lightweight layer-wise self-attention adapters are inserted to predict robot actions from intermediate features, effectively learning an inverse-dynamics-style mapping conditioned on temporally predictive representations. At deployment, the action module runs in a single forward pass without iteratively denoising future frames, enabling higher-frequency closed-loop control than full video generation.

On CALVIN~\citep{CALVIN_2022} and subsets of OXE~\citep{OXE_ICRA_2024}, VidMan reports improved action prediction and task performance over action-only baselines (e.g., improvements over prior video-pretrained baselines on CALVIN), suggesting that diffusion-based video prediction can provide useful dynamics-aware features for closed-loop action generation. The main trade-off of this factorization is that the frozen video representations may not align perfectly with downstream control objectives, so performance can be sensitive to embodiment and domain shift.

\begin{table*}[t]
    \centering
    \footnotesize
    \setlength{\tabcolsep}{4pt}
    \renewcommand{\arraystretch}{1.25}
\caption{\textbf{Reported quantitative snapshots for direct video--action methods.} These results are \emph{not} a leaderboard: papers differ in robot/data budgets, pretraining corpora, embodiments, and evaluation protocols. We use the numbers only to indicate evidence type and maturity---within-paper claims, weakly comparable shared-benchmark slices (e.g., CALVIN, LIBERO), and deployment-feasibility evidence. Entries are reported as stated in the original papers; ``Source'' gives the original location.}
    \label{tab:direct_video_action_quant}
    \begin{tabularx}{\textwidth}{
        @{}
        >{\raggedright\arraybackslash}p{3.2cm}
        Y
        >{\raggedright\arraybackslash}p{5.2cm} 
        >{\raggedright\arraybackslash}p{1.0cm}
        @{}
    }
    \toprule
    \textbf{Method} & \textbf{Metric (as reported)} & \textbf{Setting / Protocol Note} & \textbf{Source} \\
    \midrule
    \rowcolor{gray!8}
    \multicolumn{4}{l}{\textit{CALVIN} \citep{CALVIN_2022}: \textit{long-horizon language-conditioned manipulation (ABC\,$\rightarrow$\,D)}} \\
    \addlinespace[2pt]
GR-1~\citep{GR1_Wu_2023}   & SR@1--5$^\dagger$ (\%): 85.4\,/\,71.2\,/\,59.6\,/\,49.7\,/\,40.1;\enspace Avg.Len$^\ddagger$: 3.06 & Train on 100\% ABC, test on D & Table~1 \\
VidMan~\citep{VidMan_2024_NIPS} & SR@1--5 (\%): 91.5\,/\,76.4\,/\,68.2\,/\,59.2\,/\,46.7;\enspace Avg.Len: 3.42 & Train on 100\% ABC, test on D & Table~1 \\
VPP~\citep{VPP_Hu_2024}    & SR@1--5 (\%): 96.5\,/\,90.9\,/\,86.6\,/\,82.0\,/\,76.9;\enspace Avg.Len: 4.33 & Train on 100\% ABC, test on D
    & Table~1 \\
    \addlinespace[4pt]
    \midrule
    \rowcolor{gray!8}
    \multicolumn{4}{l}{\textit{MetaWorld} \citep{yu2020meta_world}: \textit{multi-task manipulation suite (50 tasks)}} \\
    \addlinespace[2pt]
    PAD~\citep{PAD_guo2024_NIPS}  & Avg.\ SR (\%) (50 tasks): 72.5 & Single text-conditioned policy; 50 traj. per task for training & Table~1 \\
    VPP~\citep{VPP_Hu_2024}      & Avg.\ SR (\%)(50 tasks): 68.2 & Single policy; 50 traj. per task for training & Table~2 \\
    \addlinespace[4pt]
    \midrule
    \rowcolor{gray!8}
    \multicolumn{4}{l}{\textit{LIBERO} \citep{LIBERO_Liu_2023}: \textit{multi-suite manipulation (Spatial / Object / Goal / Long)}} \\
    \addlinespace[2pt]
\phantom{.}Cosmos Policy~\citep{kim2026cosmos} & SR (\%) over Sp/Ob/Go/Lo: 98.1\,/\,100.0\,/\,98.2\,/\,97.6 & 50 demos per task; direct policy setting & Table~1 \\
    \addlinespace[2pt]
Fast-WAM~\citep{yuan2026fast_WAM} & SR (\%) over Sp/Ob/Go/Lo: 98.2\,/\,100.0\,/\,97.0\,/\,95.2 & 50 demos per task; action-only inference ablation setting & Table~2 \\
    \addlinespace[4pt]
    \midrule
    \rowcolor{gray!8}
    \multicolumn{4}{l}{\textit{Other evaluations (not shared across papers)}} \\
    \addlinespace[2pt]
    GR-2~\citep{cheang2024gr2generativevideolanguageactionmodel} & Avg.\ success: 74.7\% ($>$100 tasks) & Custom Test Environment; Unseen settings; train with data augmentation & Fig. 5\\ 
    \bottomrule
    \end{tabularx}
    
    \vspace{2pt}
    \raggedright
    \scriptsize
    $^{\dagger}$Success Rate (SR) for different No. of chained instruction.
    $^{\ddagger}$Average task completion length.
    \textit{Sp\,/\,Ob\,/\,Go\,/\,Lo}: LIBERO Spatial\,/\,Object\,/\,Goal\,/\,Long suites.
\end{table*}

\subsubsection{VPP: building on video foundation models.}
\label{subsec:vpp}

VPP~\citep{VPP_Hu_2024} follows the same factorized control strategy as VidMan, but changes where the predictive representation comes from: instead of training a robot-video predictor from scratch, it adapts an existing video foundation model as the representation provider for control.
VPP therefore asks whether temporally predictive features already present in large video diffusion models can be transferred to robot action learning.

The adapted video model (e.g., Stable Video Diffusion~\citep{Stable_video_diffusion_2023}) is kept fixed during policy learning, and VPP extracts predictive representations from an early video-model pass rather than generating a full future video at deployment.
An action diffusion policy then conditions on these features to output robot actions directly in the robot's native control space.
This preserves the benefit shared with VidMan---future-aware visual features without expensive test-time video generation---while relying more heavily on the foundation model's pretrained dynamics prior.

The advantage is efficient closed-loop control with less robot-specific video-model training.
The trade-off is alignment: if the foundation model's predictive features emphasize visually plausible motion rather than control-relevant contact, embodiment, or gripper-state cues, the downstream action policy may inherit a representation mismatch despite efficient inference.

\paragraph{Takeaways for frozen-feature policies.}
Within the direct video--action family, VidMan and VPP occupy a more factorized design point than joint generators. Video prediction remains a representation-learning mechanism: at deployment the controller consumes internal features from a frozen video predictor (e.g., intermediate diffusion features or a single-pass video-model embedding) and predicts actions without explicitly generating future frames at test time. This improves inference efficiency, but shifts more responsibility onto whether the frozen video representations transfer cleanly to the target embodiment and task distribution. This factorized position is reflected in Table~\ref{tab:direct-video-action-capabilities}, which shows that transfer occurs through internal predictive features rather than explicit visual targets.

\subsection{Boundary case: Latent-state world models from action-free video}

The core methods above connect video supervision to control by learning policies that predict actions directly in the robot's native action space. As shown in Fig.~\ref{fig:direct_video_action_pipeline}c, a closely related but conceptually distinct line of work instead uses action-free video to pretrain a temporally predictive latent state that later serves as an internal world model for planning or model-based RL.

These approaches differ from joint video--action generators and frozen-feature reuse in where the video-derived structure is used. Rather than treating video prediction as auxiliary supervision for an action policy, they treat prediction as a way to learn a compact latent state with dynamics structure, and then learn action-conditioned dynamics, rewards/values, and a policy on top through interaction.

We include APV~\citep{APV_2022} and ContextWM~\citep{ContextWM_NIPS_2023} as boundary cases because they share the same training signal---temporal prediction from action-free video---but shift the deployment interface from direct action generation to latent-state planning and model-based RL. These boundary cases appear in the bottom block of Table~\ref{tab:direct-video-action-comparison}; Table~\ref{tab:direct-video-action-capabilities} highlights that actions are grounded through RL interaction rather than demonstrations.

\subsubsection{APV: action-free video prediction pretraining for latent world models.}

APV~\citep{APV_2022} uses action-free video to initialize a temporally predictive latent world model, then grounds actions later through online model-based RL rather than through supervised action prediction.
Its role is therefore different from direct video--action generators: video prediction does not directly regularize an action head, but instead shapes the latent state space in which downstream dynamics learning and planning will occur.

During pretraining, APV trains a recurrent state-space model (RSSM)~\citep{RSSM_hafner2019, DreamerV2_2022} on passive manipulation videos, encouraging the latent state to support observation reconstruction and future latent prediction without action labels.
After this initialization, an action-conditioned dynamics model, reward/value functions, and a policy are learned through DreamerV2-style model-based RL~\citep{DreamerV2_2022}.
Control is then performed by planning and updating in latent space through imagined rollouts, rather than by directly mapping pixels or video-pretrained features to actions.

This design makes long-horizon replanning more explicit than in direct action predictors, but it also changes the grounding burden: success depends on whether the video-pretrained latent dynamics remain useful after action-conditioned fine-tuning, and on whether reward/value learning can support reliable model-based control.

\subsubsection{ContextWM: contextualized world models from in-the-wild videos.}

ContextWM~\citep{ContextWM_NIPS_2023} refines APV-style action-free video pretraining for more diverse, in-the-wild video by separating static scene context from time-varying dynamics.
Its motivation is to address a weakness of passive video prediction: reconstruction losses can overfit to backgrounds, textures, and layouts, leaving the latent state less focused on transferable dynamics useful for downstream control.

As in APV, ContextWM pretrains a latent video prediction model without action labels.
The key modification is a context pathway---extracted from a separately sampled reference frame---that absorbs time-invariant appearance information, allowing the recurrent latent state to concentrate more on temporal evolution.
This factorization is intended to make the pretrained world model less sensitive to appearance variation when transferred across scenes and domains.

Downstream control follows the same boundary-case pattern as APV: an action-conditioned latent dynamics model, reward/value functions, and a policy are learned through DreamerV2-style model-based RL~\citep{DreamerV2_2022} on top of the video-pretrained state space.
Thus, ContextWM does not introduce a direct action predictor; it improves the reliability of latent world-model initialization from action-free video, while retaining the same dependence on model-based RL grounding and latent-rollout planning.

\paragraph{Takeaways for latent-state world models.}
As boundary cases within the direct video--action section, APV and ContextWM use action-free video to initialize a latent state space for model-based RL rather than to directly supervise an action predictor. This shift makes long-horizon reasoning and replanning explicit through imagined rollouts in latent space, but it also introduces additional assumptions (latent state sufficiency and reward/value modeling) and typically relies on online interaction for grounding. ContextWM further highlights that for in-the-wild video, separating static context from dynamics can improve the transferability of video-pretrained world models under appearance and environment shift. Consistent with Table~\ref{tab:direct-video-action-capabilities}, these methods differ from the demo-grounded policies above in how actions are grounded (online RL interaction rather than supervised action prediction), which in turn changes the failure modes and modeling assumptions.

\subsection{Execution and control integration}
\label{subsec:direct-control-integration}

Direct video--action methods differ not only in their training objectives, but also in how the learned predictor is inserted into the closed-loop control cycle.
In this family, video-derived temporal structure is usually not preserved as an explicit target at deployment.
Instead, temporal prediction mainly shapes the policy representation during training, while the deployed system outputs actions directly in the robot's native control space.
Boundary latent-state world models retain internal rollouts for planning, but these rollouts are still not exposed as inspectable visual targets or constraint objects in the same way as the explicit-interface methods discussed later.

Execution mode therefore becomes a central design choice.
As summarized in Table~\ref{tab:direct-video-action-comparison}, GR-1 predicts actions step by step; GR-2, UWM, UVA, and Fast-WAM use short action chunks; Cosmos Policy either executes denoised action chunks directly or selects among imagined candidates with a value-guided planner; PAD replans in a receding-horizon loop; VidMan and VPP run lightweight action heads on predictive video features; and APV/ContextWM plan through latent rollouts.
These deployment patterns do not align one-to-one with model architecture: diffusion models may be used for receding-horizon control, chunked execution, or feature-conditioned action prediction, while autoregressive models may also move from stepwise to chunked decoding.
The resulting trade-off is therefore governed by the observation-update rate, the length of the open-loop action segment, and the computation spent on future prediction before each control update.

This deployment view also clarifies the main limitation of the direct family.
Because actions are produced without an inspectable intermediate, there is no natural checkpoint at which the robot can verify reachability, contact consistency, collision risk, or dynamic feasibility before moving.
Classical safeguards such as workspace filtering, collision checking, or constraint projection can still be added around the learned policy, but they are not native to the video-to-action interface itself.
As a result, failures are often easy to observe but difficult to localize: an incorrect action may originate from weak temporal prediction, missing embodiment cues in frozen video features, rollout drift, action decoding error, or an execution horizon that is too open-loop for the task.
Boundary world-model methods offer somewhat more internal structure through latent rollouts, but they shift the burden to latent-state sufficiency, model bias, and reward/value learning.
Compared with the latent-action and explicit-interface families, direct video--action policies therefore trade modular verification and intervention for a simpler, more tightly coupled route from video-supervised representations to executable control.

\subsection{Summary and takeaways}
Direct video--action methods use temporal visual prediction to shape representations that are later grounded to robot control.
Across the three coupling points in Fig.~\ref{fig:direct_video_action_pipeline} and Table~\ref{tab:direct-video-action-comparison}, the key distinction is where the video-derived temporal structure enters the control stack: joint generators co-train video prediction and action generation in one backbone, frozen-feature policies freeze a predictive video model and train a lightweight action head on its internal features, and boundary world-model methods use action-free video to initialize latent dynamics for model-based RL.
This makes the family scalable to action-free video and heterogeneous robot data, but it also keeps the video-to-action link largely implicit.

\paragraph{How temporal prediction transfers to control.}
The main pattern is that video prediction often matters more during training than during deployment: many direct policies can bypass explicit future-video generation at test time, with the future-video objective serving mainly to encode motion, contact, and interaction outcomes into the policy representation rather than to provide an explicit plan for the controller.
Different methods implement this transfer through different coupling mechanisms, including shared token spaces, joint diffusion with masking, modality-specific denoising queries, decoupled video/action heads, and latent-frame interfaces for non-image quantities.
This creates a recurring tension: tighter joint generators keep temporal cues close to action prediction, whereas more factorized variants improve efficiency and modularity but depend more heavily on whether the learned or frozen representation preserves control-relevant dynamics.

\paragraph{Empirical evidence and evaluation fragmentation.}
Within-family comparisons are most meaningful when methods share a benchmark and protocol (e.g., parts of the CALVIN~\citep{CALVIN_2022}, MetaWorld~\citep{yu2020meta_world}, or LIBERO~\citep{LIBERO_Liu_2023} evidence), whereas cross-suite comparisons remain suggestive because reported gains may also reflect differences in data scale, embodiment coverage, or task distribution.

\paragraph{Design patterns and failure modes.}
Because direct methods do not expose an inspectable intermediate target, execution mode (Section~\ref{subsec:direct-control-integration}) becomes the main deployment-level control lever.
These execution choices in turn shape failure modes: joint generators may suffer from compounding temporal or denoising error, frozen-feature policies may inherit representation mismatch from the frozen predictor's features, and latent-state boundary cases depend on latent sufficiency and reward/value learning.
In all cases, as detailed in Section~\ref{subsec:direct-control-integration}, physically infeasible or out-of-distribution actions are hard to intercept before execution because the interface exposes no separate control object for feasibility.

\paragraph{Relation to intermediate interfaces.}
The practical advantage of direct video--action policies is their simple route from video-supervised representations to executable actions.
The cost is that the relation between ``what changes'' in video and ``what should be executed'' remains embedded in model parameters rather than exposed as a controllable interface.
The next two families introduce more structured intermediates between video-derived temporal structure and robot control: latent-action methods learn compact action-like variables from transitions, while explicit visual interface methods expose interpretable predictive targets such as trajectories, subgoals, or video plans to downstream controllers.
We return to cross-cutting issues---including controllability, temporal abstraction, and grounding protocols---in Section~\ref{sec:discussion_future}.

\section{Latent-Action Interfaces for Control}
\label{sec:latent_actions}
Latent-action methods introduce an intermediate \emph{action abstraction} learned from how observations change over time, and then use a comparatively small amount of action-labeled robot data to connect this abstraction to executable commands.
Unlike direct video--action policies (Section~\ref{sec:direct-video-action}), which keep the connection between visual change and control implicit inside a policy network, latent-action methods introduce a structured intermediate variable intended to represent the \emph{cause} of an observed transition.
It serves as a compact interface that can support planning, model-predictive control, or policy learning in an abstract action space before being grounded to a specific robot.
This action-like abstraction differs from the \emph{latent state} of an action-free world model: a latent action summarizes \emph{what changed} between observations and is meant to be selected or grounded as a control primitive, whereas a latent world-model state represents the environment for downstream action-conditioned planning or policy learning.

\begin{table*}[t]
  \caption{\textbf{Discovery and grounding workflows of latent-action methods.}
  ``Discovery Data'' indicates the action-free video used to learn latent actions.
  ``Grounding Workflow'' describes the procedure used to establish the latent-to-action or latent-conditioned policy connection.
  ``Grounding Data'' specifies the action-labeled supervision or interaction required to connect latents to executable controls.
  ``Task Scope'' summarizes the broad task family reported in the original papers.
  ``Real-Robot Scope'' summarizes the real-robot experiments reported (or their absence).}
  \label{tab:latent-action-capabilities}
  \centering
  \footnotesize
  \setlength{\tabcolsep}{6pt}
  \renewcommand{\arraystretch}{1.15}
  \begin{tabularx}{\textwidth}{
    >{\raggedright\arraybackslash}p{1.6cm}
    >{\raggedright\arraybackslash}p{3.1cm}
    >{\raggedright\arraybackslash}X
    >{\raggedright\arraybackslash}p{1.7cm}
    >{\raggedright\arraybackslash}p{2.0cm}
    >{\centering\arraybackslash}p{2.0cm}
  }
    \toprule
    \textbf{Method} &
    \textbf{Discovery Data} &
    \textbf{Grounding Workflow} &
    \textbf{Grounding Data} &
    \textbf{Task Scope} &
    \textbf{Real-Robot Scope} \\
    \midrule

    \rowcolor{gray!8}
    \multicolumn{6}{l}{\textit{Latent actions as standalone control interfaces}} \\
    \addlinespace[3pt]

    CLASP~\citep{CLASP_ICLR2019}
      & Robot video (BAIR pushing)
      & Small latent$\rightarrow$action map; plan over latents (image-goal MPC)
      & Small labeled set
      & Tabletop manip. 
      & \textit{--} \\

    \addlinespace[5pt]
    \midrule
    \rowcolor{gray!8}
    \multicolumn{6}{l}{\textit{Latent actions in instruction-conditioned policy learning}} \\
    \addlinespace[3pt]

    LAPA~\citep{LAPA_ye2025}
      & Robot + human manip. video
      & Latent-action prediction as pretraining; head replacement + fine-tune
      & Robot demos
      & Tabletop manip.
      & 3 Franka tabletop tasks, 15 objects \\

    Moto~\citep{Moto_chen2025}
      & Cross-embodiment robot + human video
      & Autoregressive latent-token pretraining; co-train action head
      & Robot demos
      & Long-horizon manip.
      & 3 FANUC tabletop tasks \\ 


    UniVLA~\citep{UniVLA_RSS_25}
      & Cross-embodiment robot + human video
      & Task-centric latent tokens; lightweight decoder $\rightarrow$ actions
      & Small robot demos
      & Manip., navigation
      & 4 Piper-arm long-horizon tasks \\

    ConLA~\citep{ConLA_dai2026}
      & Human manip. video + action-category labels
      & Contrastive latent-token discovery; head replacement + fine-tune
      & Action-labeled robot demos
      & Tabletop manip.
      & 3 Franka tabletop tasks \\

    HiLAM~\citep{HiLAM_kim2026}
      & Manip. video; latent actions inferred by a pretrained model
      & Chunk latent actions into skills; skill-conditioned action policy
      & Robot demos
      & Long-horizon manip.
      & \textit{--} \\

    RotVLA~\citep{RotVLA_li2026}
      & Robot + human manip. video
      & Predict rotational latent actions; co-train joint flow-matching latent+action
      & Robot demos
      & General and bimanual manip.
      & 3 ARX R5 tasks, incl. dual-arm \\

    ALAM~\citep{ALAM_tang2026}
      & Action-free video frame triplets
      & Additive/reversal-consistent latent transitions; co-generate latent+action
      & Robot demos
      & Tabletop manip.
      & 4 Piper tabletop tasks \\
    \bottomrule
  \end{tabularx}

  \vspace{3pt}
  \raggedright
  \footnotesize
  ``Small labeled set'' is relative to the discovery video scale.
  \textit{manip.} = manipulation; \textit{--} = no real-robot execution evaluation reported.
\end{table*}

The motivation is that videos of physical interaction---human or robotic---contain structured, action-like information: the observed transition often constrains what interaction produced it, even when the underlying control commands are unobserved.
Latent-action methods operationalize this by separating \emph{discovery}---learning the abstraction from action-free transitions---from \emph{grounding}---mapping it to robot controls with limited labeled data.
The central question is therefore: \emph{Can we discover an action-like abstraction purely from observation---one that supports planning and search---and then ground it to a specific robot's control space with minimal supervision?}

\paragraph{Core decomposition.}
Latent-action methods answer this by separating three roles that are conflated in direct models.
First, a \emph{transition dynamics model} is learned from action-free video to explain observation transitions through a capacity-limited latent variable (continuous or discrete), encouraging it to capture \emph{what changed} rather than static scene content; we refer to this transition latent as a \emph{latent action}.
Second, a \emph{latent policy or planner} operates in this latent-action space, selecting latent codes from the current observation, optionally conditioned on a goal image and/or a language instruction.
Third, a lightweight grounding module connects these latent codes to executable actions using modest action-labeled supervision (e.g., a decoder, a co-occurrence dictionary/adapter, or head replacement in instruction-conditioned policies).
These components are typically learned in stages: latent-action discovery from observation-only transitions, followed by grounding and control learning with action-labeled data or interaction.
This factorization treats the latent action as a dedicated interface between perception and control: discovery can exploit large action-free corpora, while grounding remains robot- and embodiment-specific.
In practice, discovery data ranges from small robot datasets (CLASP) to large-scale observation corpora in non-robot domains (e.g., games) and diverse cross-embodiment manipulation datasets (UniVLA), whereas grounding data is consistently limited across methods (Table~\ref{tab:latent-action-capabilities}), illustrating how this design decouples discovery scale from robot-action supervision requirements. 

\begin{figure*}
    \centering
    \includegraphics[width=0.9\linewidth, trim= 480 200 480 200, clip]{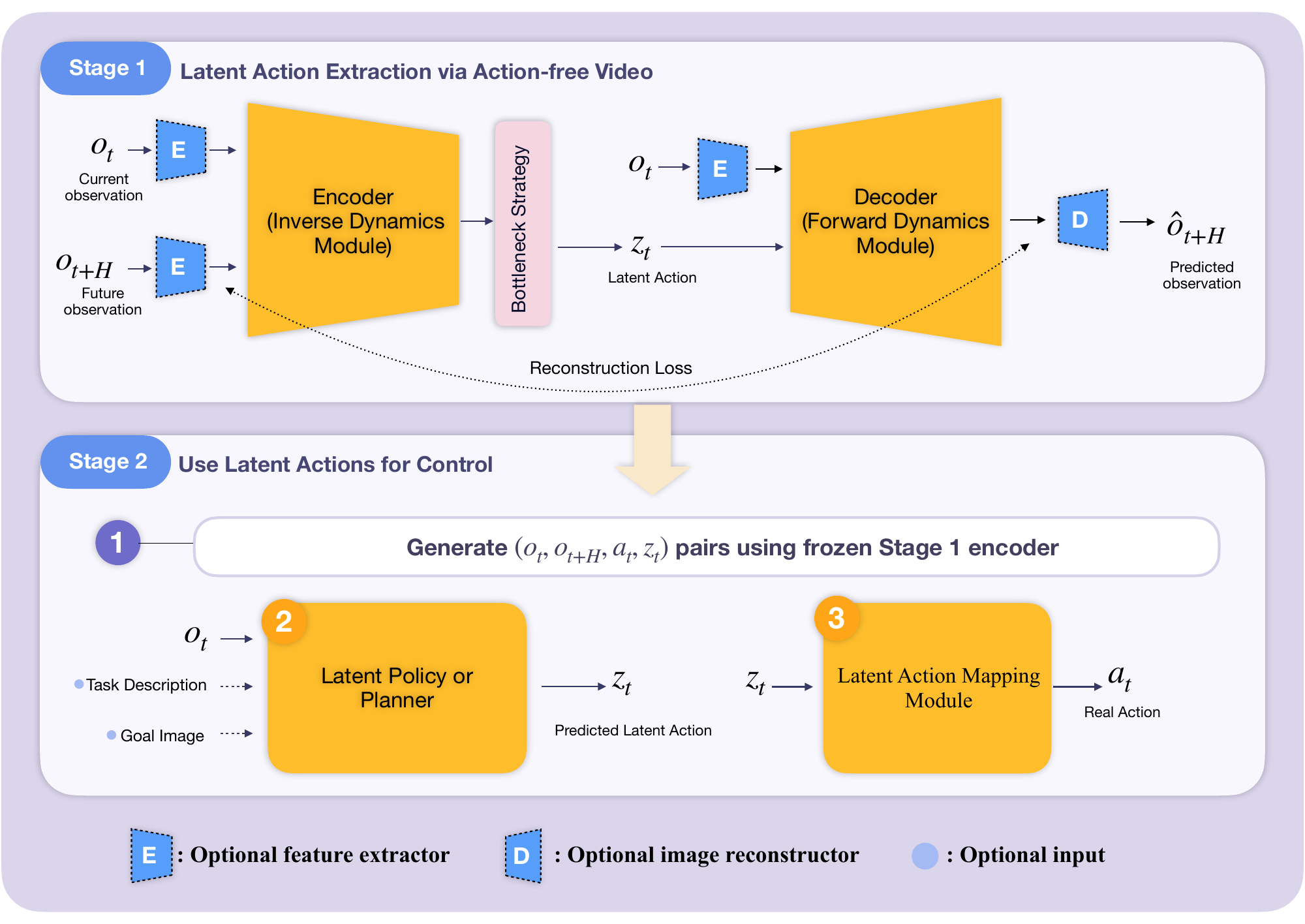}
    \vspace{-3mm}
    \caption{\textbf{Generic latent-action pipeline.}
    Latent-action methods typically follow a two-stage logic.
    \textbf{Stage~1}: action-free video is used to infer a latent action $z_t$ from an observed transition $(o_t,o_{t+H})$ through an inverse-dynamics-style encoder, constrain it with a bottleneck, and train a forward model to predict the future observation or representation.
    \textbf{Stage~2}: the learned latent-action interface is connected to control using action-labeled robot data: a frozen encoder can label transitions with $z_t$, a planner or policy can select latent actions from the current observation and task context, and a grounding mechanism maps or couples $z_t$ to executable robot actions $a_t$.
    The grounding block is drawn as a mapping module for clarity, but may also be implemented through head replacement, persistent latent-token decoding, or joint latent--action generation.}
    \label{fig:latent_action_pipeline}
    \vspace{-2mm}
\end{figure*}

\paragraph{Organization.}
We first summarize common building blocks (bottleneck mechanisms and the inverse/forward dynamics factorization) and a generic two-stage latent-action pipeline (Sections~\ref{subsec:latent_prelims}--\ref{subsec:latent_framework}).
We then review representative methods in two groups, distinguished by how the latent action enters control.
The first group treats latent actions as \emph{standalone control interfaces} (Section~\ref{subsec:standalone_latent}): a planner or policy selects latent actions, and a separate grounding mechanism maps them to commands.
Within this group, we first discuss non-manipulation precursors that establish the interface template, followed by CLASP as the manipulation-oriented representative.
The second group embeds latent actions \emph{inside instruction-conditioned policies} (Section~\ref{subsec:lapa_univla}), where they act as pretraining targets, auxiliary objectives, persistent action tokens, co-generated streams, or hierarchical skills.
Tables~\ref{tab:latent-action-comparison} and~\ref{tab:latent-action-capabilities} summarize the methods, data sources, and grounding mechanisms.

\subsection{Building blocks}
\label{subsec:latent_prelims}
Compared to direct video--action policies, latent-action methods more often introduce dedicated latent-variable machinery (e.g., bottlenecks and discrete codebooks). We briefly review these recurring components before surveying specific methods.

We continue to write $o_t$ for the observation at time $t$ (typically an image or a short observation window), $a_t$ for the true robot action when available, and $z_t$ for a latent action.
We use ``encoder'' to mean a network that infers $z_t$ from an observed transition, and ``decoder'' (or predictor) to mean a network that predicts the future observation (or its representation) conditioned on $z_t$ and current observation $o_t$.

\paragraph{Latent actions as inverse/forward dynamics in a chosen space.}
A common modeling choice is to treat latent-action discovery as a factorization of inverse and forward dynamics:
\begin{equation}
    z_t \sim q_\phi(\,\cdot \mid o_t, o_{t+H}), 
    \qquad
    \hat{o}_{t+H} \sim p_\theta(\,\cdot \mid o_t, z_t),
\end{equation}
where $H$ is typically 1 (next frame) or a fixed prediction horizon.
Interpreting $z_t$ as an ``action'' makes the encoder $q_\phi$ functionally a \emph{latent inverse dynamics model} (infer the cause of a transition), and the decoder $p_\theta$ a \emph{latent forward model} (predict the effect of applying $z_t$). To learn a meaningful latent action, a reconstruction objective between $o_{t+H}$ and $\hat{o}_{t+H}$ is adopted with bottleneck constraints, encouraging the latent to capture the change between observations.
Importantly, different papers instantiate $p_\theta$ in different spaces: it may predict pixels, pixel differences, learned visual features, or a latent state used by a world model. This choice affects robustness, scalability, and what the learned latent captures.

\paragraph{Bottleneck mechanisms.}
Many latent-action methods encourage $z_t$ to be action-like by limiting how much transition information it can carry.
This capacity constraint can be imposed in several forms, not only as a binary continuous-versus-discrete choice.
Continuous bottlenecks, such as $\beta$-VAE / information-bottleneck objectives~\citep{InformationBottleneck_2017, VAEs_Kingma_2019}, penalize the information carried by a stochastic latent so that $z_t$ retains what is needed to predict the transition while suppressing static scene content.
Hard VQ codebooks~\citep{VQVAE_NIPS_2017} impose a different bottleneck by mapping a continuous encoder output to a discrete code from a finite vocabulary, yielding \emph{action tokens} that can support dictionary-style grounding and language-model-compatible representations.
Soft codebooks, such as SoftVQ~\citep{SoftVQVAE_chen2025}, interpolate between these cases: they retain a finite set of prototypes but represent the latent as a weighted combination of codebook entries, so the downstream latent can vary continuously.
These bottlenecks bias $z_t$ toward compact transition descriptors, but action-likeness also depends on the prediction space, auxiliary constraints, and the grounding procedure.

\subsection{A generic latent-action pipeline}
\label{subsec:latent_framework}
Despite varied architectures, most latent-action methods follow a two-stage logic (Figure~\ref{fig:latent_action_pipeline}): latent actions are first discovered from observation-only transitions, and are then used for planning, policy pretraining, or robot-action generation. In early methods these stages are often trained independently, whereas recent instruction-conditioned methods may co-train the latent and robot-action streams during fine-tuning.

\paragraph{Stage 1: discover latent actions from action-free video.}
Using action-free videos, a model is trained to explain video transition dynamics through a bottleneck latent action $z_t$.
The encoder infers $z_t$ from observed change, and the decoder or predictor reconstructs the future observation or its representation from the current observation and $z_t$.
Design choices in this stage include:
(i) prediction target (pixels vs.\ features vs.\ latent states);
(ii) bottleneck type (continuous bottleneck vs.\ codebook-based, including hard and soft variants);
and (iii) auxiliary constraints (e.g., composability, cycle consistency, contrastive separation, or algebraic consistency) that bias $z_t$ toward reusable primitives rather than arbitrary transition hashes.

\paragraph{Stage 2: use latent actions for control.}
At test time, the agent must choose actions without access to $o_{t+H}$.
Thus, methods add one or more of the following mechanisms:
\begin{itemize}
    \item \textbf{Latent-action selection:} a latent policy or planner that proposes $z_t$ from the current observation (optionally conditioned on language or a goal), trained for example by imitation of inferred latents or by planning through the learned forward model.
    \item \textbf{Grounding to robot controls:} a mapping between $z_t$ and real robot actions $a_t$, learned from action-labeled trajectories (e.g., a decoder $z_t \!\rightarrow\! a_t$, a code-to-action dictionary, an action head replacing a latent head, or a single generative policy module that jointly produces latent and robot-action streams).
\end{itemize}

The key distinction from direct methods is that the latent variable is treated as a dedicated intermediate interface: it is learned from videos first, then connected to executable controls through decoding, policy fine-tuning, persistent latent-token grounding, or joint latent--action generation.
Figure~\ref{fig:latent_action_pipeline} summarizes this two-stage pipeline.

\begin{figure}
    \centering
    \includegraphics[width=1\linewidth, trim=420 100 420 90, clip]{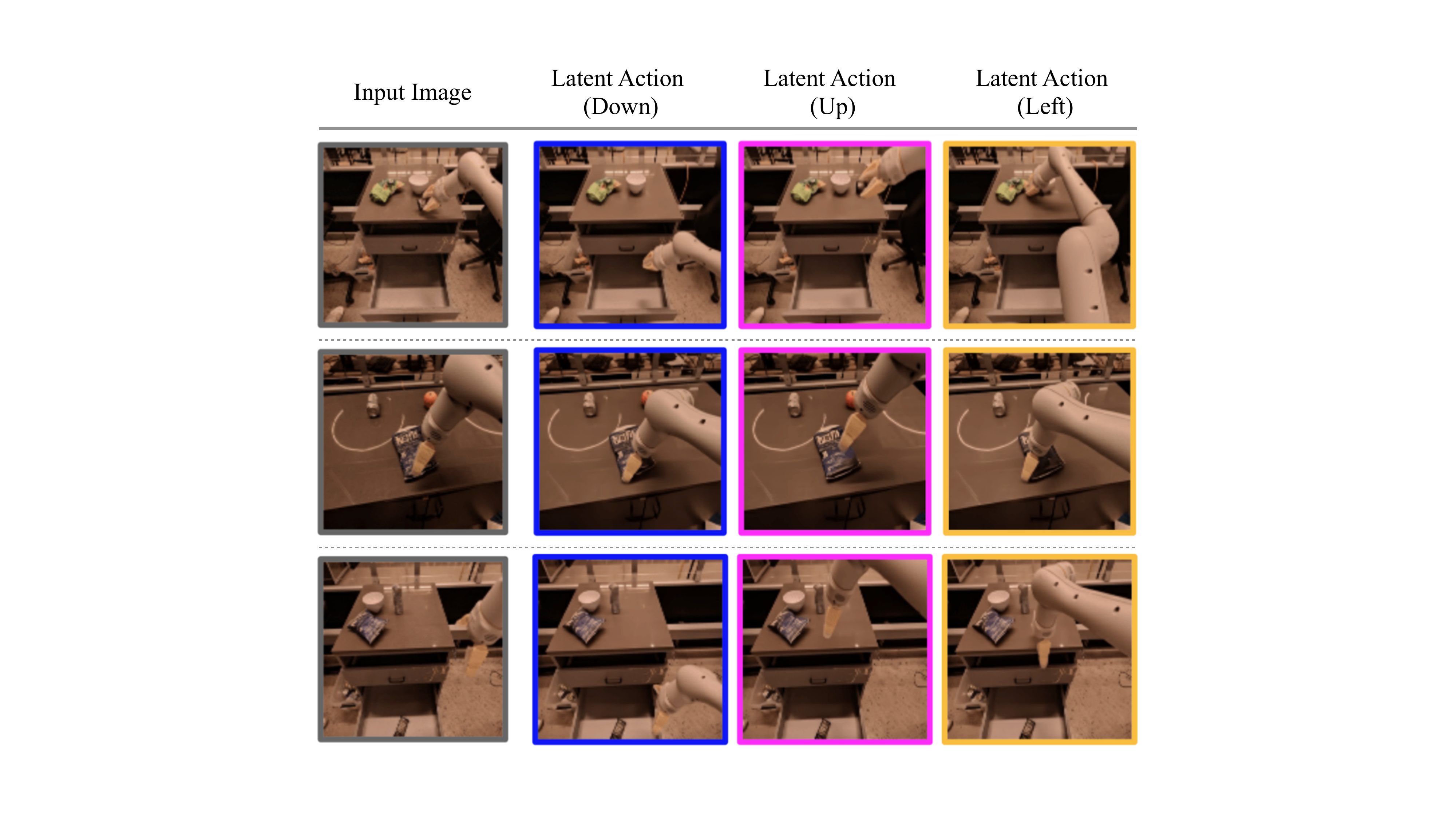}
    \caption{\textbf{Latent actions induce action-relevant visual change (example adapted from Genie~\citep{Genie_1_ICML_2024}).} Each row is a different scene. The first column is the input frame, and columns~2--4 show future observations predicted by applying different latent actions to it. Although unlabeled, the latent actions produce consistent, interpretable effects (e.g., end-effector or object motion in different directions), illustrating an emergent alignment between the learned latent-action space and robot-action-relevant visual change.}
    \label{fig:latent_action_example}
\end{figure}

Figure~\ref{fig:latent_action_example} provides a qualitative visualization of this idea: latent actions learned without action labels can induce stable, semantically consistent visual changes in robotic scenes, such as different end-effector motion directions. This does not by itself establish executable control, but it illustrates why such codes are plausible candidates for later grounding.

\subsection{Latent actions as standalone control interfaces}
\label{subsec:standalone_latent}
A first line of work treats the latent action as an explicit control object: a planner or policy selects latent actions from observations, and a separate grounding mechanism maps them to executable commands.
This contrasts with the instruction-conditioned methods in the next subsection, where latent actions are usually embedded inside a larger policy-learning pipeline rather than serving as the sole control interface.

\paragraph{Precursors from non-manipulation control domains.}
Several non-manipulation methods established templates that later manipulation work adapted.
ILPO~\citep{ILPO_edwards2019_ICML} introduced discrete latent actions for imitation from state-only demonstrations: observed transitions are labeled with latent variables, a policy is trained over these latents, and a separate grounding step connects them to the true action space. It also surfaces a recurring failure mode for discrete bottlenecks learned from observation only---maximum-likelihood latent labeling can become degenerate, using only a few codes---so latent actions are not automatically identifiable as independently controllable primitives.

FICC~\citep{FICC_ICLR_2023} learns a finite set of VQ latent actions from observation-only experience in a learned feature/state space, then aligns these codes to real actions through a lightweight co-occurrence \emph{action adapter} built from interaction data, allowing a pretrained latent-dynamics model to be reused for sample-efficient model-based RL in game domains. Its relevance here is the grounding template: associating discrete latent codes with executable actions through co-occurrence statistics, a pattern that manipulation methods later adapt to ground video-derived latents to robot actions.

LAPO~\citep{LAPO_schmidt2024_ICLR} learns a discrete latent-action vocabulary from action-free video with a VQ bottleneck---an inverse model infers a code from consecutive observations and a forward model predicts the next observation---then labels the data with inferred codes and trains a latent policy by imitation, grounded through a small decoder or online reward-driven fine-tuning (evaluated on Procgen). Its relevance is the use of inferred latent codes as imitation supervision for a latent policy, a pattern later adapted as a latent-code pretraining target for instruction-conditioned policies in LAPA and Moto.

Genie~\citep{Genie_1_ICML_2024} demonstrates unsupervised latent-action discovery at scale.
It learns discrete VQ latent actions from large action-free video corpora, with an additional demonstration on robotic videos such as RT-1~\citep{RT1_Brohan_2022}, and conditions a video generator on these codes to produce controllable futures.
This suggests that observation-only video can support semantically consistent latent-action vocabularies, but Genie remains a controllable world-model / video-generation method rather than a robotic manipulation policy: it does not ground the discovered codes to executable robot commands or evaluate robotic manipulation control.
We therefore treat it as a non-manipulation precursor, alongside FICC and LAPO.

\begin{table*}[t]
  \caption{\textbf{Structural comparison of latent-action methods.} ``Latent'' and ``Bottleneck'' describe the form and capacity constraint of the learned latent-action variable. ``Prediction Space'' indicates the observation or representation space used by the latent-action discovery model. ``Grounding mechanism'' describes the architectural connection between the latent interface and executable actions. ``Deployment role'' indicates whether latent actions persist as a test-time control interface or serve only as training-time supervision.}
  \label{tab:latent-action-comparison}
  \centering
  \footnotesize
  \setlength{\tabcolsep}{4pt}
  \renewcommand{\arraystretch}{1.15}

  \begin{tabularx}{\textwidth}{
      @{}
      >{\raggedright\arraybackslash}p{1.8cm}
      >{\centering\arraybackslash}p{1.2cm}
      >{\centering\arraybackslash}p{2.2cm}
      >{\raggedright\arraybackslash}p{3.8cm}
      Y
      >{\raggedright\arraybackslash}p{1.9cm}
      @{}
    }
    \toprule
    \textbf{Method} & 
    \textbf{Latent} & 
    \textbf{Bottleneck} & 
    \textbf{Prediction Space} & 
    \textbf{Grounding Mechanism} & 
    \textbf{Deployment Role} \\
    \midrule
    \rowcolor{gray!8}
    \multicolumn{6}{l}{\textit{Latent actions as standalone control interfaces}} \\
    \addlinespace[2pt]
    CLASP~\citep{CLASP_ICLR2019}
      & Continuous
      & $\beta$-VAE / IB
      & Pixels (next-frame / rollout)
      & MLP latent$\rightarrow$action (labeled rollouts)
      & Used for planning (MPC) \\
    \addlinespace[4pt]
    \midrule
    \rowcolor{gray!8}
    \multicolumn{6}{l}{\textit{Latent actions in instruction-conditioned policy learning}} \\
    \addlinespace[2pt]
    LAPA~\citep{LAPA_ye2025}
      & Discrete
      & VQ
      & Pixels / features (fixed-horizon)
      & Replace latent head with action head; fine-tune
      & Pretraining only \\
    Moto~\citep{Moto_chen2025}
      & Discrete
      & VQ
      & Pixels / features (fixed-horizon)
      & Add action head; co-train with autoregressive latent action prediction
      & Pretraining + co-training \\
    UniVLA~\citep{UniVLA_RSS_25}
      & Discrete
      & Two-stage VQ
      & DINO~\citep{Dinov2_2023} features (short-horizon)
      & Lightweight latent$\rightarrow$action decoder
      & Action tokens \\
    ConLA~\citep{ConLA_dai2026}
      & Discrete
      & Contrastive VQ
      & Pixels (future-frame)
      & Replace latent head with action head; fine-tune
      & Pretraining only \\
    HiLAM~\citep{HiLAM_kim2026}
      & Discrete
      & VQ
      & Inherited from pretrained latent-action model
      & Chunk latents into skills; skill-conditioned policy with short-horizon head adapted to robot actions
      & Latent skill conditioning \\
    RotVLA~\citep{RotVLA_li2026}
      & Continuous
      & SoftVQ on $SO(n)$
      & Pixels / features (triplet consistency)
      & Joint flow-matching (latent action + robot actions)
      & Latent action conditioning \\
    ALAM~\citep{ALAM_tang2026}
      & Continuous
      & Algebraic consistency
      & Pixels / features (triplet)
      & Joint flow-matching; co-generate latent action + robot actions
      & Latent action conditioning \\
    \bottomrule
  \end{tabularx}
  
  \vspace{3pt}
  \raggedright
  \footnotesize
  \textit{Two-stage VQ} and \textit{contrastive VQ} denote UniVLA's staged codebook learning and ConLA's contrastive action--vision disentanglement, respectively.
\end{table*}

\subsubsection{CLASP: minimal and composable latent actions.}
\label{subsec:clasp}
CLASP~\citep{CLASP_ICLR2019} instantiates the generic latent-action pipeline in Fig.~\ref{fig:latent_action_pipeline}: it discovers continuous transition latents from action-free video and then grounds them to robot actions for image-goal planning.
Its key contribution is to add two biases that later reappear throughout the latent-action family: the latent should be \emph{minimal}, carrying only the information needed to explain visual change, and \emph{composable}, so that short-horizon latents can combine into longer-horizon behavior.

In the discovery stage, CLASP follows the inverse/forward factorization introduced above.
A latent inverse model infers the latent action from an observed transition, and a latent forward model predicts the future observation from the current observation and the inferred latent.
Minimality is encouraged by limiting the information carried by the latent, while composability is encouraged by a learned composer network that combines consecutive latent actions into a latent representing the longer transition.
Together, these choices produce a continuous latent-action space that supports planning before grounding to real robot commands.

For control and deployment, CLASP searches over latent-action sequences whose predicted futures best match a goal image.
The selected latent actions are then mapped to real robot actions by a lightweight grounding module learned from a small action-labeled dataset, and execution proceeds in a receding-horizon loop.
The trade-off is that the latent is not guaranteed to be uniquely controllable, and planning can be sensitive to forward-model error and sampling cost.

\paragraph{Takeaways for standalone latent-action interfaces.}
CLASP illustrates the standalone use of latent actions as explicit control objects in manipulation: continuous latents are discovered from action-free video, planned over with an MPC-style objective, and grounded to executable commands with limited action-labeled data.
This route offers a modular separation between latent discovery, latent-space planning, and robot-specific grounding, but it also exposes the main weaknesses of standalone latent control: sensitivity to forward-model error, sampling cost, and ambiguity in whether a visually predictive latent corresponds to a uniquely executable action.
The discrete counterpart to this template appears mainly in non-manipulation precursors: FICC and LAPO ground discrete codes through a co-occurrence dictionary or small decoder, while Genie uses them to drive a controllable world model rather than to issue robot commands.
These limitations help explain why recent manipulation work has largely shifted toward the instruction-conditioned designs below, where latent actions are embedded into broader policy-learning pipelines and the central question becomes how deeply the latent signal persists into deployment.

\subsection{Latent actions in instruction-conditioned policy learning}
\label{subsec:lapa_univla}

Latent actions have been incorporated into instruction-conditioned policies (often termed VLA policies in recent work) as video-derived, action-centric signals for manipulation. Methods in this group still retain a latent-action discovery stage: an action-free transition model extracts codes or continuous latents from video, and the resulting latent labels are then used to supervise, condition, or co-train a downstream policy. The main difference from the standalone interfaces above is that the latent action is usually no longer the sole control object at deployment; it may be discarded after pretraining, retained as an auxiliary training signal, decoded as a persistent action token, used as hierarchical conditioning, or generated alongside robot actions.

This subsection follows two recurring design directions. One line improves \emph{latent-action quality}: it asks whether transition-derived latents capture controllable, task-relevant change rather than appearance, camera motion, or other nuisance dynamics, and whether their local structure supports useful composition. The other line improves \emph{policy integration and grounding}: it asks how the discovered latents should be used once a policy is trained for executable robot control, ranging from head replacement and co-training to persistent token decoding, hierarchical conditioning, and joint latent--action generation.

\subsubsection{LAPA: latent action pretraining for instruction-conditioned policies.}
\label{subsec:lapa}

LAPA~\citep{LAPA_ye2025} uses discrete latent actions primarily as a \emph{pretraining target} for instruction-conditioned policies, rather than as a deployment-time control interface. The goal is to transfer temporal, action-relevant structure from videos into a VLA backbone before learning real robot actions.

Like Genie, LAPA first learns a VQ-style latent-action model from videos: a transition encoder produces a discrete code summarizing the change between observations at a fixed horizon, and a decoder predicts the future observation from the current observation and this code. Rather than building a dictionary or training a latent policy, LAPA uses these inferred codes as pretraining supervision: given an observation and language instruction, the VLA model is trained to predict the corresponding latent-action code (obtained by running the frozen transition encoder on video pairs), providing an action-centric learning signal without requiring action labels during this stage.

To produce executable policies, LAPA replaces the latent-code head with a real-action head and fine-tunes on a comparatively small action-labeled robot dataset. At deployment, latent actions play no explicit role---the policy outputs real robot actions directly. A limitation is that latent-action prediction is not itself the final control interface: swapping to a real-action head can introduce a representation gap between the pretraining objective and the target action space, which may require additional adaptation during fine-tuning.

\subsubsection{Moto: autoregressive modeling of latent action sequences.}
\label{subsec:moto}

Moto~\citep{Moto_chen2025}, like LAPA~\citep{LAPA_ye2025}, does not use discrete latent actions as the final executable interface, but extends the template from per-transition code prediction to sequence-level prediction. After learning a VQ-style latent-action model from action-free video, Moto labels videos with latent-action codes and trains a GPT-style policy to autoregressively predict latent action sequences over a clip, conditioned on language, visual observation, and previously generated codes. This shifts the role of the latent action from a single-step pseudo-action label to a temporally ordered code, allowing action-free video to supervise dependencies across successive transitions.

For executable control, Moto adds an action head alongside the latent-action prediction head, so the latent-action objective continues to shape the backbone during fine-tuning rather than being discarded. Deployment behavior nonetheless remains close to LAPA: the policy outputs real robot actions directly, so the latent-action interface lives inside training rather than at deployment. Moto's addition to the latent-action family is therefore autoregressive latent-action prediction as a sequential pretraining and co-training objective. A remaining concern, shared with other VQ-style latent-action methods, is that transition-based latent-action discovery does not by itself separate action-caused changes from scene- or camera-induced changes; as a result, the discrete vocabulary may encode controllable and task-irrelevant dynamics in the same latent space.

LAPA and Moto use transition-derived latent codes mainly as policy-learning signals, focusing on how those codes supervise or co-train the policy rather than substantially changing the latent-action discovery objective. The methods that follow instead target the \emph{quality} of the latent-action space itself---whether the discovered latents capture controllable, task-relevant change rather than appearance or other nuisance dynamics, and whether their local structure supports useful composition.

\subsubsection{UniVLA: task-centric latent actions.}
\label{subsec:univla}

UniVLA~\citep{UniVLA_RSS_25} targets cross-embodiment instruction-conditioned policy learning by addressing a recurring failure mode of latent-action discovery: transition codes can entangle controllable task progress with camera motion, other agents, background changes, or scene clutter.
The main change is to make the latent-action vocabulary task-centric, so that the policy predicts tokens associated with task-relevant change rather than arbitrary visual difference.

UniVLA learns latent actions in a learned feature space rather than pixels, using DINOv2~\citep{Dinov2_2023} patch features as both inputs and prediction targets.
Its key mechanism is a language-guided factorization of transition information.
With language conditioning, one latent branch is encouraged to absorb task-irrelevant variation needed for reconstruction, such as camera or background change.
A second task-centric codebook is then trained to explain the remaining transition structure without language conditioning in the discovery model.
This gives the downstream policy a latent vocabulary intended to track task progress rather than generic visual change.

The task-centric codes are used as persistent action tokens for policy learning: videos are labeled with these tokens, an autoregressive vision--language policy predicts them from observation and instruction, and a lightweight action decoder grounds each predicted token into the robot's native control space.
Unlike LAPA, which discards the latent-action head after pretraining, UniVLA retains the latent vocabulary as the policy's output interface at deployment.

The trade-off is that task-relevant separation remains inherently approximate: UniVLA mitigates nuisance dynamics by factorizing transition information, but the retained task-centric tokens may still absorb viewpoint, embodiment, or scene correlations that are predictive in the discovery data yet brittle during grounding and execution.

\subsubsection{ConLA: contrastive latent action learning.}
\label{subsec:conla}

ConLA~\citep{ConLA_dai2026} shares UniVLA's concern that transition-derived codes can absorb nuisance visual variation, but frames the problem as action--appearance separation rather than task-centric factorization.
The goal is to make the latent-action vocabulary capture controllable motion dynamics instead of appearance, background, or future-content correlations that help reconstruct the next frame but do not define an action.

ConLA introduces contrastive constraints during latent-action discovery so that the action branch is pulled toward motion-related distinctions and pushed away from appearance-related content before quantization.
Action-category priors and temporal-order cues provide the supervision signal for this separation, after which the learned codebook is used for reconstruction and policy pretraining.
This keeps the method within the LAPA-style template, but changes the discovery objective so that the pretraining labels are more action-centric.

For executable control, ConLA follows LAPA's head-based pattern: latent-action codes serve as a pretraining target for an instruction-conditioned policy, and a real-action head is fine-tuned on action-labeled robot demonstrations.
At deployment, latent actions play no explicit role.
The trade-off is that the action--appearance separation is not fully unsupervised; it depends on action-category labels or pseudo-labels extracted from language, so the granularity and reliability of those categories can shape what distinctions the latent vocabulary learns.

\subsubsection{RotVLA: rotational structure for compositional latent actions.}
\label{subsec:rotvla}

RotVLA~\citep{RotVLA_li2026} revisits two desiderata that appeared earlier in CLASP---continuous latent actions and compositional structure---but gives them a rotational form for instruction-conditioned policy learning.

The method keeps a codebook-style bottleneck but uses SoftVQ~\citep{SoftVQVAE_chen2025} so that each latent action is represented as a soft, continuous combination of prototypes.
It then constrains these continuous latents to follow a rotational composition rule on $SO(n)$: two adjacent latent transitions should compose into a latent that explains the corresponding longer transition, with composition given by a fixed group operation rather than the learned composer network used in CLASP.
This local constraint biases the latent space toward reusable action increments rather than arbitrary next-frame descriptors, without requiring the latent action to be a separately executed control primitive.

For downstream control, RotVLA trains the policy to predict rotational latent actions from video--language inputs and couples this prediction pathway with robot-action generation during fine-tuning.
The latent action therefore conditions executable action generation instead of being discarded after pretraining or mapped through a separate dictionary.
The main trade-off is that rotational consistency is an imposed geometric bias: a visually consistent latent rotation need not correspond to an executable or dynamically valid composition of robot actions.

\subsubsection{ALAM: additive transition geometry for latent actions.}
\label{subsec:alam}

ALAM~\citep{ALAM_tang2026} shares RotVLA's goal of making latent actions more compositional, but uses an additive transition geometry rather than a rotational one.
It asks whether a continuous latent action can behave like a local displacement in transition space: consecutive latents should approximately add to the longer transition they span, and reversing a transition should approximately cancel the original latent.

During latent-action discovery, these algebraic constraints are applied to continuous transition latents learned from action-free video, while reconstruction keeps the latents tied to observable visual change.
The resulting representation is then used as a video-derived supervision signal for instruction-conditioned policy learning.
For executable control, ALAM co-generates latent-transition trajectories and robot-action trajectories during policy training, but only the robot-action stream is executed at deployment.

Compared with RotVLA, ALAM makes the compositional bias easier to interpret as local vector arithmetic, but it inherits the same basic limitation: local consistency over short transition tuples does not establish exact long-horizon closure or guarantee that latent addition matches the physical composition of robot actions.

The methods so far operate primarily at the short-horizon transition level. A final direction instead introduces an explicit temporal hierarchy over latent actions.

\subsubsection{HiLAM: hierarchical latent actions.}
\label{subsec:hilam}

HiLAM~\citep{HiLAM_kim2026} focuses on an aspect not emphasized by many transition-level latent-action methods: a code that explains a short visual change may be too local to represent task phase, subgoal progress, or longer-horizon intent.
It therefore uses extracted short-horizon latent-action codes as building blocks for longer-horizon latent skills, which guide instruction-conditioned task execution.

Rather than changing the basic latent-action discovery objective, HiLAM changes how discovered latents are used during policy learning.
Sequences of inferred latent actions are grouped into skills that summarize extended behavior segments without requiring manually specified boundaries.
A long-horizon policy predicts the skill from the current observation and instruction, while a short-horizon policy uses that skill condition to produce the next short-horizon control output.
This shifts the latent interface from single-transition prediction toward temporally abstract guidance for instruction-conditioned control.

For executable control, the short-horizon policy is grounded with action-labeled robot demonstrations so that it outputs native robot actions while remaining conditioned on the inferred skill.
The benefit is a clearer separation between long-horizon task structure and short-horizon control; the trade-off is that the hierarchy still depends on whether the inferred skill captures the correct task phase from the current observation and instruction, especially when visual evidence for progress is ambiguous.

\paragraph{Other extensions.}
Beyond the main instruction-conditioned designs above, several methods extend the same broad latent-action pretraining template along scale, data, or representation design. GO-1~\citep{GO1_Bu_IROS_2025} scales the Moto-style latent-action pipeline to a larger-scale data and evaluation platform. CoMo~\citep{CoMo_yang2026} explores VQ-free continuous latent-motion embeddings from internet videos using temporal-difference and contrastive objectives. We treat these as extensions rather than core taxonomy anchors because they primarily scale or vary the pretraining template rather than introducing a new grounding role within this section.

\paragraph{Takeaways for instruction-conditioned policies.}
Across instruction-conditioned latent-action methods, the recurring pattern is that discovery and grounding are tightly coupled: a latent code that predicts video transitions well may still encode nuisance dynamics, and a downstream policy cannot fully recover controllability from a poorly aligned latent space.
The main design axis is how long the latent remains active in the control pipeline---discarded after pretraining, retained as an auxiliary fine-tuning signal, or preserved more deeply as persistent tokens, co-generated streams, or long-horizon skills---a spectrum summarized in Table~\ref{tab:latent-action-comparison}.
Greater persistence makes the intermediate structure more explicit, but also carries latent-discovery errors further into control.
Accordingly, Table~\ref{tab:latent_action_quant} summarizes reported evidence and evaluation settings rather than ranking methods or isolating the effect of a particular latent design, since reported gains also vary with pretraining corpora, backbones, grounding budgets, and evaluation protocols.
Because several of the newest variants in this subsection are preprints, we read them as illustrations of emerging objective and integration designs, not as settled comparative evidence.
Taken together, these results suggest that recent objectives make latent actions more \emph{action-like} by separating task-relevant motion from nuisance appearance, imposing local compositional structure, or organizing short-horizon codes into longer-horizon skills.
These objectives are useful inductive biases rather than guarantees of causal identifiability or physical controllability, which must still be secured by the grounding mechanism and embodiment-specific control.

\begin{table*}[t]
    \centering
    \footnotesize
    \setlength{\tabcolsep}{4pt}
    \renewcommand{\arraystretch}{1.25}
    \caption{\textbf{Reported quantitative snapshots for latent-action methods.} These results are \emph{not} a leaderboard: papers differ in robot embodiments, simulation/real-robot settings, data regimes, and evaluation protocols. We use the numbers only to indicate evidence type and maturity---within-paper claims, weakly comparable shared-benchmark slices, and deployment-feasibility evidence. Entries are reported as stated in the original papers; ``Source'' gives the original location.}
    \label{tab:latent_action_quant}
    \begin{tabularx}{\textwidth}{
        @{}
        >{\raggedright\arraybackslash}p{3.2cm}
        Y
        >{\raggedright\arraybackslash}p{5.3cm} 
        >{\raggedright\arraybackslash}p{2.0cm}
        @{}
    }
    \toprule
    \textbf{Method} & \textbf{Metric (as reported)} & \textbf{Setting / Protocol Note} & \textbf{Source} \\
    \midrule

    \rowcolor{gray!8}
    \multicolumn{4}{l}{\textit{LIBERO~\citep{LIBERO_Liu_2023}/CALVIN~\citep{CALVIN_2022} manipulation benchmarks --- task success}} \\
    \addlinespace[2pt]
    UniVLA~\citep{UniVLA_RSS_25} &
    LIBERO Avg.\ SR (\%) over Sp/Ob/Go/Lo: {95.2} (full pretrain); 92.5 (Bridge-V2); 88.7 (human-video) &
    Four suites, 50 traj. per task for training; result avg.\ over Sp/Ob/Go/Lo &
    Table~I \\
    \addlinespace[2pt]

    LAPA$^{*}$~\citep{LAPA_ye2025} &
    LIBERO Avg.\ SR (\%) over Sp/Ob/Go/Lo: {65.7} &
    Four suites, 50 traj. per task for training; result avg.\ over Sp/Ob/Go/Lo &
    UniVLA Table~I \\
    \addlinespace[2pt]

    HiLAM~\citep{HiLAM_kim2026} &
    LIBERO-Lo SR (\%): {94}; Sp/Ob/Go are reported in Fig.~3a bar chart without exact numeric values &
    Four LIBERO suites, 10 tasks per suite and 50 demos per task; exact data-efficiency values are reported for Lo &
    Fig.~3a--b; Sec.~4.2.1 \\
    \addlinespace[2pt]

    RotVLA~\citep{RotVLA_li2026} &
    LIBERO SR (\%) over Sp/Ob/Go/Lo: {98.2}\,/\,{99.6}\,/\,{98.4}\,/\,{96.4}; Avg.\ {98.2} &
    Four LIBERO suites&
    Table~1 \\
    \addlinespace[2pt]

    ALAM~\citep{ALAM_tang2026} &
    LIBERO SR (\%) over Sp/Ob/Go/Lo: {99.2}\,/\,{99.6}\,/\,{99.0}\,/\,{94.4}; Avg.\ {98.1} &
    Four LIBERO suites; reported as $\pi_0$ + ALAM with a 3B flow-matching backbone &
    Table~2 \\
    \addlinespace[2pt]


    Moto~\citep{Moto_chen2025} &
    CALVIN (ABC$\rightarrow$D) Avg.\ Len.\ {3.10}; chain SR (\%, 1-5 tasks): 89.7 / 72.9 / 60.1 / 48.4 / 38.6 &
    Long-horizon multi-task chains (1000 sequences); Static RGB only &
    Table~3 \\

    \addlinespace[4pt]
    \midrule
    \rowcolor{gray!8}
    \multicolumn{4}{l}{\textit{SIMPLER-family~\citep{SIMPLER_li_CORL_2024} simulated manipulation evaluations}} \\

    LAPA$^{\dagger}$~\citep{LAPA_ye2025} &
    Avg.\ SR (\%): 57.3 (BridgeV2 video pretraining); 52.1 (SomethingV2 pretraining) &
    Four tasks: stack block, carrot-to-plate, spoon-to-towel, eggplant-to-basket &
    LAPA Table~11; ConLA Table~1 \\
    \addlinespace[2pt]
    
    ConLA~\citep{ConLA_dai2026} &
    Avg.\ SR (\%): {60.4} (BridgeV2 video pretraining); {64.6} (SomethingV2 pretraining) &
    Same four-task SIMPLER setting as the LAPA BridgeV2 evaluation &
    Table~1 \\
    \addlinespace[2pt]
    
    Moto~\citep{Moto_chen2025} &
    Avg.\ SR (\%): {61.4} (Open-X pretraining) &
    Tasks include Pick Coke Can, Move Near, and Open/Close Drawer &
    Table~2 \\

    \addlinespace[4pt]
    \midrule
    \rowcolor{gray!8}
    \multicolumn{4}{l}{\textit{Other evaluations (not shared across papers)}} \\
    \addlinespace[2pt]

    RotVLA~\citep{RotVLA_li2026} &
    RoboTwin2.0 SR (\%): Clean {89.6}, Random {88.5} &
    50 bimanual tasks; 100 rollouts per task &
    Table~1 \\
    \addlinespace[2pt]


    CLASP~\citep{CLASP_ICLR2019} &
    Reacher visual servoing, final distance [deg]: {1.6$\pm$1.0}; 3.0$\pm$2.2 (varied bg.); 2.8$\pm$2.9 (varied agents) &
    MPC/CEM in latent-action space; cosine distance on VGG16 features &
    Table~2 \\

    \bottomrule
    \end{tabularx}

    \vspace{2pt}
    \raggedright
    \scriptsize
    $^{*}$ Reproduced in UniVLA using the Prismatic-7B VLM (per UniVLA Table~I). $^{\dagger}$ BridgeV2 result is reported in LAPA Table~11; SomethingV2 result is reported/reproduced in ConLA under the same four-task SIMPLER setting.
    \textit{Sp\,/\,Ob\,/\,Go\,/\,Lo}: LIBERO Spatial\,/\,Object\,/\,Goal\,/\,Long suites.
\end{table*}

\subsection{Execution and control integration}
\label{subsec:latent-control-integration}

Latent-action methods make control modular by inserting a learned action abstraction between video observations and robot commands.
This abstraction can be planned over, used as a pretraining target, retained as a policy output token, co-generated with robot actions, or organized into longer-horizon skills.
The same factorization also creates three recurring failure points: the latent must correspond to controllable change, the latent forward model that defines or uses it must remain physically meaningful, and the grounding mechanism must preserve embodiment-specific control details.

\paragraph{Identifiability: does the latent correspond to controllable change?}
The central promise of latent actions is that the learned variable captures ``what changed'' between observations.
However, a video transition can reflect many simultaneous causes: the robot's action, camera motion, other agents, lighting, background change, or passive object dynamics.
Information bottlenecks and VQ codebooks encourage compression, but they do not guarantee that the retained information is \emph{controllable} by the robot.
A latent code may therefore predict visual change while mixing robot-caused effects with exogenous variation, making the subsequent grounding map reliable only in the training distribution.

This issue appears directly in CLASP-style latent planning, where a continuous transition latent discovered from pixels is later mapped to robot commands.
If the latent absorbs viewpoint, background, or coincident object motion, visually similar codes can require different robot actions under changed conditions.
Instruction-conditioned methods address the same problem by changing the discovery objective: UniVLA separates task-centric change from nuisance variation, ConLA pushes the codebook toward action-related motion rather than appearance, and RotVLA / ALAM impose local compositional structure.
These biases improve action-likeness, but they do not provide causal identifiability; the learned latent can still encode correlations that are predictive in the discovery data but brittle under embodiment or scene shift.

\paragraph{Physical consistency of latent forward models.}
Even when latent actions capture controllable change, the transition model that defines or uses the latent space may remain visually predictive rather than physically grounded.
For CLASP-style planning, this issue is direct: latent-action sequences are selected through a learned forward model, so visually plausible but dynamically infeasible predictions can lead to poor grounded actions.
For example, the model may predict object motion without a contact-consistent end-effector configuration that could have produced it.

The same concern appears more indirectly in instruction-conditioned methods.
RotVLA and ALAM encourage local algebraic consistency, while HiLAM imposes temporal structure through latent skills.
These biases organize the latent space, but they do not guarantee that latent composition matches the physical composition of executable robot actions.
Physical consistency therefore also matters when latent tokens, co-generated streams, or skills continue to shape executable behavior.

\paragraph{Grounding brittleness and the action-space gap.}
Grounding connects video-derived latents to executable robot actions, but it also exposes an action-space gap.
A latent may specify coarse visual change while omitting timing, force, compliance, gripper state, or other embodiment-specific information needed for execution.
The gap is most visible in standalone interfaces: CLASP maps planned continuous latents to robot commands, so one latent transition may correspond to different low-level actions depending on contact state, robot configuration, or object pose.
Non-manipulation precursors such as FICC and LAPO illustrate related discrete-code grounding mechanisms through co-occurrence statistics, decoders, or online adaptation, but these mechanisms do not by themselves resolve the manipulation-specific grounding problem.

In instruction-conditioned policies, the action-space gap depends on how long the latent remains active.
Head-replacement and co-training variants avoid executing the latent directly, but can inherit a representation gap if latent-code supervision under-specifies fine-grained control before the policy switches to native actions.
Persistent-token, structured-stream, and hierarchical variants keep more latent structure active during control, which can preserve temporal abstraction but also lets discovery errors influence executable behavior more directly.
Thus, persistence does not automatically solve grounding; it shifts the burden from learning a one-shot latent$\rightarrow$action map to maintaining alignment between latent structure and native control throughout policy learning and deployment.
In practice, this can require grounding modules to encode embodiment constraints---joint limits, collision avoidance, gripper state, or contact feasibility---rather than treating latent$\rightarrow$action mapping as purely statistical regression.

\subsection{Summary and takeaways}

Latent-action methods make the video-to-control interface more explicit than direct video--action policies, but still more opaque than hand-specified visual plans or trajectories.
Their central move is to compress observation transitions into an action-like variable learned from video, then connect that variable to executable control with limited robot-action supervision.
This creates a useful modular decomposition---discovery from action-free video, followed by robot-specific grounding---but also exposes a recurring tension: a compact latent may be predictive of visual change without being identifiable, physically meaningful, or reliably executable.
Figure~\ref{fig:latent_action_pipeline} and Tables~\ref{tab:latent-action-comparison}--\ref{tab:latent-action-capabilities} summarize the pipeline, structural choices, and grounding mechanisms; Table~\ref{tab:latent_action_quant} provides a reported quantitative snapshot with protocol caveats.

\paragraph{What video supervision contributes.}
Video supervision contributes most directly by turning temporal change into a reusable training signal.
Instead of learning only static visual features, latent-action methods train a transition model whose bottleneck is pressured to retain what changes between observations.
This can reduce the need for dense robot-action labels, support latent-space planning or policy pretraining, and enable cross-domain transfer from action-free videos.
The benefit is strongest when the latent captures controllable task progress rather than nuisance variation; otherwise, the same compression can preserve camera motion, background change, or coincidental scene dynamics that are visually predictive but not executable by the robot.

\paragraph{Discovery, grounding, and deployment persistence.}
Spanning both subfamilies, the persistence spectrum runs from standalone interfaces such as CLASP, which retain the latent as an explicit planning object exposed to forward-model error and grounding ambiguity, to instruction-conditioned designs that either discard the latent after pretraining or keep it active as tokens, streams, or skills.
Deployment persistence therefore trades interpretability and modularity against accumulated mismatch between the learned latent space and the robot's executable action space.

\paragraph{Design patterns and failure modes.}
Across architectures, the recurring lesson is that a latent action is useful only when its abstraction stays aligned with what the robot can execute; the three failure modes behind this---identifiability, physical consistency of the latent forward model, and grounding brittleness---are developed in Section~\ref{subsec:latent-control-integration}.

\paragraph{Evidence and evaluation fragmentation.}
The more reliable evidence in this family is often internal: ablations showing whether video-derived latent supervision improves sample efficiency, generalization, or policy learning within a given method.
Cross-method claims about bottleneck type, latent persistence, or grounding strategy remain suggestive unless evaluated under shared tasks and action spaces.

\paragraph{Adjacent directions outside the survey's scope.}
A few closely related methods relax the action-free discovery assumption of the latent-action family: CLAM~\citep{CLAM_liang2025} and LAOM~\citep{LAOM_2025latent} introduce action supervision into latent-action discovery, while villa-X~\citep{villaX_chen2026} incorporates robot proprioceptive state and actions when learning or grounding latent actions.
These methods remain centered on latent-action interfaces, but robot-side supervision participates in shaping the interface rather than only grounding an interface learned from non-action-annotated video.
We therefore treat them as adjacent to the core latent-action taxonomy rather than as core methods, while noting the design point they illustrate: modest embodiment-specific supervision may improve controllability and grounding.
Characterizing when such supervision is useful---and how much action-side supervision suffices for which downstream gains---is a natural extension of the analysis presented here.

\paragraph{Relation to explicit interfaces.}
A defining feature of latent-action methods is that the intermediate abstraction is learned from data rather than specified by a designer: latent actions emerge from transition modeling and are not inherently inspectable or interpretable.
This makes them flexible---they can capture diverse dynamics from large video corpora---but also difficult to verify, debug, or constrain to physically valid behaviors.
The next family of methods, explicit visual interfaces, addresses this by exposing structured, human-readable predictive signals (e.g., goal images, drawn trajectories, waypoint sequences) as the interface between video-derived temporal understanding and downstream controllers.
This creates a progression across the three families surveyed: direct methods keep the video-to-control link implicit in learned features, latent-action methods make it an explicit but opaque variable, and explicit interface methods make it a structured, inspectable plan.
We return to cross-cutting issues---including controllability, temporal abstraction, and grounding protocols---in Section~\ref{sec:discussion_future}.

\section{Explicit Visual Control Interfaces}
\label{sec:visual-interfaces}

\begin{figure*}
    \centering
    \includegraphics[width=1\linewidth, trim= 30 10 30 10, clip]{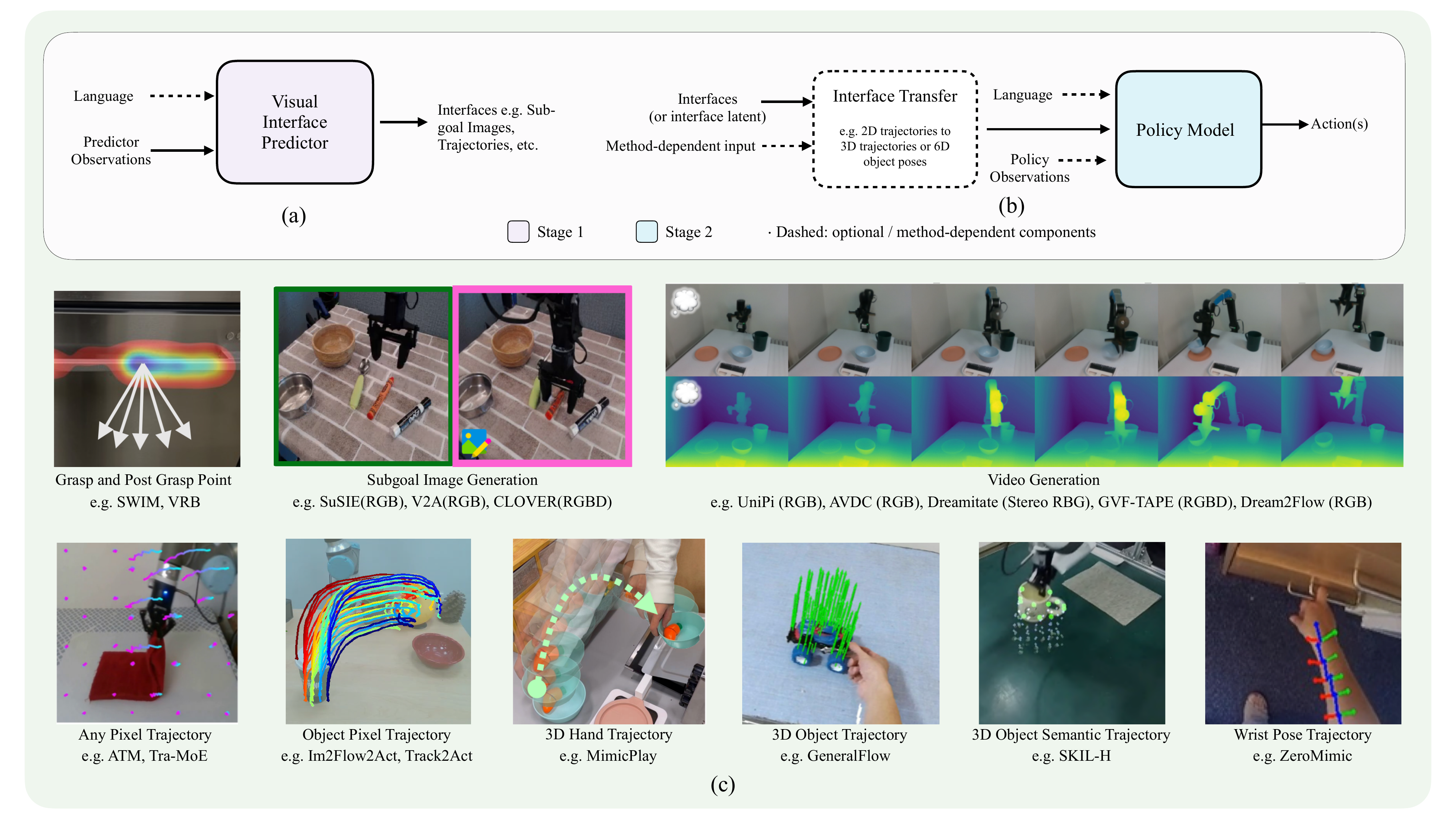}
    \caption{\textbf{Explicit visual interface--based methods for robotic manipulation.} Sub-figure \textbf{(a--b)} These methods decouple perception and control by first predicting a human-interpretable interface (e.g., subgoal images, trajectories, or pose sequences) from visual observations and optional language, and then mapping the interface to robot actions with a downstream policy. Some approaches additionally apply an interface transfer step (e.g., 2D trajectories $\rightarrow$ 6D object poses) before policy execution. Sub-figure \textbf{(c)} Representative explicit interfaces used in prior work, parentheses denote interface format (e.g., RGB, RGB-D) where methods in the same group differ; examples are reproduced/adapted from~\citep{SWIM_2023,susie_2023,GVFTAPE_CoRL_2025,ATM_Wen_RSS_2024,Im2Flow2Act_Xu_CoRL_2024,MimicPlay_CoRL_2023,GeneralFlow_CoRL_2024,SKIL_RSS_2025,ZeroMimic_ICRA_2025}. \textit{Methods in this section typically pretrain the interface predictor on video without action labels, and learn the downstream policy from robot data.}}
    \label{fig:explicit_interface_pipeline}
\end{figure*}

The two previous families connect video understanding to robot control through mechanisms that keep the video-to-action relationship largely implicit.
Direct video--action policies (Section~\ref{sec:direct-video-action}) embed video and action prediction in a single model, where the connection between visual dynamics and control remains in shared latent features rather than an interpretable intermediate signal.
Latent-action methods (Section~\ref{sec:latent_actions}) introduce an abstract action variable that explains visual transitions, but the resulting latent space is compact and difficult to inspect, complicating debugging and analysis.

This section explores a third design philosophy: methods that extract \emph{explicit visual interfaces} from video---structured, human-interpretable signals such as subgoal images, video plans, point trajectories, or pose sequences---and train robot policies to consume these signals as transparent, mid-level control targets.
The key insight is that, rather than compressing video knowledge into shared latent features or abstract action codes, these methods predict visual quantities that a human can inspect and that a separate policy can follow.
This modular separation between interface prediction (trained on large-scale action-free video) and action generation (trained on smaller robot datasets) offers three main benefits.
(i)~\emph{transparency}---one can visualize predicted trajectories or subgoal images before execution to verify plausibility;
(ii)~\emph{cross-embodiment transfer}---interfaces defined in visual space (e.g., pixel trajectories from human video) can guide policies on different robots; and
(iii)~\emph{flexible composition}---video models and controllers can be improved independently, and interfaces can be combined with other inputs such as language or proprioception.
The cost is an additional design choice: what structure to extract and how to represent it, balancing expressiveness against robustness to perception noise and domain shift.
Here, \emph{action-free} refers to video without synchronized action labels: the interface predictor may be trained on human, Internet, or robot videos as long as actions are not required, while the downstream controller may still use action-labeled robot data for grounding.

Figure~\ref{fig:explicit_interface_pipeline} illustrates the generic pipeline: an interface predictor trained from action-free video outputs an explicit visual signal (e.g., a video plan, subgoal image, or point trajectories) from the current observation and an optional language instruction; a downstream controller then maps this interface, together with robot state, to low-level actions. 

We group methods by this \emph{primary predicted interface} produced by the visual module (Table~\ref{tab:explicit-interface-comparison}). 
Some approaches additionally apply an \emph{interface-transfer} step---for example, converting a generated video plan into object/EE pose trajectories (e.g., AVDC, RIGVid, GVF-TAPE) or lifting predicted tracks into SE(3) motion targets (e.g., Track2Act)---which we treat as a downstream grounding choice rather than the basis of the taxonomy.

We treat a method as part of this family when it exposes a temporally structured, visually derived target to the controller, or when the controller operates on a compact encoding of such an inspectable target, such as a subgoal image, video plan, point trajectory, object flow, or pose sequence.
We exclude methods whose intermediate is primarily an unconstrained latent feature vector or a learned action abstraction; these belong instead to the direct video--action and latent-action families, respectively.

\noindent\textbf{Organization.}
We divide explicit-interface methods into two categories based on the nature of the predicted signal.
\emph{Frame-based interfaces: goal images and video plans} (\S\ref{sec:goal-images}) synthesize future visual states that goal-reaching or plan-following policies execute.
\emph{Trajectory-based interfaces} (\S\ref{sec:trajectory-interfaces}) predict lower-dimensional motion signals---pixel tracks, 3D keypoint trajectories, affordance waypoints, or 6D pose sequences---that policies consume as explicit motion targets.
The distinction reflects a trade-off between \emph{holistic guidance} (images provide rich context but are harder to track precisely) and \emph{compact precision} (trajectories are easier to ground to actions but may miss visual context).
Tables~\ref{tab:explicit-interface-comparison} and~\ref{tab:explicit-interface-capabilities} summarize the structural design choices and training sources of these methods.

\subsection{Goal images and video plans}
\label{sec:goal-images}

Methods in this category expose an explicit \emph{frame-based visual target} for control---either a subgoal image or a short-horizon video plan---that a downstream controller attempts to realize in the physical scene.
Given the current observation and a task specification (e.g., language or a goal image), the method predicts a future visual state (single-frame subgoal) or a sequence of intermediate frames (video plan) that can be inspected before execution.
A separate goal-reaching module then grounds this interface to actions, for example via goal-conditioned visuomotor policies, inverse dynamics between consecutive frames, or pose/trajectory tracking after interface transfer.
Within this category, methods differ along several axes: whether the interface is a single subgoal image or a dense video sequence; how visual predictions are converted to actions (inverse dynamics, pose estimation, or goal-conditioned policies); and whether execution is open-loop (follow the plan once) or closed-loop (replan adaptively).
Some methods further perform interface transfer, converting video predictions into lower-dimensional targets such as object or end-effector poses before control.
Figure~\ref{fig:explicit_interface_pipeline}(c) illustrates the variety of visual interfaces in this category, from dense video rollouts to single subgoal images; Table~\ref{tab:explicit-interface-capabilities} shows that most methods train the interface predictor from action-free Internet or human video, while grounding is trained on comparatively small robot datasets.
We organize the methods below into three design-pattern clusters based on how the predicted visual interface is grounded to robot actions.

\paragraph{Historical precursor.}
AVID~\citep{AVID1_smith2020} is an early explicit-interface formulation that draws on action-free human video for multi-stage robotic tasks by translating human demonstrations into robot-relevant visual targets.
Rather than predicting robot actions from video directly, AVID learns a pixel-level translation that maps a human video segment into a corresponding robot-view visual plan, which can then be consumed by a downstream controller trained on comparatively limited robot data.
This design illustrates a recurring theme in explicit interfaces: temporal structure can be harvested from large, unlabelled video corpora as an inspectable intermediate signal, while robot-specific action grounding is delegated to a separate module that operates on the translated visual targets.

\begin{table*}[t]
  \caption{\textbf{Training sources, deployment requirements, and real-robot evaluation scope of explicit visual interface methods.}
``Video Source'' = data for the interface predictor (action-free unless noted);
``Robot Data'' = supervision needed to ground the interface on a robot;
``Real-Robot Scope'' = scope of real-robot experiments reported in the paper.
Quantitative results on shared benchmarks (CALVIN~\citep{CALVIN_2022}, LIBERO~\citep{LIBERO_Liu_2023}) are in Table~\ref{tab:explicit_interface_quant}.}
  \label{tab:explicit-interface-capabilities}
  \centering
  \footnotesize
  \setlength{\tabcolsep}{3pt}
  \renewcommand{\arraystretch}{1.35}
  \begin{tabularx}{\textwidth}{
      >{\raggedright\arraybackslash}p{3.2cm}
      >{\raggedright\arraybackslash}X
      >{\raggedright\arraybackslash}p{2.9cm}
      >{\centering\arraybackslash}p{1.2cm}
      >{\raggedright\arraybackslash}p{2.0cm}
      >{\raggedright\arraybackslash}p{2.8cm}
    }
    \toprule
    \textbf{Method} & \textbf{Video Source} & \textbf{Robot Data} & \textbf{Cross-E.} & \textbf{Task Scope} & \textbf{Real-Robot Scope} \\
    \midrule

    \rowcolor{gray!8}
    \multicolumn{6}{l}{\textit{Frame-based interfaces}} \\
    \addlinespace[2pt]

    UniPi~\citep{UniPi_Du_NIPS_2023}
      & Internet text-video + robot video-text finetuning 
      & Robot demos
      & \checkmark
      & General
      & Limited (reimpl.\ by SuSIE) \\

    Gen2Act~\citep{Gen2Act_bharadhwaj2024}
      & Internet (pretrained generator; frozen)
      & Robot demos + paired generated human videos
      & \checkmark
      & General
      & multi-setting generalization test \\

    AVDC~\citep{AVDC_2023}
      & Robot video (action-free); diffusion predictor finetuned in-domain
      & None (zero-shot; grasp required)
      & \checkmark
      & Rigid objects
      & Real robot + Meta-World sim \\

    RIGVid~\citep{RIGVid_patel2025}
      & Pretrained video generator (frozen)
      & None (zero-shot; online tracking)
      & \checkmark
      & Rigid objects
      & 4 tasks; zero-shot \\

    Dreamitate~\citep{Dreamitate_Liang_CoRL_2024}
      & Collected stereo human tool-use video 
      & None (zero-shot after pose tracking)
      & Limited
      & Tool manip.
      & 4 tool tasks \\

    GVF-TAPE~\citep{GVFTAPE_CoRL_2025}
      & Robot video (action-free); RGB + monocular depth
      & Robot motions with EE pose labels
      & Limited
      & General
      & 5 tasks (rigid/deform/artic.) \\

    Dream2Flow~\citep{dream2flow_dharmarajan2025}
      & Internet (pretrained generator; frozen) + perception
      & Test-time optimization / RL
      & \checkmark
      & Rigid, artic., deform., granular 
      & 3 tasks; zero-shot \\

    SuSIE~\citep{susie_2023}
      & Internet-pretrained image editor + human/robot videos
      & Robot demos
      & \checkmark
      & General
      & 3 scenes, 9 tasks \\

    CLOVER~\citep{CLOVER_2024}
      & Human/robot RGB-D
      & Robot demos
      & Limited
      & General
      & 3 tasks (short + long-horizon) \\

    V2A~\citep{V2A_2025}
      & Internet (pretrained; frozen video model)
      & Self-exploration / interaction
      & \checkmark
      & General
      & --- \\

    \addlinespace[4pt]
    \midrule
    \rowcolor{gray!8}
    \multicolumn{6}{l}{\textit{Trajectory-based interfaces}} \\
    \addlinespace[2pt]

    VRB~\citep{VRB_2023}
      & Human egocentric video
      & Robot demos / RL / planning
      & \checkmark
      & Contact-rich
      & 8 tasks, 2 robots \\

    SWIM~\citep{SWIM_2023}
      & Human egocentric video
      & Small robot dataset (adaptation)
      & \checkmark
      & Grasp-centric
      & 6 tasks, 2 robots \\

    MimicPlay~\citep{MimicPlay_CoRL_2023}
      & Human play (calibrated multi-view)
      & Robot demos
      & \checkmark
      & Long-horizon
      & 4 long-horizon tasks \\

    ATM~\citep{ATM_Wen_RSS_2024}
      & Large action-free video (traj.\ pseudo-labels)
      & Small robot dataset
      & \checkmark
      & General
      & 3 transfer tasks (human$\rightarrow$robot) \\

    Im2Flow2Act~\citep{Im2Flow2Act_Xu_CoRL_2024}
      & Human demo video
      & Sim play (no robot demos)
      & \checkmark
      & Object manip.
      & 4 tasks; no real training \\

    Track2Act~\citep{track2act_ECCV_2024}
      & Web human + passive robot videos; depth at execution
      & Small robot dataset (residual)
      & \checkmark
      & Rigid objects
      & 25 tasks, 5 locations \\

      SKIL-H~\citep{SKIL_RSS_2025}
      & Human + robot videos (for trajectory prediction)
      & Robot demos only for traj.$\rightarrow$action grounding
      & \checkmark
      & General
      & 3 tasks; cross-embodiment study \\

    GeneralFlow~\citep{GeneralFlow_CoRL_2024}
      & Human RGB-D video
      & None (zero-shot)
      & \checkmark
      & Rigid, artic., deform.
      & 18 tasks, 6 scenes; zero-shot \\

    ZeroMimic~\citep{ZeroMimic_ICRA_2025}
      & Web egocentric video
      & None (post-grasp); separate grasp module
      & \checkmark
      & Post-grasp
      & 9 skills, 2 robots; zero-shot \\

    \bottomrule
  \end{tabularx}

  \vspace{3pt}
  \raggedright
  \footnotesize
  \textit{Cross-E.} = Cross-embodiment transfer. \checkmark\ = supported; \textit{Limited} = constrained by data/assumptions.
  ``---'' = no separate real-robot evaluation reported (sim/shared-benchmark only).
\end{table*}

\subsubsection{Dense video plans with direct action grounding.}
\label{subsec:dense-video-plans}

Dense video-plan methods generate a sequence of future frames as the interface and derive actions directly from that sequence---through inverse dynamics or policy conditioning (see the ``Action Derivation'' column of Table~\ref{tab:explicit-interface-comparison})---without an intermediate geometric transfer step.
The plan is inspectable in visual space, while feasibility is enforced only indirectly through the downstream grounding module and the robot data used to train it.

\paragraph{UniPi~\citep{UniPi_Du_NIPS_2023}.}
UniPi uses a dense \emph{video plan} as the explicit interface for control: given a current frame and a text goal, it synthesizes a future-frame sequence and then grounds the plan to actions by applying an inverse-dynamics model between consecutive generated frames.
The video plan is produced by a text-conditioned video diffusion model, and is generated hierarchically---first predicting a temporally sparse set of keyframes and then filling in intermediate frames via temporal super-resolution to form a coherent dense rollout.
Pretraining on large-scale Internet text--video data provides broad visual priors and supports combinatorial generalization to novel instructions that compose previously seen concepts.
Within this design, the plan remains human-inspectable before execution, while execution quality depends on how reliably the inverse-dynamics module can translate frame-to-frame changes into feasible robot actions.

\paragraph{Gen2Act~\citep{Gen2Act_bharadhwaj2024}.}
Gen2Act also uses a generated rollout as the interface, but shifts grounding into a single closed-loop visuomotor policy: instead of an explicit inverse-dynamics module, the policy directly conditions on the generated video and maps it to robot actions using observation history for feedback during execution.
The method synthesizes a short \emph{human} demonstration video with a frozen off-the-shelf generator, and trains on robot demonstrations paired with corresponding generated guidance videos; an auxiliary point-track motion objective further encourages the policy to encode the rollout’s motion structure.
This trades an explicit frame-to-action grounding step for end-to-end video-conditioned control, placing more of the interface-to-action burden on supervised video-to-action translation.

\subsubsection{Video plans with interface transfer to poses and trajectories.}
\label{subsec:video-plan-transfer}

A second group keeps the predicted video or frame sequence as the primary interface, but adds a geometric \emph{interface-transfer} step that converts the visual plan into lower-dimensional pose or trajectory targets before control; the ``Transferred To'' column in Table~\ref{tab:explicit-interface-comparison} summarizes the target used by each method.
This transfer trades the holistic richness of video for a compact, action-grounding-friendly representation---at the cost of additional perception steps such as depth estimation, correspondence matching, and pose fitting, which can introduce errors under clutter, occlusion, or deformable objects.

\paragraph{AVDC~\citep{AVDC_2023}.}
AVDC uses a generated \emph{video plan} as an explicit planning interface, but grounds it to control through geometric \emph{video$\rightarrow$pose transfer} rather than a learned inverse-dynamics model.
Given the current RGB-D observation and a task specification (e.g., language), a diffusion model synthesizes an ``imagined execution'' video; dense correspondences (optical flow) between successive predicted frames are lifted to 3D using the depth of the first frame, and a rigid SE(3) motion is recovered by fitting the transformation consistent with the 3D--2D correspondences (PnP-style with robust estimation~\citep{EPnP_IJCV_2009, RANSAC}).
For execution, the robot grasps the target object and applies the inferred rigid motion to produce end-effector commands that move the grasped object along the recovered pose trajectory.
This design avoids action annotations during interface prediction, but depends on the grasped rigid-object assumption and on reliable correspondence/depth lifting, which can be brittle in clutter or for deformable objects.

\paragraph{RIGVid~\citep{RIGVid_patel2025}.}
RIGVid keeps the generated video rollout as the primary visual interface, but makes AVDC-style video-plan-to-pose transfer more robust by addressing two common failure modes: semantic mismatch between the rollout and the instruction, and drift during pose execution.
Rather than committing to a single generated video plan, it samples multiple candidates from a pretrained video diffusion model and uses a vision--language model to select an instruction-consistent rollout.
The selected rollout is then transferred into a 6D object-pose trajectory using depth estimation, metric alignment to the real initial observation, and object pose tracking; this pose trajectory is retargeted to end-effector motion under a grasp/contact assumption and tracked in closed loop.
Compared with AVDC, this design improves the reliability of the video-plan interface through rollout selection and closed-loop tracking, but it also adds new dependencies on semantic filtering, depth alignment, and reliable 6D pose estimation.

\paragraph{Dreamitate~\citep{Dreamitate_Liang_CoRL_2024}.}
Dreamitate makes video-plan-to-pose transfer tool-centric: its primary visual interface is a short-horizon sequence of stereo frame pairs, which is then converted into a 6D \emph{tool-pose sequence} for execution.
The stereo predictions are produced by fine-tuning a foundation video generation model; given each generated stereo pair and a known CAD model of the tool, a pose estimator (MegaPose~\citep{MegaPose_2022}) recovers the tool's 6D pose.
The robot follows the resulting tool-pose trajectory with the tool attached to the end-effector.
This design makes the transferred interface well aligned with tool-use tasks, but it also narrows applicability: reliable execution depends on stereo prediction, known tool models, and accurate tool-pose estimation.

\paragraph{GVF-TAPE~\citep{GVFTAPE_CoRL_2025}.}
GVF-TAPE makes two shifts in video-plan interface transfer: it predicts RGB-D future observations from an RGB input, and transfers those predictions to \emph{end-effector pose} targets rather than manipulated-object poses.
A language-conditioned video predictor forecasts short-horizon RGB-D observations from the current view and instruction, and a pose estimator then extracts the end-effector poses implied by the predicted frames.
This weakens the rigid manipulated-object assumption used by AVDC-style transfer, making the interface better suited to deformable or hard-to-track objects.
The trade-off is that execution depends on the quality and diversity of robot videos used to train the predictor, as well as the reliability of depth prediction and end-effector pose estimation from generated frames.

\paragraph{Dream2Flow~\citep{dream2flow_dharmarajan2025}.}
Dream2Flow keeps generated interaction video as the primary frame-based interface, but transfers this video plan into object-centric \emph{3D motion targets} rather than a single rigid SE(3) pose sequence.
Given an initial RGB observation and a language instruction, an off-the-shelf image-to-video model synthesizes a plausible interaction video; the generated object motion is then lifted into 3D using segmentation, point tracking, monocular video depth, and scale alignment to the robot's initial RGB-D observation.
Execution tracks the resulting object trajectories through trajectory optimization or reinforcement learning, enabling video-plan interfaces for non-rigid or distributed motion such as deformable and granular manipulation.
The benefit is a richer transfer target than rigid pose tracking, but the cost is a heavier perception-and-control pipeline whose reliability depends on accurate segmentation, tracking, depth estimation, and trajectory tracking.

\subsubsection{Subgoal-image interfaces with goal-conditioned policies.}
\label{subsec:subgoal-images}

A third pattern uses \emph{visual subgoals} as the interface, paired with a goal-conditioned policy that executes short-horizon actions to reduce visual mismatch.
These subgoals may be single RGB images or ordered RGB-D subgoal sequences, but in both cases the controller treats them as intermediate visual targets rather than dense frame-by-frame rollouts.
By iterating subgoal prediction, goal-reaching, and replanning, these methods compose long-horizon behaviors from repeated short-horizon segments.

\paragraph{SuSIE~\citep{susie_2023}.}
SuSIE uses a \emph{single subgoal image} as the explicit interface: a high-level planner proposes an intermediate visual target from the current observation and a language command, and a separate goal-conditioned policy executes short-horizon actions to reach it in a repeated predict-then-reach loop.
The planner is implemented by fine-tuning an Internet-pretrained image-editing diffusion model (e.g., InstructPix2Pix~\citep{InstructPix2Pix_2023}) on a mixture of human videos and robot rollouts so that it outputs a hypothetical future observation a few steps ahead.
Grounding is provided by a low-level goal-conditioned policy trained on robot data to move the scene toward a given goal image; at test time, SuSIE alternates between generating a new subgoal and executing the controller to compose long-horizon behaviors.
The interface is human-inspectable and modular, but the approach relies on subgoals being reachable within the controller's horizon; unrealistic edits or large visual jumps can break the composition.

\paragraph{CLOVER~\citep{CLOVER_2024}.}
CLOVER follows the same ``visual planner + goal-conditioned policy'' pattern as SuSIE, but replaces RGB subgoal images with text-conditioned \emph{RGB-D subgoal-sequence} plans and adds explicit error-driven feedback for closed-loop progression and replanning.
A multimodal encoder compares the current RGB-D observation with the active visual subgoal to form an error signal, which an inverse-dynamics policy decodes into end-effector and gripper actions.
Execution advances to the next subgoal when progress is sufficient, or replans when the current sequence appears infeasible.
Compared with SuSIE, CLOVER exposes more temporal and geometric structure, but it also relies on accurate RGB-D subgoal prediction and a learned error metric for robust switching and replanning.

\paragraph{V2A~\citep{V2A_2025}.}
V2A keeps the subgoal-image interface but changes the supervision source: instead of demonstration-based imitation, a goal-reaching policy is trained from self-collected rollouts with hindsight relabeling, guided by subgoals predicted from a frozen pretrained video model.
This eliminates the need for action-labeled demonstrations, but shifts the burden to extensive environment interaction, which can be difficult to scale safely on real robots.

\paragraph{Takeaways for frame-based interfaces.}
Across frame-based interfaces, the main design axis is how much temporal structure the interface exposes: dense video plans specify a visual rollout over time, whereas subgoal-image methods provide a sparse future target and leave intermediate behavior to a goal-conditioned policy (Table~\ref{tab:explicit-interface-comparison}).
This creates a central grounding trade-off.
Dense plans are more inspectable and can express how a task should unfold, but they require the generated frames to be translated into actions through inverse dynamics, geometric interface transfer, or pose/trajectory tracking, which introduces failure points when predictions are unrealistic, misaligned, or hard to track.
Subgoal-image methods simplify the visual interface and support frequent replanning, but push more of the image-to-action burden into the goal-conditioned controller.
The clearest shared numeric slice is CALVIN ABC$\rightarrow$D: combined with the method descriptions above, Table~\ref{tab:explicit_interface_quant} shows higher chained-instruction success rates for closed-loop subgoal-image methods than for the open-loop dense-video-plan baseline under the same train/test split.
The shared slice therefore points more directly to the importance of feedback and grounding than to interface density alone: it also conflates plan realism, the grounding pipeline, and replanning rate, so it does not establish that sparse subgoals are intrinsically superior to dense plans.

\subsection{Trajectory-based interfaces}
\label{sec:trajectory-interfaces}

Rather than predicting full images or videos, methods in this category expose an explicit \emph{motion target} for control---pixel tracks, 3D point/keypoint trajectories, affordance waypoints, or 6D pose sequences---that a downstream controller tracks over time.
Because these signals describe how scene elements should move, they can often be extracted from action-free human or web video and then retargeted to robot execution. 

Methods differ primarily in what is represented (scene points, objects, hands/end-effectors) and in how the target enters the control stack (trajectory-conditioned policies, geometric tracking/IK, MPC-style replanning, or exploration priors).
A useful distinction is where geometric commitment happens: some methods keep the interface in image space, while others lift it into 3D keypoints, object trajectories, or 6D pose targets.
This progression, visualized in Figure~\ref{fig:explicit_interface_pipeline}(c), motivates our organization below into local contact-centric interfaces, image-space 2D trajectories, and metric 3D/6D trajectories.
Table~\ref{tab:explicit-interface-comparison} complements this grouping by showing how each interface is consumed by the downstream controller, ranging from learned trajectory-conditioned policies to geometric execution and retargeting.

\subsubsection{Affordance-based contact interfaces.}
\label{subsec:affordance-interfaces}

Affordance-based methods expose \emph{contact-centric} visual targets---where to touch and how to move after contact---distilled from egocentric human manipulation video (Figure~\ref{fig:explicit_interface_pipeline}(c), leftmost).
The interface is compact and interpretable, and can be lifted to 3D waypoints for execution, but it primarily captures local interaction structure rather than full long-horizon plans.

\begin{table*}[t]
  \caption{\textbf{Structural comparison of explicit visual interface methods.} 
  Methods are grouped by interface type: \emph{frame-based interfaces} predict future visual states, while \emph{trajectory-based interfaces} predict lower-dimensional motion signals. 
  ``Transferred To'' indicates whether an explicit interface-transfer step is applied; 
  ``Loop'' indicates open-loop (execute plan once) or closed-loop (replan or update targets during execution). See footnotes for marker definitions.}
  \label{tab:explicit-interface-comparison}
  \centering
  \footnotesize
  \setlength{\tabcolsep}{3pt}
  \renewcommand{\arraystretch}{1.25}

    \begin{tabularx}{\textwidth}{
    @{}
    >{\raggedright\arraybackslash}p{3.6cm}
    >{\raggedright\arraybackslash}p{3.9cm} 
    >{\raggedright\arraybackslash}p{2.0cm}
    >{\centering\arraybackslash}p{0.9cm}
    Y
    >{\centering\arraybackslash}p{0.9cm}
    @{}
  }
    \toprule
    \textbf{Method} & \textbf{Predicted Interface} & \textbf{Transferred To} & \textbf{Input} & \textbf{Action Derivation} & \textbf{Loop} \\
    \midrule
    \rowcolor{gray!8}
    \multicolumn{6}{l}{\textit{Frame-based interfaces}} \\
    \addlinespace[2pt]
    UniPi~\citep{UniPi_Du_NIPS_2023} 
      & Video plan 
      & — 
      & RGB 
      & Learned (inverse dynamics) 
      & Open \\
    Gen2Act~\citep{Gen2Act_bharadhwaj2024}
      & Human video plan
      & —
      & RGB
      & Learned (video-conditioned policy)
      & Closed \\
    \addlinespace[2pt]
    AVDC~\citep{AVDC_2023}$^\dagger$ 
      & Video plan 
      & Object 6D poses 
      & RGB-D 
      & Geometric (flow + PnP~\citep{EPnP_IJCV_2009})
      & Open \\
    {RIGVid~\citep{RIGVid_patel2025}$^{\dagger\ddagger}$}
      & {Video plan (sampled \& filtered)}
      & {Object 6D poses}
      & {RGB-D}
      & {Geometric (6D pose tracking + retargeting)}
      & {Closed} \\
    Dreamitate~\citep{Dreamitate_Liang_CoRL_2024} 
      & Stereo video plan 
      & Tool 6D poses 
      & Stereo 
      & Geometric (MegaPose~\citep{MegaPose_2022}) 
      & Open \\
    GVF-TAPE~\citep{GVFTAPE_CoRL_2025}$^\ddagger$ 
      & RGB-D video plan (monocular depth) 
      & EE 6D poses 
      & RGB 
      & Geometric (EE pose extraction/tracking from predicted RGB-D)
      & Closed \\
    Dream2Flow~\citep{dream2flow_dharmarajan2025}$^\dagger$ 
      & Video plan 
      & 3D object flow 
      & RGB-D 
      & Optimization / RL 
      & Closed \\
    \addlinespace[2pt]
    SuSIE~\citep{susie_2023} 
      & Subgoal image 
      & — 
      & RGB 
      & Learned (goal-reaching) 
      & Closed \\
    CLOVER~\citep{CLOVER_2024} 
      & Subgoal image sequence (RGB-D) 
      & — 
      & RGB-D 
      & Learned (goal-reaching) 
      & Closed \\
    V2A~\citep{V2A_2025} 
      & Subgoal images 
      & — 
      & RGB 
      & Learned (goal-reaching) 
      & Closed \\
    \addlinespace[4pt]
    \midrule
    \rowcolor{gray!8}
    \multicolumn{6}{l}{\textit{Trajectory-based interfaces}} \\
    \addlinespace[2pt]
    VRB~\citep{VRB_2023}$^\dagger$ 
      & 2D contact point + post-contact 2D traj. 
      & 3D waypoints 
      & RGB-D 
      & Learned / RL / planning 
      & Varies \\
    SWIM~\citep{SWIM_2023}$^\dagger$ 
      & 2D affordance waypoints 
      & 3D waypoints 
      & RGB-D 
      & World-model planning (CEM) 
      & Open \\
    ATM~\citep{ATM_Wen_RSS_2024} 
      & 2D pixel traj.
      & — 
      & RGB 
      & Learned (traj.-conditioned) 
      & Closed \\
    Im2Flow2Act~\citep{Im2Flow2Act_Xu_CoRL_2024} 
      & 2D object flow 
      & — 
      & RGB 
      & Learned (flow-conditioned) 
      & Closed \\
    Track2Act~\citep{track2act_ECCV_2024}$^\dagger$ 
      & 2D pixel tracks 
      & Object 6D poses 
      & RGB-D 
      & Geometric + residual policy 
      & Closed \\
    GeneralFlow~\citep{GeneralFlow_CoRL_2024} 
      & 3D object-point traj. 
      & EE SE(3) motion 
      & RGB-D 
      & Geometric (SVD alignment) 
      & Closed \\
    SKIL-H~\citep{SKIL_RSS_2025} 
      & 3D semantic keypoint traj.
      & — 
      & RGB-D 
      & Learned (traj.-conditioned) 
      & Closed \\
    MimicPlay~\citep{MimicPlay_CoRL_2023} 
      & 3D hand traj.\ (latent) 
      & — 
      & RGB 
      & Learned (plan-conditioned policy) 
      & Closed \\
    ZeroMimic~\citep{ZeroMimic_ICRA_2025} 
      & 6D wrist pose traj. 
      & — 
      & RGB 
      & Direct execution (retarget + tracking)
      & Closed \\
    \bottomrule
  \end{tabularx}

  \vspace{2pt}
  \raggedright
  \scriptsize
  \textit{traj.} = trajectory/trajectories; \textit{EE} = end-effector; \textit{CEM} = Cross-Entropy Method.\\
  $^\dagger$~Predicts interface from RGB but requires depth (sensor or RGB-D) for transfer or execution.
  $^\ddagger$~Interface includes \textit{predicted} monocular depth (not sensor depth); depth supervision may come from an external estimator (e.g., Video-Depth-Anything~\citep{video_depth_anything_chen2025}).\\
  Input modality reflects system \textit{sensor} requirements.
\end{table*}

\paragraph{VRB~\citep{VRB_2023}.}
VRB uses \emph{contact points} and \emph{post-contact 2D trajectories} as an explicit control interface: it predicts where interaction should occur in the image and how motion should proceed immediately after contact, yielding an interpretable, contact-centric target that downstream robot learning can follow.
These affordances are learned from large-scale, action-free egocentric human videos using automatically extracted hand--object interaction cues, without requiring robot actions or task labels.
At deployment, predicted contact points and trajectories can be lifted to 3D waypoints using depth and calibration and then integrated into multiple control pipelines, including collecting data for offline imitation, biasing exploration and goal-conditioned learning toward human-salient interactions, or serving as a discrete action parameterization for planning or reinforcement learning.
The main limitation is that the interface is deliberately local and contact-centric: it can underspecify longer-horizon task structure and depends on reliable perception (contact/trajectory prediction and, when used, depth-based lifting), so errors in the interface can propagate directly into execution.

\paragraph{SWIM~\citep{SWIM_2023}.}
SWIM uses pixel-grounded \emph{affordance actions} as an explicit interface for goal-directed manipulation, but couples this interface to control through \emph{latent world-model planning}.
Following VRB, actions are represented as image-space grasp and post-grasp interaction targets (lifted to 3D at execution time), distilling an interpretable contact-centric control input from egocentric human manipulation video.
The method first trains an affordance predictor from automatically extracted hand--object interaction cues in human video, then pretrains a latent world model on the same video with the affordance interface as control inputs (sampling robot-specific components such as depth/rotation when these are not present in human data), and finally adapts the world model to a target robot by fine-tuning on a small reward-free dataset collected by executing affordance proposals.
At deployment, SWIM plans in latent space using CEM over a hybrid action space that includes affordance actions and Cartesian end-effector deltas (selected via a discrete mode), scores imagined rollouts by feature-space similarity to a goal image, and executes the optimized sequence open-loop.
We include SWIM here because its primary control input is an interpretable image-space affordance interface; however, because action selection relies on world-model imagination and search, it also relates to latent-state planning methods and is best viewed as a boundary case.

\subsubsection{2D pixel-trajectory interfaces.}
\label{subsec:2d-trajectory-interfaces}

Two-dimensional pixel-trajectory methods predict the future motion of selected scene points in \emph{image-plane (pixel) coordinates}---arbitrary pixels, object-associated points, or dense flow fields---and expose these tracks as an explicit conditioning signal for control.
Because the interface is defined in the image space, it is naturally cross-embodiment and can be pretrained from action-free human or Internet video.
The trade-off is that 2D tracks discard metric 3D structure and contact forces; precise execution may therefore require robust tracking under occlusion and, in some methods, additional geometric lifting (e.g., depth-based back-projection) or robot-specific supervision.

\paragraph{ATM~\citep{ATM_Wen_RSS_2024}.}
ATM uses predicted \emph{any-point 2D trajectories} as the explicit interface between action-free video and robot control: a pretrained trajectory model forecasts how queried pixels will move over a short horizon, and a downstream visuomotor policy conditions on these tracks to produce robot actions.
The method first builds an action-free supervision signal by generating point tracks with an off-the-shelf tracker (e.g., CoTracker~\citep{Cotracker_2024}) and trains a track transformer to predict future 2D coordinates for arbitrary query points over a fixed horizon in image-plane coordinates.
For grounding, ATM trains a track-guided policy from a small action-labeled robot dataset by sampling query points (e.g., a grid), predicting their future tracks, and conditioning the policy on the observation, predicted trajectories, and optional proprioception.
The interface is compact and transferable, but it inherits tracking sensitivity (occlusion, fast motion); because ATM samples query points uniformly at inference, many queried tracks can be uninformative (e.g., static background or irrelevant objects), which dilutes the motion signal available to the policy and motivates later methods that prioritize object- or semantics-relevant points.

\paragraph{Im2Flow2Act~\citep{Im2Flow2Act_Xu_CoRL_2024}.}
Im2Flow2Act uses \emph{object-centric 2D point flow} as the explicit interface: it predicts how points on the manipulated object should move in image space over a long horizon, excluding background and embodiment motion so that the same flow target can transfer across domains (human$\rightarrow$robot; sim$\rightarrow$real).
A language-conditioned flow generator trained on action-free human videos produces object-point trajectories from an initial frame and task description.
Grounding is provided by a separate flow-conditioned policy trained entirely in simulation; at deployment it receives updated object point locations from an online tracker and operates in closed loop to reduce mismatch to the predicted flow.
Because the control loop is driven by detected/tracked object points, performance depends on reliable object localization and point tracking; tracking drift or occlusion can directly corrupt the interface. While object-centric flow is designed to reduce sim-to-real appearance mismatch, performance can still degrade when real interactions induce object motions that diverge from the predicted image-space flow (e.g., contact-induced slippage or compliance not reflected in the predicted motion target).

\paragraph{Track2Act~\citep{track2act_ECCV_2024}.}
Track2Act starts from the \emph{pixel-trajectory} interface of the preceding methods but uses predicted tracks primarily as an intermediate signal to recover a \emph{rigid SE(3) plan} for execution.
Given an initial and goal image, it predicts short-horizon tracks for sampled query pixels and filters for salient motion before back-projecting the selected points to 3D using the initial depth.
A per-timestep rigid transform is then fit so that the projected 3D points match the predicted tracks (PnP-style alignment~\citep{EPnP_IJCV_2009, RANSAC}), yielding an object pose trajectory that the robot follows after grasping.
A lightweight goal-conditioned residual policy provides closed-loop corrections, but reliability hinges on track quality and depth-based lifting: occlusion, identity drift, or depth noise can perturb the recovered SE(3) trajectory.

\paragraph{Other extensions.}
Beyond the core 2D pixel-trajectory designs above, Tra-MoE~\citep{tra_moe_CVPR2025} extends the ATM-style any-point trajectory interface to more heterogeneous video domains by using sparsely-gated Mixture-of-Experts layers~\citep{MoE_2017} in the trajectory predictor and adaptive conditioning between predicted trajectory masks and observations.
We treat it as an extension rather than a core taxonomy anchor because it scales and specializes the trajectory predictor without introducing a new interface type or grounding role.

\subsubsection{3D/6D structured trajectory interfaces.}
\label{subsec:3d6d-trajectory-interfaces}

A third trajectory family predicts geometric motion targets directly---3D point trajectories, sparse semantic 3D keypoint trajectories, or 6D pose sequences---and uses these as the explicit interface for execution.
Compared with 2D pixel-track interfaces, these representations make metric structure explicit and reduce ambiguity when retargeting across embodiments, enabling controllers that track targets in 3D or SE(3) space.
The trade-off is heavier reliance on the 3D perception stack (depth, segmentation, pose estimation, or calibrated reconstruction): as Table~\ref{tab:explicit-interface-capabilities} shows, most methods in this cluster require RGB-D sensing or calibrated multi-view capture, and errors in these perception components produce incorrect motion targets that downstream controllers may faithfully track.

\paragraph{GeneralFlow~\citep{GeneralFlow_CoRL_2024}.}
GeneralFlow uses language-conditioned \emph{3D object-point trajectories} as the explicit interface: given an RGB-D observation and an instruction, it predicts short-horizon 3D trajectories of queried object points, specifying desired object motion rather than robot actions.
The interface predictor is trained from cross-embodiment human RGB-D video (e.g., HOI4D~\citep{HOI4D_CVPR_2022}) by deriving 3D trajectory supervision from object masks and depth-based back-projection, enabling learning without robot action labels.
For grounding, GeneralFlow executes with a geometric tracking controller: points near the gripper are tracked online and matched to their predicted targets, and an SVD-based alignment step produces SE(3) end-effector updates~\citep{ICP_1987} in a closed loop, enabling zero-shot execution without robot-domain training.
A practical failure mode is that segmentation/point-tracking errors or partial occlusion corrupt the correspondence set used by the alignment step, which can yield incorrect SE(3) updates even when the predicted 3D targets are plausible.

\paragraph{SKIL-H~\citep{SKIL_RSS_2025}.}
\label{subsubsec:skil_h}
Where GeneralFlow queries dense object-point motion, SKIL-H compresses the explicit interface to \emph{semantic 3D keypoint trajectories}: it discovers a small set of consistent object keypoints, predicts how they should move under an instruction, and trains a downstream policy to track the resulting 3D targets in closed loop.
The keypoints are obtained by clustering foundation-model features within segmented object regions, localized by descriptor matching at runtime, and lifted to 3D using depth and camera intrinsics; action-free human video then supervises a short-horizon predictor over these semantic keypoints.
This yields a more structured and interpretable 3D interface than uniform point flow, but it shifts robustness onto \emph{keypoint identity}: if semantic matching fails under viewpoint change, occlusion, or novel object appearance, the sparse target can jump abruptly, and the controller may faithfully track an incorrect 3D trajectory with little redundancy to average out the error.

Whereas the preceding methods predict where \emph{objects} should move, the next two use \emph{human hand/wrist motion} as the explicit interface and rely on retargeting to bridge the embodiment gap.

\paragraph{MimicPlay~\citep{MimicPlay_CoRL_2023}.}
MimicPlay uses \emph{3D human hand trajectories} as the explicit interface: from calibrated multi-view ``human play'' recordings it reconstructs 3D hand trajectories and treats them as motion targets that downstream robot policies should realize.
A goal-conditioned planner maps the current observation and a goal image to a compact plan code that is decodable into a distribution over future 3D hand trajectories, making the interface inspectable despite its low-dimensional representation.
For grounding, the frozen planner conditions a low-level robot policy trained from a small set of teleoperated demonstrations, together with wrist-camera features and proprioception.
This interface is compact and interpretable through trajectory decoding, but it relies on calibrated capture and accurate 3D hand reconstruction; downstream transfer ultimately depends on how well the human-play trajectory prior matches the robot’s visual observations and interaction dynamics.

\begin{table*}[t]
  \centering
  \footnotesize
  \setlength{\tabcolsep}{4pt}
  \renewcommand{\arraystretch}{1.25}
  \caption{\textbf{Reported quantitative snapshots for explicit visual interface methods.} These results are \emph{not} a leaderboard: papers differ in data regimes, sensors, and evaluation protocols. We use the numbers only to indicate evidence type and maturity---within-paper claims, weakly comparable shared-benchmark slices, and deployment-feasibility evidence. Entries are reported as stated in the original papers; ``Source'' gives the original location.} 
  \label{tab:explicit_interface_quant}
  \begin{tabularx}{\textwidth}{
      @{}
      >{\raggedright\arraybackslash}p{3.2cm}
      Y
      >{\raggedright\arraybackslash}p{5.0cm}
      >{\raggedright\arraybackslash}p{1.7cm}
      @{}
  }
    \toprule
    \textbf{Method} & \textbf{Metric (as reported)} & \textbf{Setting / Protocol Note} & \textbf{Source} \\
    \midrule

    \rowcolor{gray!8}
    \multicolumn{4}{l}{\textit{CALVIN} \citep{CALVIN_2022}: \textit{long-horizon language-conditioned (ABC\,$\rightarrow$\,D)}} \\
    \addlinespace[2pt]
    UniPi$^{\dagger}$~\citep{UniPi_Du_NIPS_2023} &
    SR@1--5$^{\ddagger}$ (\%): 56\,/\,16\,/\,08\,/\,08\,/\,04 &
    Train on 100\% ABC, test on D&
    SuSIE Table~1 \\
    \addlinespace[2pt]

    SuSIE~\citep{susie_2023} &
    SR@1--5 (\%): 87\,/\,69\,/\,49\,/\,38\,/\,26 &
    Train on 100\% ABC, test on D&
    Table~1 \\
    \addlinespace[2pt]

    CLOVER~\citep{CLOVER_2024} &
    SR@1--5 (\%): 96\,/\,84\,/\,71\,/\,58\,/\,45; Avg.Len$^{\S}$: 3.53 &
    Train on 100\% ABC, test on D&
    Table~1 \\

    \addlinespace[4pt]
    \midrule
    \rowcolor{gray!8}
    \multicolumn{4}{l}{\textit{LIBERO} \citep{LIBERO_Liu_2023}: \textit{multi-suite manipulation (Spatial / Object / Goal / Long)}} \\
    \addlinespace[2pt]
    UniPi$^{\dagger}$~\citep{UniPi_Du_NIPS_2023} &
    SR (\%) over Sp/Ob/Go/Lo: 69.2\,/\,59.8\,/\,11.8\,/\,5.8 &
    Reported by ATM; 10 demos + 50 action-free videos per task &
    ATM Table~V \\
    \addlinespace[2pt]

    ATM~\citep{ATM_Wen_RSS_2024} &
    SR (\%) over Sp/Ob/Go/Lo: 68.5\,/\,68.0\,/\,77.8\,/\,39.3; L-90: {48.4} &
    10 demos + 50 action-free videos/task &
    Table~VI \\
    \addlinespace[2pt]

    GVF-TAPE~\citep{GVFTAPE_CoRL_2025} &
    SR (\%) over Sp/Ob/Go: {95.5}\,/\,86.7\,/\,66.8; mean (3 suites): 83.0 &
    LIBERO-Long not reported; mean over Sp/Ob/Go only &
    Table~5 \\

    \addlinespace[4pt]
    \midrule
    \rowcolor{gray!8}
    \multicolumn{4}{l}{\textit{Other evaluations (not shared across papers)}} \\
    \addlinespace[2pt]
    AVDC~\citep{AVDC_2023} &
    Meta-World (10 tasks) avg.\ SR (\%): {43.1} &
    3 camera poses $\times$ 25 trials per task for evaluation (sim only) &
    Table~1 \\
    \addlinespace[2pt]

    GeneralFlow$^{\star}$~\citep{GeneralFlow_CoRL_2024} &
    Real robot (18 tasks, 6 scenes) avg.\ SR (\%): {81} &
    Zero-shot human-to-robot transfer; no robot-domain training &
    Table~2 \\
    \addlinespace[2pt]

    ZeroMimic$^{\star}$~\citep{ZeroMimic_ICRA_2025} &
    Real robot Avg. SR (\%): {71.9} (Franka, 9 skills), {65.0} (WidowX, 4 skills) &
    Zero-shot human-to-robot transfer from EpicKitchens dataset; 34 scenarios, 18 obj.\ categories &
    Fig.~5 \\

    \bottomrule
  \end{tabularx}

  \vspace{2pt}
  \raggedright
  \scriptsize
  $^{\dagger}$Reported by another paper (see source column).
  $^{\ddagger}$Success Rate (SR) for different No. of chained instruction.
  $^{\star}$Zero-shot real-robot evaluation (no robot-domain training).
  $^{\S}$Average task completion length.
  \textit{Sp\,/\,Ob\,/\,Go\,/\,Lo}: LIBERO Spatial\,/\,Object\,/\,Goal\,/\,Long suites; \textit{L-90}: LIBERO-90.

\end{table*}

\paragraph{ZeroMimic~\citep{ZeroMimic_ICRA_2025}.}
ZeroMimic uses human wrist SE(3) motion as an explicit \emph{post-grasp} interface, separating object acquisition from subsequent motion transfer.
Given the current observation and a goal image, it predicts short chunks of 6D wrist-pose trajectories in the camera frame and retargets them to robot end-effector motion for execution.
The interface is distilled from action-free egocentric human video by reconstructing hand motion and retaining only the wrist SE(3) signal as the transferable target, rather than attempting to imitate full human actions.
At deployment, a human-affordance-guided grasping stage first acquires the object; the learned post-grasp policy then outputs wrist-pose chunks that are transformed into the robot frame and tracked.
This decomposition makes the post-grasp motion interface compact and transferable, but success still depends on both stages: failed grasp acquisition, calibration/viewpoint mismatch, or wrist-pose reconstruction errors can directly distort the executed trajectory.

\paragraph{Interface-transfer links to 6D pose control.}
Several methods in this section ultimately execute a pose-sequence controller, even when their primary predicted interface is a video plan or 2D tracks.
A common interface-transfer pattern is to convert the predicted visual target into an object- or end-effector SE(3) sequence for downstream tracking: AVDC~\citep{AVDC_2023} and RIGVid~\citep{RIGVid_patel2025} transfer a generated video plan into an object-pose trajectory; Dreamitate~\citep{Dreamitate_Liang_CoRL_2024} transfers stereo video predictions into tool 6D poses; GVF-TAPE~\citep{GVFTAPE_CoRL_2025} extracts end-effector poses from predicted RGB-D future observations; Dream2Flow~\citep{dream2flow_dharmarajan2025} lifts generated object motion into 3D motion targets; and Track2Act~\citep{track2act_ECCV_2024} lifts 2D tracks into rigid object poses.
In our taxonomy, we group these methods by the interface predicted by the video/trajectory module (video plan, subgoal image, or 2D tracks), and treat pose conversion as a downstream grounding choice that trades holistic visual guidance for a compact geometric target.

RT-Affordance (RT-A)~\citep{Rt-affordance_ICRA_2025} is closely related in that it conditions a policy on an explicit end-effector pose sequence, but it relies primarily on robot trajectory supervision rather than action-free video, and is therefore outside this section's scope.

\paragraph{Takeaways for trajectory-based interfaces.}
Across trajectory-based methods, the shared idea is to expose temporal structure directly as a motion target---where points, objects, hands, or end-effectors should move over time---rather than as predicted images (Table~\ref{tab:explicit-interface-comparison}).
The main design axis is where geometric commitment enters the control pipeline.
Contact-centric affordance interfaces specify local interaction points and post-contact motion; 2D trajectory interfaces remain in image space, which makes them easy to pretrain from action-free human or Internet video but leaves metric structure and action grounding to downstream control; and 3D/6D interfaces expose more directly trackable geometric targets, but depend more heavily on depth, calibration, pose estimation, correspondence, or online localization.
This placement of the grounding burden also explains why many trajectory-based methods favor closed-loop execution: the target is compact but perception-sensitive, so systems often re-localize and correct during control rather than treating the trajectory as a one-shot plan.
Accordingly, Table~\ref{tab:explicit_interface_quant} summarizes reported evidence and evaluation settings rather than ranking 2D against 3D/6D interfaces, because the results come from a fragmented mix of shared suites, custom simulations, and real-robot studies.
The 2D-versus-3D/6D distinction should therefore be read primarily as a design trade-off in where grounding effort is spent, not as a settled empirical ordering.

\subsection{Execution and control integration}
\label{subsec:explicit-control-integration}

Explicit visual interfaces impose a clear two-level control hierarchy: a video-pretrained predictor produces the interface (subgoal, plan, trajectory), and a separate controller maps it to motor commands.
This modularity is a strength---it enables inspection, transfer, and independent improvement---but it also introduces control-integration challenges that are distinct from those faced by direct or latent-action methods.

\paragraph{The tracking-error problem.}
When an explicit interface specifies a target (e.g., a subgoal image, a 6D pose trajectory, or a set of point tracks), the low-level controller must \emph{track} that target under the robot's physical constraints---a problem with direct analogues in classical trajectory tracking, inverse kinematics, and operational-space control.
The failure modes are well understood: kinematic singularities can make smooth tracking impossible; the predicted target may lie outside the robot's reachable workspace; self-collision constraints may block the planned motion; and for non-rigid or high-dimensional interfaces (e.g., object flows or dense point tracks), the downstream controller faces an underdetermined mapping from visual targets to robot motion, whether that mapping is handled explicitly by task-space control or implicitly by a learned policy.
For instance, GeneralFlow's SVD-based SE(3) alignment may request end-effector displacements that violate joint limits, and ZeroMimic's 6D pose chunks can place the wrist near singularity boundaries where small target changes demand large joint-space motions.
Video models that produce visually plausible predictions may still generate targets that require the robot to pass through itself, reach beyond its workspace envelope, or exceed actuator velocity limits---yet feasibility checks (reachability, collision-freeness, joint-limit compliance) are rarely integrated into current systems.

\paragraph{Open-loop plans vs.\ closed-loop tracking.}
Explicit-interface methods span a wide range of execution strategies that mirror a classical spectrum from feedforward trajectory following to closed-loop visual servoing.
Some execute a predicted plan open-loop (UniPi, Dreamitate): the video model generates a full sequence, and the robot follows it without re-observing---analogous to feedforward execution of a pre-planned trajectory, where robustness depends entirely on prediction accuracy.
Others incorporate closed-loop feedback at various granularities: SuSIE and CLOVER replan subgoals when execution error exceeds a threshold; Im2Flow2Act and GeneralFlow track predicted targets in closed loop using online object detection; Track2Act adds a residual correction policy.
These closed-loop designs resemble look-and-move visual servoing in spirit, with the predicted interface serving as the reference signal, although in some methods (e.g., Im2Flow2Act) the correction law is learned rather than explicitly geometric.
Closed-loop execution mitigates compounding prediction errors but requires that the interface can be refreshed often enough during execution, which is especially expensive for video-generation-based interfaces.
A practical middle ground adopted by several methods is \emph{periodic replanning}: execute for a short horizon, then re-predict and switch to the updated interface.

\paragraph{Interface-transfer pipelines as fragile links.}
Many methods in this family include an interface-transfer step that converts the primary predicted signal into a lower-dimensional control target---for example, video plans or 2D tracks into object, tool, or end-effector pose sequences (enumerated above).
Each step in these transfer pipelines---segmentation, tracking, depth estimation, correspondence matching, rigid-body fitting---can fail under clutter, occlusion, specularities, or deformable objects.
This mirrors a well-known concern in classical cascaded estimation-and-control pipelines, where each perception stage introduces errors that propagate downstream; the difference is that classical systems can often cross-check intermediate estimates against known dynamics or geometric models, whereas video-derived pipelines typically lack such verification.
Crucially, errors \emph{compound}: a small segmentation error propagates through depth lifting, through pose estimation, and into the controller, which ``faithfully'' tracks an incorrect target.
Making these pipelines robust---or designing interfaces that reduce the depth of the transfer pipeline (e.g., direct 2D trajectory prediction in ATM, or direct 3D motion prediction in GeneralFlow)---is a practical priority for deployment.

\paragraph{Hallucinated physics in generated plans.}
A risk shared with direct video--action models but especially visible here is \emph{physical hallucination} in generated video plans.
Generative video models, trained to maximize visual likelihood rather than physical plausibility, may produce subgoal images or video sequences that violate contact mechanics (e.g., objects sliding before contact), geometry (object penetration, disappearing parts), or dynamics (instantaneous accelerations, gravity-defying motion)---failures that are absent by construction in classical physics-based planning, which operates on dynamics models that enforce physical constraints.
This risk is most acute for dense video-plan methods (UniPi, AVDC, Dreamitate, Dream2Flow) and subgoal-image methods (SuSIE, CLOVER) that rely on generated visual predictions as the primary interface; methods that predict compact motion targets directly rather than rendering a full visual rollout (e.g., ATM, GeneralFlow) partially sidestep this problem, though their predicted targets can still be unrealistic when the trajectory predictor extrapolates beyond its training distribution.
Methods that filter generated candidates for plausibility or instruction consistency (e.g., RIGVid's VLM-based rollout selection) partially mitigate the problem, but systematic physical-consistency checking---analogous to constraint satisfaction in motion planning---remains an open problem.

These failure modes---tracking difficulty, open-loop fragility, transfer-pipeline compounding, and hallucinated physics---represent the primary execution-level risks of explicit-interface designs, and each method's interface choice determines which of these risks dominates.

\subsection{Summary and takeaways}
\label{sec:explicit-interfaces-summary}

Explicit visual interface methods factor video understanding into a predict--then--control pipeline: a video-pretrained predictor produces a structured, inspectable intermediate target---either a future visual target or a visually grounded motion target---and a separate controller grounds that target to robot actions.
Across this section, the main design variation is not simply which network is used, but what temporal structure is exposed explicitly and where the grounding burden sits.
Figure~\ref{fig:explicit_interface_pipeline} and Tables~\ref{tab:explicit-interface-comparison}--\ref{tab:explicit-interface-capabilities} summarize the pipeline, structural choices, training sources, and real-robot scope.

\paragraph{How explicit interfaces transfer temporal structure to control.}
The central design choice is the division of labor between the interface predictor and the downstream controller: each interface type determines what structure is provided explicitly and what the controller must recover on its own.
Goal images and video plans expose future visual states, whereas trajectory-based interfaces expose motion targets such as affordance waypoints, pixel tracks, keypoint trajectories, or pose sequences.
These interfaces also differ in how much temporal structure they make explicit---from sparse subgoals, to dense visual rollouts, to compact motion targets---and in whether execution relies primarily on learned goal-reaching, inverse dynamics, geometric transfer, or repeated replanning.
Seen this way, explicit-interface methods are best understood not as a single technique class, but as a family of design choices about how video-derived structure is handed off to control.

\paragraph{Empirical evidence and evaluation fragmentation.}
Tables~\ref{tab:explicit-interface-capabilities} and~\ref{tab:explicit_interface_quant} indicate that the predict--then--control factorization is already practically useful: many methods learn interface predictors from action-free video, ground them with comparatively small robot datasets, and report cross-embodiment transfer or real-robot deployment.
The evaluation picture, however, remains fragmented.
Table~\ref{tab:explicit_interface_quant} captures a few shared benchmarks such as CALVIN and LIBERO, but most methods are evaluated on non-comparable custom suites, and interface design, video-generation quality, grounding pathway, and execution strategy interact strongly.
The reported results are therefore most useful for identifying recurring design patterns and failure modes, rather than for establishing a stable cross-method ranking.

\paragraph{Design patterns and failure modes.}
The main trade-off in this family is that making temporal structure explicit improves inspectability and modularity, but shifts difficulty into the interface--control handoff.
Each interface relocates the grounding problem rather than removing it: dense video plans stress visual realism and transfer; subgoal images push more of the missing temporal structure into the controller; image-space trajectories defer metric grounding; and 3D/6D targets depend more heavily on geometric perception.
The resulting failure modes differ in form but share the same underlying cause: the explicit interface is useful only insofar as it can be grounded robustly under real sensing, estimation, and execution constraints.
In this sense, explicit interfaces do not eliminate the grounding problem; they make it visible, modular, and therefore easier to analyze, but still difficult to solve robustly.

\paragraph{Relation to direct and latent methods.}
Relative to the earlier two families, explicit interfaces occupy the most interpretable point in the design space.
Direct video--action policies keep video-to-control coupling implicit in a single policy, and latent-action methods introduce a dedicated but opaque intermediate variable; by contrast, explicit-interface methods expose a structured target that can be visualized, debugged, and in some cases transferred across embodiments more naturally.
This makes them especially attractive when action-free human, Internet, or robot video is abundant but robot action data is limited.
Their central challenge, however, is precisely that interpretability and modularity come with a visible interface that must still be grounded reliably.
The long-term promise of this family therefore lies not only in predicting better interfaces, but in designing interfaces whose structure matches what downstream control can robustly execute.
We return to these cross-cutting issues---including controllability, temporal abstraction, grounding protocols, and evaluation standards---in Section~\ref{sec:discussion_future}.

\section{Datasets for Video-Based Robotic Manipulation}
\label{sec:datasets}

Video-based manipulation methods typically draw on two complementary supervision sources:
(i) \textbf{action-free video} (often human, sometimes robot) used to learn temporal structure and predictive priors, and
(ii) \textbf{action-labeled robot trajectories} used to ground video-derived structure into executable control and to evaluate policies under standardized protocols.
This section provides a concise reference to the datasets and benchmarks most commonly used in this literature. Method-level empirical comparisons are presented in the family sections, while broader evaluation issues are revisited in the cross-family discussion.
We use ``dataset'' broadly to include resources used for pretraining, grounding, or evaluation; several widely adopted datasets (e.g., CALVIN~\citep{CALVIN_2022}, LIBERO~\citep{LIBERO_Liu_2023}, RLBench~\citep{james2020rlbench}) also function as benchmarks with fixed task suites and metrics.

\paragraph{A practical taxonomy.}
For survey purposes, datasets can be categorized by:
(i) \textbf{embodiment} (human vs.\ robot),
(ii) \textbf{labels} (action-free vs.\ action-labeled; auxiliary labels such as poses, masks, language),
(iii) \textbf{modality} (RGB, RGB-D, multi-view, force/audio),
and (iv) \textbf{evaluation role} (pretraining corpora vs.\ standardized benchmarks).

\subsection{Human Manipulation Video Datasets}
\label{subsec:human_video_datasets}

Human manipulation videos provide large-scale, mostly action-free recordings of object interactions and task execution.
They are widely used to pretrain video predictors, learn motion/affordance priors, or distill intermediate interfaces (e.g., subgoals, trajectories) before grounding on robot data.

\paragraph{Egocentric, in-the-wild.}
Datasets such as Ego4D~\citep{Ego4d_CVPR_2022} and EPIC-Kitchens~\citep{EPIC-KITCHENS-100_IJCV_2022} provide large-scale first-person video with rich hand--object interactions and long-horizon activities. These are attractive for learning manipulation-relevant temporal structure, but include substantial confounds (camera motion, occlusions, scene diversity) that can complicate grounding.

\paragraph{Multi-view / RGB-D with stronger geometry.}
Datasets such as HOI4D~\citep{HOI4D_CVPR_2022}, H2O~\citep{H2O_Kwon_2021_ICCV}, and DexYCB~\citep{DexYCB_chao2021_CVPR} provide richer geometric cues (e.g., RGB-D, multi-view capture, hand/object pose annotations), which can be particularly helpful for interface-transfer methods that lift 2D motion into 3D trajectories or poses.

\paragraph{Web-scale and generic video corpora (often used as pretraining).}
Large generic video--language resources such as HowTo100M~\citep{howto100m_ICCV_2019} and interaction-focused datasets such as Something-Something V2~\citep{somethingSomethingV2_2017} are frequently used to pretrain video or video--language backbones that are later adapted to manipulation. These corpora provide scale and diversity, but they are less targeted to manipulation geometry and often underrepresent fine contact dynamics.

Overall, human video datasets trade off \emph{scale and diversity} against \emph{geometric reliability and controllability}. This trade-off aligns closely with the method families: approaches that rely on interface transfer often benefit disproportionately from reliable geometry (multi-view, RGB-D, pose/mask cues), while approaches that learn dynamics priors implicitly can benefit more from scale.

\subsection{Robotic Manipulation Datasets and Benchmarks}
\label{subsec:robot_datasets}

Robot datasets supplement video with action labels (and often language or proprioception), providing the supervision needed to connect video-derived representations to executable control.

\subsubsection{Real-world robot datasets.}
Large-scale real-robot datasets are commonly used for grounding and for training generalist policies.
RT-1~\citep{RT1_Brohan_2022} reports learning from 130k real-world episodes collected across 13 robots and 700+ tasks, while Open X-Embodiment (OXE)~\citep{OXE_ICRA_2024} aggregates over one million trajectories across 22 robot embodiments (527 skills) to study cross-embodiment generalization.
BridgeData V2~\citep{Bridgedata_Walke_CoRL_2023} supports scalable collection across varied scenes and is frequently used for imitation and diffusion-policy training.
DROID~\citep{khazatsky2024droid} emphasizes large-scale data collection across diverse real environments (reported across many buildings/cities/countries), enabling stress tests under substantial environment shift.

Some datasets target sensing beyond RGB for contact-rich manipulation. For instance, RH20T~\citep{fang2023rh20t} provides multimodal real-robot trajectories (including force/torque), which can be valuable when downstream controllers must realize video-derived targets under tight physical constraints.

\subsubsection{Simulation datasets and standardized benchmarks.}
Simulation remains essential for controlled comparisons, reproducibility, and stress testing.
CALVIN~\citep{CALVIN_2022} is a widely used language-conditioned long-horizon benchmark with standardized evaluation over multi-step sequences.
LIBERO~\citep{LIBERO_Liu_2023} and RLBench~\citep{james2020rlbench} provide broad task suites under unified APIs and are commonly used for transfer and robustness evaluation.
COLOSSEUM~\citep{pumacay2024colosseum} is a robustness-focused benchmark designed to quantify sensitivity to distribution shift.
Tooling-oriented resources such as RoboMimic~\citep{mandlekar2021robomimic} provide open-source datasets and reproducible learning pipelines for offline imitation / offline RL from demonstrations, while simulation platforms such as ManiSkill2~\citep{gu2023maniskill2} support scalable experimentation and benchmark-style task suites.

\subsection{How Datasets Map to Method Families}
\label{subsec:dataset_usage_across_families}

Dataset usage aligns closely with where each family places the video-to-control interface.
\textbf{Direct video--action policies} keep the learned interface close to the robot's action space, so action-labeled robot trajectories remain central both for grounding and for evaluating whether video-shaped representations translate into executable low-level behavior.
\textbf{Latent-action methods} and \textbf{explicit visual interfaces} can exploit action-free video more aggressively, because they first learn transition structure or an intermediate interface from video and only later connect it to robot control through decoders, inverse dynamics, goal-reaching controllers, or downstream optimization.
Across all families, simulation benchmarks remain critical for controlled ablations and robustness tests, while real-world datasets provide the more demanding measure of whether the learned temporal structure survives grounding under sensing, actuation, and environment shift.

Evaluation protocols vary substantially across and within families: methods differ not only in benchmark choice but also in pretraining corpora, robot-data budgets, observation modalities, and reported metrics.
Per-family quantitative snapshots, with comparability caveats, are provided in \S\ref{sec:direct-video-action}--\S\ref{sec:visual-interfaces}; \S\ref{sec:discussion_future} discusses what current benchmarks do not yet test and what evaluation infrastructure the field still needs.
\section{Cross-Family Synthesis, Deployment Challenges, and Future Directions}
\label{sec:discussion_future}

The preceding sections analyzed three families of video-based manipulation learning---\emph{direct video--action policies} (\S\ref{sec:direct-video-action}), \emph{latent-action methods} (\S\ref{sec:latent_actions}), and \emph{explicit visual interfaces} (\S\ref{sec:visual-interfaces})---each with per-family execution and control-integration analyses.
This section shifts to cross-family comparison, synthesizing shared design axes, examining what the families' differences imply for control-loop integration and deployment, and identifying open challenges and future directions.

\vspace{4pt}
\noindent\fcolorbox{black!45}{gray!8}{%
\begin{minipage}{\dimexpr\linewidth-2\fboxsep-2\fboxrule\relax}
\small
\textbf{The robotics integration layer.} The deployment-facing stack that turns video-derived predictions into closed-loop robot behavior: grounding interfaces to executable actions, closing the control loop at an adequate rate, checking physical feasibility, and recovering from detected errors, all under a specific robot's embodiment, sensing, and actuation constraints.
\end{minipage}%
}
\vspace{4pt}

\begin{table*}[t]
\centering
\caption{\textbf{Family-level comparison across three design philosophies.}
Direct video--action models keep the video--action link implicit in shared representations; latent-action methods introduce an intermediate action abstraction that may be used at deployment or as a training signal; explicit visual-interface methods expose structured, human-interpretable targets for a downstream controller.}
\label{tab:family_level_comparison}
\footnotesize
\setlength{\tabcolsep}{5pt}
\renewcommand{\arraystretch}{1.35}
\begin{tabularx}{\textwidth}{@{}>{\raggedright\arraybackslash}p{2.2cm}YYY@{}}
\toprule
\textbf{Axis} &
\textbf{Direct Video--Action} &
\textbf{Latent-Action} &
\textbf{Explicit Visual Interfaces} \\
\midrule

\rowcolor{gray!5}
\textbf{What action-free video supervises} &
Shared dynamics-aware \emph{representations} via temporal prediction &
A bottleneck variable (latent action) summarizing \emph{transition-causing change} &
A \emph{visualizable interface} (subgoal image / video plan / traj.\ / poses) \\
\addlinespace[6pt]

\textbf{Interface at deployment} &
None exposed; link lives in hidden features and the action head &
Abstract latent-action interface; may be exposed for planning/control or retained mainly as a training signal &
Explicit and inspectable: plans, subgoals, tracks, flows, poses \\
\addlinespace[6pt]

\rowcolor{gray!5}
\textbf{Action production} &
Predict actions directly in native control space &
Ground latent actions to robot actions, or transfer their supervision into a native-action policy &
Map interface to robot actions via controller, optionally after interface transfer \\
\addlinespace[6pt]

\textbf{Typical training factorization} &
Joint/interleaved video+action on mixed data; boundary: action-free pretrain then RL grounding &
Stage~1: discover latent actions from obs.-only data; Stage~2: grounding, policy training, or head transfer with limited action-labeled data &
Predictor from action-free video; controller trained separately (demos, sim, RL, or zero-shot geometry) \\
\addlinespace[6pt]

\rowcolor{gray!5}
\textbf{Inference-time requirements} &
Often bypasses video gen.; some models can still produce rollouts &
Usually no video gen.; infer/select latent actions, decode persistent tokens/skills, or output native actions &
Depends on interface: may need generative prediction + post-processing; or lightweight predictors; execution via tracking/controller \\
\addlinespace[5pt]
\midrule
\addlinespace[2pt]

\textbf{Benefits} &
End-to-end in native action space; simple interface; scales with mixed data; efficient when video bypassed &
Modular discovery vs.\ grounding; compact action-like interface or supervision signal; reduced action supervision &
Transparency / debuggability; natural cross-embodiment transfer; predictor/controller improve independently; interface transfer makes embodiment-agnostic targets explicit \\
\addlinespace[6pt]

\rowcolor{gray!5}
\textbf{Costs} &
Hard to inspect/debug; implicit video-to-control linkage; one-to-many futures and exogenous dynamics blur controllability &
Latents may entangle confounders with controllable effects; identifiability not guaranteed; grounding or head-transfer mismatch can be brittle &
Interface reliability often dominates (prediction realism + perception/transfer failures); embodiment feasibility may be violated; multi-stage calibration burden \\

\bottomrule
\end{tabularx}

\vspace{2pt}
\raggedright
\scriptsize
\textit{RL} = reinforcement learning; \textit{obs.} = observation; \textit{gen.} = generation; \textit{traj.} = trajectories.
\end{table*}

Table~\ref{tab:family_level_comparison} provides the cross-family lens, contrasting interface design, grounding mechanisms, execution structure, and the resulting benefits and costs.
Its rows show that the three families differ less in model architecture than in three coupled design choices: how explicitly the video-to-control interface is represented, where grounding to robot action occurs, and what deployment burdens follow.
More explicit interfaces improve inspectability and cross-embodiment transfer but introduce estimation stages whose errors can dominate execution; more implicit designs reduce design bias and simplify deployment at the cost of reduced transparency and fewer opportunities for verification and intervention.
The subsections below unpack these trade-offs: \S\ref{subsec:synthesis} compares conceptual design axes, \S\ref{subsec:control-comparison} examines cross-family control properties and deployment challenges, and \S\ref{subsec:challenges-future} consolidates open problems and research directions.

\subsection{Cross-Family Synthesis}
\label{subsec:synthesis}

All three families aim to exploit large-scale video without action labels to learn priors about interaction dynamics, while reducing reliance on expensive action-labeled robot trajectories.
They share this high-level goal but diverge in how they structure the path from video-derived knowledge to executable robot actions.

\paragraph{Interface location and explicitness.}
The clearest cross-family distinction is where the video-to-control interface lives and how visible it remains at deployment.
Direct video--action models keep this interface implicit inside shared representations and action heads, which simplifies the deployment stack but makes it difficult to inspect why a particular action is produced.
Latent-action methods learn a structured intermediate variable from observation transitions, which may be retained as a planning/control interface or used mainly as action-centric supervision before native-action fine-tuning; in both cases, its semantics are not guaranteed and may still entangle controllable and non-controllable change.
Explicit visual-interface methods move the interface into human-interpretable outputs---subgoal images, video plans, trajectories, or pose targets---which improves inspectability and debugging, but also introduces design bias and additional perception or transfer stages whose errors can dominate execution.
Within the explicit family, the same spectrum continues: dense video plans provide the richest context but are expensive and fragile, subgoal images simplify planning but underspecify motion, and trajectory or pose interfaces offer the most precise targets while depending heavily on tracking and geometric estimation.

\paragraph{Training factorization and what action-free supervision buys.}
A second cross-family axis is how video supervision is separated from robot-action grounding.
Direct video--action models typically optimize video and action objectives jointly or interleaved on mixed data, so action generation remains tightly coupled to the shared representation.
Latent-action and explicit-interface methods more often factor training into two stages: first learn a latent forward model or interface predictor from action-free video, then connect it to robot control using a smaller amount of action-labeled data, demonstrations, interaction, or downstream optimization.
This factorization can improve data efficiency and modularity, but it also reveals a recurring tension: a video-derived representation may be predictive without being realizable under the robot's kinematic, contact, or embodiment constraints.
In practice, what action-free video buys is therefore not direct executability, but a scalable prior over how scenes change---one that must still be grounded carefully to robot behavior.

\begin{table*}[t]
\centering
\caption{\textbf{Control-level comparison across the three families.}
Summary of loop closure, pre-execution verification, dominant control risk, and common mitigation strategies; \S\ref{subsec:control-comparison} unpacks each column.}
\label{tab:control_properties}
\footnotesize
\setlength{\tabcolsep}{5pt}
\renewcommand{\arraystretch}{1.35}
\begin{tabularx}{\textwidth}{@{}>{\raggedright\arraybackslash}p{1.9cm} >{\raggedright\arraybackslash}X >{\centering\arraybackslash}p{1.5cm} >{\raggedright\arraybackslash}X >{\raggedright\arraybackslash}X@{}}
\toprule
\textbf{Family} &
\textbf{Loop closure} &
\textbf{Pre-exec.\ verification} &
\textbf{Main control risk} &
\textbf{Common mitigation} \\
\midrule

\rowcolor{gray!5}
\textbf{Direct video--action} &
Native-action decoding (stepwise, chunked, or receding-horizon); no inspectable intermediate &
Low &
Shared representation may be visually predictive yet not controllable; feasibility enters through action outputs &
Robot-action grounding; frequent re-decoding to limit drift; bypass/amortize video generation for control rate; no explicit feasibility checks \\
\addlinespace[6pt]

\textbf{Latent-action} &
Latent planning/search (e.g., MPC over latents) or latent-conditioned policy; some variants output native actions &
Medium-low &
Codes may be predictive but not identifiable or executable; risk enters at latent prediction and grounding alignment &
Information bottlenecks, disentanglement/contrastive objectives, native-action fine-tuning; online monitoring of latent rollouts when retained \\
\addlinespace[6pt]

\rowcolor{gray!5}
\textbf{Explicit visual interfaces} &
Controller tracks a subgoal, trajectory, or pose target; open-loop, periodic replanning, or closed-loop tracking &
Medium-high &
Target may be visually plausible yet physically infeasible; risk enters through predicted visual/geometric targets and the transfer pipeline &
Closed-loop tracking and replanning, residual correction, candidate filtering; reachability/collision checks (still rarely integrated) \\

\bottomrule
\end{tabularx}

\vspace{2pt}
\raggedright
\scriptsize
\textit{MPC} = model predictive control; \textit{pre-exec.} = pre-execution.
\end{table*}

\paragraph{Temporal abstraction and planning horizon.}
A third cross-family axis is how interface design shapes what kind of temporal abstraction can be learned, planned over, and validated before execution.
Direct models absorb temporal structure into the policy or generator itself, so whatever planning they perform remains largely implicit: they can smooth or chunk behavior over short windows, but the resulting multi-step intent is difficult to inspect or recombine.
Latent-action methods are naturally abstraction-oriented: discrete or continuous transition latents can serve as compact units for search, policy learning, persistent-token decoding, co-generated latent streams, or skill construction.
Because many are discovered from short-horizon observation transitions, their usefulness depends on whether the learned semantics remain stable, controllable, and aligned with executable actions as they are grounded, composed, or carried into deployment.
Explicit interfaces provide the most transparent temporal targets: subgoals, video plans, and trajectories make future structure visible before action, and in practice current systems typically replan at short horizons rather than committing to long-range predictions, leaving longer-horizon temporal abstraction as an open opportunity.
Across all three families, the unresolved challenge is therefore not only to predict farther ahead, but to learn reusable multi-step structures whose semantics remain executable, physically grounded, and robust under feedback.

\paragraph{Generalization and transfer: what helps, what breaks.}
Cross-embodiment and cross-domain transfer are enabled by different mechanisms across the three families.
Direct models can benefit from broad multi-robot pretraining, but their action heads remain tightly tied to a specific embodiment.
Latent-action methods aim to separate transition structure from embodiment-specific execution through modular grounding, but learned latents can conflate multiple causes of visual change, making the grounding step sensitive to distribution shift.
Explicit interfaces can transfer more naturally when defined in observation or geometric space, but this advantage depends on reliable perception, viewpoint alignment, and whether the predicted targets remain feasible for the target robot.
Taken together, these differences suggest that transfer improves as the learned interface becomes less tied to a single action space, but robustness then depends more heavily on accurate grounding, estimation, and feasibility under the new embodiment.

\subsection{Cross-Family Control Properties and Deployment Challenges}
\label{subsec:control-comparison}

The three families differ not only in where they place the video-to-control interface, but also in how they close the control loop, where physical inconsistency enters, what can be verified before execution, and how readily they transfer across embodiments and domains.
The comparisons below synthesize these deployment-level differences, with Table~\ref{tab:control_properties} providing a compact overview before the following paragraphs discuss each dimension in turn.

\paragraph{Execution loop and replanning.}
The three families close the control loop in fundamentally different ways.
Direct video--action methods execute in the robot's native action space without an inspectable intermediate. As Section~\ref{subsec:direct-control-integration} showed, their execution mode---stepwise, chunked, receding-horizon, or feature-conditioned---is a deployment-level design choice orthogonal to training architecture, analogous to choosing between reactive control and open-loop trajectory segments in classical robotics.
Standalone latent-action methods route decisions through a compact latent space, supporting MPC-style search over latent actions (CLASP), so execution depends on both the latent forward model and the grounding map. Instruction-conditioned variants span a wider range: some use latent codes mainly as pretraining or co-training signals before outputting robot actions directly, while others retain persistent tokens, co-generated latent streams, or latent skills that continue to condition execution.
Explicit-interface methods span the widest range: some execute video plans open-loop (UniPi, Dreamitate), others replan subgoals adaptively (SuSIE, CLOVER), and trajectory-tracking methods operate in closed loop (Im2Flow2Act, GeneralFlow)---resembling the classical hierarchy from feedforward trajectory execution through look-and-move visual servoing to online trajectory tracking.
Across all families, the inference latency of large generative models can substantially limit how tightly the loop can be closed, favoring methods that bypass video generation at deployment (VidMan, VPP, UVA) or use lightweight interface predictors (ATM, SKIL-H, GeneralFlow).

\paragraph{Physical feasibility and consistency.}
Across all three families, predicted behavior is typically not accompanied by explicit guarantees of kinematic feasibility, collision avoidance, or contact consistency---a contrast with classical model-based planning, where such constraints can be enforced directly.
Direct methods encode these constraints only implicitly through the action-labeled training distribution, so violations appear as infeasible or brittle actions at execution time rather than being ruled out beforehand.
Latent-action methods face a related but distinct risk: the latent forward model used for discovery or planning may remain visually predictive while drifting away from controllable, dynamically valid transitions, and the grounding mechanism may then align executable actions to a physically imperfect latent space.
Explicit interfaces come closest to exposing physically meaningful targets, but their predictions are still produced by learned perceptual or generative models rather than by constrained motion planners, so they can remain visually plausible while being kinematically unreachable or dynamically inconsistent.
Thus, the central physical-consistency problem differs across families not in whether it exists, but in where it enters: in direct methods through action outputs, in latent methods through latent prediction and grounding alignment, and in explicit methods through predicted visual or geometric targets.

\paragraph{Failure detection, verification, and recovery.}
The families differ more sharply in what they expose for checking and correction than in whether failures occur.
Direct methods are the most opaque: because there is no explicit intermediate target, failures are easy to observe but difficult to diagnose, and there is little opportunity to verify the planned behavior or insert corrective logic before execution.
Latent-action methods provide somewhat more structure when the latent remains active at deployment, since predicted latent transitions, token choices, or rollout discrepancies can in principle be monitored online; when the latent is used mainly for pretraining, however, this checkpoint largely disappears. In both cases, the diagnostic signal remains indirect and depends on whether the latent space cleanly separates controllable from exogenous change.
Explicit interfaces provide the clearest checkpoint: discrepancies between a predicted subgoal, trajectory, or pose target and the observed state can be measured directly and used to trigger replanning or residual correction, as in CLOVER or Track2Act.
The practical consequence is that explicit interfaces are currently the most amenable to structured verification and recovery, whereas direct and latent methods still rely more heavily on implicit robustness of the learned controller than on explicit error-monitoring mechanisms.

\paragraph{Embodiment mismatch, domain gaps, and deployment practicality.}
Deployment becomes harder as video-derived structure must be grounded to a specific robot under embodiment, observation, and compute constraints.
Although transfer is generally easier when the interface is further from the robot's native action space, ``embodiment-agnostic'' predictions can remain embodiment-\emph{infeasible}: a human wrist trajectory may require joint configurations impossible for a 6-DoF arm, which is why methods such as ZeroMimic and GeneralFlow incorporate explicit retargeting or geometric alignment rather than assuming that a small robot dataset will close the gap automatically.
Video-based pretraining introduces a second deployment challenge distinct from classical sim-to-real transfer: Internet and egocentric videos differ from robot observations in viewpoint, resolution, lighting, and motion statistics, and explicit-interface methods often face an additional \emph{interface-domain gap} when predictors trained on human video output targets with human-hand kinematics.
Finally, all transfer benefits are contingent on the full pipeline meeting the control-rate requirements discussed above, which further favors lightweight or video-generation-free inference strategies.

\subsection{Open Challenges and Future Directions}
\label{subsec:challenges-future}

The cross-family analysis above reveals four clusters of open problems and corresponding research directions.

\paragraph{Execution-aware and physically grounded learning.}
A central unresolved question is what video prediction objectives actually constrain for control.
Future prediction from pixels is inherently one-to-many: for direct models, it remains unclear whether gains arise from better visual representations, regularization, stronger generative priors, or alignment of video and action prediction.
For latent-action and explicit-interface methods, a related gap is the tension between predictability and controllability---a latent transition, predicted subgoal, or visual/geometric interface may be visually plausible yet kinematically unreachable or not realizable by the robot.
Current video generation models can further compound this problem by maximizing visual likelihood without explicit physical constraints, yielding visually plausible but physically inconsistent contact or motion predictions.
Addressing these limitations calls for execution-aware learning at multiple levels: augmenting predictors with feasibility signals such as constraint violations or uncertainty estimates, incorporating lightweight physics priors into generation or decoding, coupling temporal-abstraction learning to execution feedback so that discovered primitives remain controllable rather than merely predictive, and exploring hybrid multi-resolution interfaces that pair high-level visual context with short-horizon targets that are easier to verify and execute.

\paragraph{Robust grounding and cross-embodiment transfer.}
Every video-based method must ultimately map visual predictions to robot actions under embodiment-specific constraints, yet no current grounding mechanism offers a principled, scalable solution across diverse robots, sensors, and tasks with minimal in-domain data.
This challenge has two intertwined facets.
The first is separating controllable from exogenous dynamics: action-free video mixes robot-caused change with camera motion, other agents, and environment-driven dynamics, and as pretraining corpora grow more diverse, disentangling controllable signals from correlated distractors becomes harder.
Possible handles include multi-view constraints, ego-motion compensation, counterfactual objectives, and multi-agent factorizations, but robust separation at scale remains unresolved.
The second facet is efficient adaptation across embodiments: current approaches range from latent-code grounding and learned inverse dynamics to retargeting and geometry-based controllers, but retraining or fine-tuning for each new robot remains the norm.
Promising directions include lightweight robot-specific adapters, shared action representations, and retargeting mechanisms that incorporate embodiment constraints explicitly, with the goal of reducing robot-domain data requirements without sacrificing robustness under distribution shift.

\paragraph{Multimodal sensing and contact-rich manipulation.}
Most current video-based methods operate in a vision-only setting and are evaluated primarily on rigid or quasi-rigid tasks, reflecting a shared limitation across all three families: current interfaces capture the visual consequences of interaction but not the underlying force and compliance state.
Extending to contact-rich assembly, deformable manipulation, and tasks requiring force modulation will likely demand interfaces that represent non-rigid state (dense 3D flow, keypoint fields) and controllers that exploit tactile and proprioceptive feedback alongside visual predictions.
Integrating tactile or force/torque sensing with video-derived interfaces could bridge this gap, enabling policies that specify both \emph{where} to move and \emph{how hard} to push.
Emerging multimodal robot datasets that include force channels (e.g., RH20T~\citep{fang2023rh20t}) provide a starting point, but learning joint video--tactile representations from heterogeneous data---where most video lacks force and most force data lacks diverse video---remains an open problem.
Curating video datasets with stronger coverage of deformable and contact-heavy interactions, and evaluating under realistic occlusion and clutter, is a concrete step toward broader applicability.

\paragraph{Evaluation, verification, and safe deployment.}
Reliable deployment requires progress on two connected fronts: trustworthy interface monitoring during execution and fair evaluation across families.
Failure signals and verification hooks differ sharply across the three families, but systematic monitoring remains underdeveloped: direct methods provide little natural checkpointing, latent methods offer only indirect rollout discrepancies, and explicit interfaces depend on intermediate estimation stages whose errors can compound under clutter, occlusion, or deformable interactions.
Addressing this reliability gap calls for lightweight verification modules that screen predicted interfaces or latent rollouts for feasibility, estimate confidence, and trigger replanning or safe fallback behaviors.
Fair comparison remains difficult because methods differ not only in benchmark choice, but also in pretraining corpora, robot-data scale, observation modalities, and evaluation procedures.
Standardized protocols should at least control for robot-data budget, observation modalities, and task difficulty, and should include metrics beyond success rate---such as robustness to perturbations, recovery behavior, and calibration of uncertainty---with widely adopted real-robot benchmarks, particularly for long-horizon and contact-rich tasks.
Relatedly, the \textit{--} entries in the Real-Robot Scope columns of the per-family capability tables (Tables~\ref{tab:direct-video-action-capabilities}, \ref{tab:latent-action-capabilities}, and~\ref{tab:explicit-interface-capabilities}) should be read not as missing bookkeeping, but as a sign of evaluation immaturity: several video-derived interfaces are still validated primarily as representation or simulation mechanisms rather than as robust robot-control components.
Together, monitoring, verification, and controlled evaluation form the infrastructure needed for video-based manipulation to move from laboratory demonstrations to dependable real-world deployment.

\subsection{Summary}

Action-free video is a powerful and scalable observation of world dynamics, but the path from predictive visual structure to reliable robotic manipulation remains incomplete.
Direct video--action models, latent-action methods, and explicit visual interfaces represent complementary points in the design space, trading off end-to-end simplicity, modular abstraction, and inspectable control targets, with distinct consequences for control-loop integration, physical feasibility, failure detection, and cross-embodiment transfer.

Our analysis suggests that the most pressing gaps now center on the \emph{robotics integration layer}: ensuring that video-derived predictions respect physical constraints, that control loops can be closed at adequate frequencies, that failures can be detected and recovered from, and that interfaces can be verified before execution.
Closing these gaps will require execution-aware and physically grounded learning, robust cross-embodiment grounding, multimodal sensing beyond vision alone, and evaluation infrastructure that enables fair comparison and safe deployment.
As video corpora and foundation models continue to scale, progress will depend increasingly on this integration layer---turning rich passive visual experience into robot behavior that is not only capable, but dependable.

\section{Conclusion}
\label{sec:conclusion}

Bridging the gap between abundant action-free video and reliable robotic control is not merely a representation-learning challenge: video-derived temporal structure must ultimately close a loop under real embodiment, sensing, and physical constraints.
This survey organized the growing literature through an interface-centric taxonomy and analyzed three families of methods through a consistent control-integration lens.
The analysis reveals that each family negotiates a distinct trade-off between end-to-end simplicity, inspectability, and cross-embodiment transfer---yet all three converge on a shared bottleneck: the robotics integration layer that grounds video-derived predictions into closed-loop behavior.
We hope the taxonomy, per-family analyses, and the research directions outlined here offer a useful foundation for advancing video-based manipulation from proof-of-concept demonstrations toward robust, deployable systems.

\section*{Statements and Declarations}
\begin{dci}{}
The author(s) declared no potential conflicts of interest with respect to the research, authorship, and/or publication of this article.
\end{dci}

\begin{funding}{}
This work was supported by the Guangdong Science and Technology Program under Grant No.\ 2024B1212010002.
\end{funding}

\subsection*{\normalsize\sagesf\bfseries Declaration of generative AI and AI-assisted technologies}
\begin{refsize}\noindent
The authors used ChatGPT (OpenAI) and Claude Opus (Anthropic) for language
polishing and wording suggestions.
ChatGPT was also used to generate a few small decorative graphical
assets that the authors incorporated into the figures; these assets are
illustrative only and are not research data. The authors reviewed all AI-assisted
text and graphical elements and take full responsibility for the content of the
manuscript.
\end{refsize}

\bibliographystyle{SageH}
\bibliography{references}

@String(IJCV = {Int. J. Comput. Vis.})

@String(CVPR= {IEEE Conf. Comput. Vis. Pattern Recog.})

@String(ICCV= {Int. Conf. Comput. Vis.})

@String(ECCV= {Eur. Conf. Comput. Vis.})

@String(ICLR = {Int. Conf. Learn. Represent.})

@String(IJCV  = {IJCV})

@String(CVPR  = {CVPR})

@String(ICCV  = {ICCV})

@String(ECCV  = {ECCV})

@String(ICLR  = {ICLR})

@inproceedings{UVA_ShuranSong_2025,
  title     = {Unified Video Action Model},
  author    = {Li, Shuang and Gao, Yihuai and Sadigh, Dorsa and Song, Shuran},
  booktitle = {Proceedings of Robotics: Science and Systems (RSS)},
  year      = {2025},
  doi       = {10.15607/RSS.2025.XXI.074},
  url       = {https://www.roboticsproceedings.org/rss21/p074.pdf}
}

@inproceedings{ATM_Wen_RSS_2024,
  title     = {Any-point Trajectory Modeling for Policy Learning},
  author    = {Wen, Chuan and Lin, Xingyu and So, John Ian Reyes and Chen, Kai and Dou, Qi and Gao, Yang and Abbeel, Pieter},
  booktitle = {Proceedings of Robotics: Science and Systems},
  year      = {2024},
  doi       = {10.15607/RSS.2024.XX.092},
  url       = {https://www.roboticsproceedings.org/rss20/p092.pdf}
}

@misc{cheang2024gr2generativevideolanguageactionmodel,
  title         = {GR-2: A Generative Video-Language-Action Model with Web-Scale Knowledge for Robot Manipulation},
  author        = {Cheang, Chi-Lam and Chen, Guangzeng and Jing, Ya and Kong, Tao and Li, Hang and Li, Yifeng and Liu, Yuxiao and Wu, Hongtao and Xu, Jiafeng and Yang, Yichu and Zhang, Hanbo and Zhu, Minzhao},
  year          = {2024},
  eprint        = {2410.06158},
  archivePrefix = {arXiv},
  primaryClass  = {cs.RO},
  url           = {https://arxiv.org/abs/2410.06158},
  note          = {Technical report}
}

@inproceedings{
Im2Flow2Act_Xu_CoRL_2024,
title={Flow as the Cross-domain Manipulation Interface},
author={Mengda Xu and Zhenjia Xu and Yinghao Xu and Cheng Chi and Gordon Wetzstein and Manuela Veloso and Shuran Song},
booktitle={8th Annual Conference on Robot Learning},
year={2024},
url={https://openreview.net/forum?id=cNI0ZkK1yC}
}

@inproceedings{Dreamitate_Liang_CoRL_2024,
title={Dreamitate: Real-World Visuomotor Policy Learning via Video Generation},
author={Junbang Liang and Ruoshi Liu and Ege Ozguroglu and Sruthi Sudhakar and Achal Dave and Pavel Tokmakov and Shuran Song and Carl Vondrick},
booktitle={8th Annual Conference on Robot Learning},
year={2024},
url={https://openreview.net/forum?id=InT87E5sr4}
}

@inproceedings{UniPi_Du_NIPS_2023,
 author = {Du, Yilun and Yang, Sherry and Dai, Bo and Dai, Hanjun and Nachum, Ofir and Tenenbaum, Josh and Schuurmans, Dale and Abbeel, Pieter},
 booktitle = {Advances in Neural Information Processing Systems},
 editor = {A. Oh and T. Naumann and A. Globerson and K. Saenko and M. Hardt and S. Levine},
 pages = {9156--9172},
 publisher = {Curran Associates, Inc.},
 title = {Learning Universal Policies via Text-Guided Video Generation},
 url = {https://proceedings.neurips.cc/paper_files/paper/2023/file/1d5b9233ad716a43be5c0d3023cb82d0-Paper-Conference.pdf},
 volume = {36},
 year = {2023}
}

@inproceedings{VPP_Hu_2024,
  title     = {Video Prediction Policy: A Generalist Robot Policy with Predictive Visual Representations},
  author    = {Hu, Yucheng and Guo, Yanjiang and Wang, Pengchao and Chen, Xiaoyu and Wang, Yen-Jen and Zhang, Jianke and Sreenath, Koushil and Lu, Chaochao and Chen, Jianyu},
  booktitle = {Proceedings of the 42nd International Conference on Machine Learning},
  series    = {Proceedings of Machine Learning Research},
  volume    = {267},
  pages     = {24328--24346},
  year      = {2025},
  publisher = {PMLR},
  url       = {https://proceedings.mlr.press/v267/hu25g.html}
}

@inproceedings{RT2_Brohan_2023,
    title={RT-2: Vision-Language-Action Models Transfer Web Knowledge to Robotic Control},
    author={Anthony Brohan and Noah Brown and Justice Carbajal and Yevgen Chebotar and Xi Chen and Krzysztof Choromanski and Tianli Ding and Danny Driess and Avinava Dubey and Chelsea Finn and Pete Florence and Chuyuan Fu and Montse Gonzalez Arenas and Keerthana Gopalakrishnan and Kehang Han and Karol Hausman and Alex Herzog and Jasmine Hsu and Brian Ichter and Alex Irpan and Nikhil Joshi and Ryan Julian and Dmitry Kalashnikov and Yuheng Kuang and Isabel Leal  and Lisa Lee and Tsang-Wei Edward Lee and Sergey Levine and Yao Lu and Henryk Michalewski and Igor Mordatch and Karl Pertsch and Kanishka Rao and Krista Reymann and Michael Ryoo and Grecia Salazar and Pannag Sanketi and Pierre Sermanet and Jaspiar Singh and Anikait Singh and Radu Soricut and Huong Tran and Vincent Vanhoucke and Quan Vuong and Ayzaan Wahid and Stefan Welker and Paul Wohlhart and  Jialin Wu and Fei Xia and Ted Xiao and Peng Xu and Sichun Xu and Tianhe Yu and Brianna Zitkovich},
    booktitle={arXiv preprint arXiv:2307.15818},
    year={2023}
}

@inproceedings{RT1_Brohan_2022,
    title={RT-1: Robotics Transformer for Real-World Control at Scale},
    author={Anthony	Brohan and  Noah Brown and  Justice Carbajal and  Yevgen Chebotar and  Joseph Dabis and  Chelsea Finn and  Keerthana Gopalakrishnan and  Karol Hausman and  Alex Herzog and  Jasmine Hsu and  Julian Ibarz and  Brian Ichter and  Alex Irpan and  Tomas Jackson and  Sally Jesmonth and  Nikhil Joshi and  Ryan Julian and  Dmitry Kalashnikov and  Yuheng Kuang and  Isabel Leal and  Kuang-Huei Lee and  Sergey Levine and  Yao Lu and  Utsav Malla and  Deeksha Manjunath and  Igor Mordatch and  Ofir Nachum and  Carolina Parada and  Jodilyn Peralta and  Emily Perez and  Karl Pertsch and  Jornell Quiambao and  Kanishka Rao and  Michael Ryoo and  Grecia Salazar and  Pannag Sanketi and  Kevin Sayed and  Jaspiar Singh and  Sumedh Sontakke and  Austin Stone and  Clayton Tan and  Huong Tran and  Vincent Vanhoucke and Steve Vega and  Quan Vuong and  Fei Xia and  Ted Xiao and  Peng Xu and  Sichun Xu and  Tianhe Yu and  Brianna Zitkovich},
    booktitle={arXiv preprint arXiv:2212.06817},
    year={2022}
}

@inproceedings{Bridgedata_Walke_CoRL_2023,
    title={BridgeData V2: A Dataset for Robot Learning at Scale},
    author={Walke, Homer and Black, Kevin and Lee, Abraham and Kim, Moo Jin and Du, Max and Zheng, Chongyi and Zhao, Tony and Hansen-Estruch, Philippe and Vuong, Quan and He, Andre and Myers, Vivek and Fang, Kuan and Finn, Chelsea and Levine, Sergey},
    booktitle={Conference on Robot Learning (CoRL)},
    year={2023}
}

@inproceedings{
GR1_Wu_2023,
title={Unleashing Large-Scale Video Generative Pre-training for Visual Robot Manipulation},
author={Hongtao Wu and Ya Jing and Chilam Cheang and Guangzeng Chen and Jiafeng Xu and Xinghang Li and Minghuan Liu and Hang Li and Tao Kong},
booktitle={The Twelfth International Conference on Learning Representations},
year={2024},
url={https://openreview.net/forum?id=NxoFmGgWC9}
}

@misc{SpatialVLA_Qu_2025,
      title={SpatialVLA: Exploring Spatial Representations for Visual-Language-Action Model}, 
      author={Delin Qu and Haoming Song and Qizhi Chen and Yuanqi Yao and Xinyi Ye and Yan Ding and Zhigang Wang and JiaYuan Gu and Bin Zhao and Dong Wang and Xuelong Li},
      year={2025},
      eprint={2501.15830},
      archivePrefix={arXiv},
      primaryClass={cs.RO},
      url={https://arxiv.org/abs/2501.15830}, 
}

@misc{pi0_Black_2024,
      title={$\pi_0$: A Vision-Language-Action Flow Model for General Robot Control}, 
      author={Kevin Black and Noah Brown and Danny Driess and Adnan Esmail and Michael Equi and Chelsea Finn and Niccolo Fusai and Lachy Groom and Karol Hausman and Brian Ichter and Szymon Jakubczak and Tim Jones and Liyiming Ke and Sergey Levine and Adrian Li-Bell and Mohith Mothukuri and Suraj Nair and Karl Pertsch and Lucy Xiaoyang Shi and James Tanner and Quan Vuong and Anna Walling and Haohuan Wang and Ury Zhilinsky},
      year={2024},
      eprint={2410.24164},
      archivePrefix={arXiv},
      primaryClass={cs.LG},
      url={https://arxiv.org/abs/2410.24164}, 
}

@misc{Stable_video_diffusion_2023,
      title={Stable Video Diffusion: Scaling Latent Video Diffusion Models to Large Datasets}, 
      author={Andreas Blattmann and Tim Dockhorn and Sumith Kulal and Daniel Mendelevitch and Maciej Kilian and Dominik Lorenz and Yam Levi and Zion English and Vikram Voleti and Adam Letts and Varun Jampani and Robin Rombach},
      year={2023},
      eprint={2311.15127},
      archivePrefix={arXiv},
      primaryClass={cs.CV},
      url={https://arxiv.org/abs/2311.15127}, 
}

@article{ReKep_Huang_2024,
  title = {ReKep: Spatio-Temporal Reasoning of Relational Keypoint Constraints for Robotic Manipulation},
  author = {Huang, Wenlong and Wang, Chen and Li, Yunzhu and Zhang, Ruohan and Fei-Fei, Li},
  journal = {arXiv preprint arXiv:2409.01652},
  year = {2024}
}

@INPROCEEDINGS{OXE_ICRA_2024,
  author={O’Neill, Abby and Rehman, Abdul and Maddukuri, Abhiram and Gupta, Abhishek and Padalkar, Abhishek and Lee, Abraham and Pooley, Acorn and Gupta, Agrim and Mandlekar, Ajay and Jain, Ajinkya and Tung, Albert and Bewley, Alex and Herzog, Alex and Irpan, Alex and Khazatsky, Alexander and Rai, Anant and Gupta, Anchit and Wang, Andrew and Singh, Anikait and Garg, Animesh and Kembhavi, Aniruddha and Xie, Annie and Brohan, Anthony and Raffin, Antonin and Sharma, Archit and Yavary, Arefeh and Jain, Arhan and Balakrishna, Ashwin and Wahid, Ayzaan and Burgess-Limerick, Ben and Kim, Beomjoon and Schölkopf, Bernhard and Wulfe, Blake and Ichter, Brian and Lu, Cewu and Xu, Charles and Le, Charlotte and Finn, Chelsea and Wang, Chen and Xu, Chenfeng and Chi, Cheng and Huang, Chenguang and Chan, Christine and Agia, Christopher and Pan, Chuer and Fu, Chuyuan and Devin, Coline and Xu, Danfei and Morton, Daniel and Driess, Danny and Chen, Daphne and Pathak, Deepak and Shah, Dhruv and Büchler, Dieter and Jayaraman, Dinesh and Kalashnikov, Dmitry and Sadigh, Dorsa and Johns, Edward and Foster, Ethan and Liu, Fangchen and Ceola, Federico and Xia, Fei and Zhao, Feiyu and Stulp, Freek and Zhou, Gaoyue and Sukhatme, Gaurav S. and Salhotra, Gautam and Yan, Ge and Feng, Gilbert and Schiavi, Giulio and Berseth, Glen and Kahn, Gregory and Wang, Guanzhi and Su, Hao and Fang, Hao-Shu and Shi, Haochen and Bao, Henghui and Ben Amor, Heni and Christensen, Henrik I and Furuta, Hiroki and Walke, Homer and Fang, Hongjie and Ha, Huy and Mordatch, Igor and Radosavovic, Ilija and Leal, Isabel and Liang, Jacky and Abou-Chakra, Jad and Kim, Jaehyung and Drake, Jaimyn and Peters, Jan and Schneider, Jan and Hsu, Jasmine and Bohg, Jeannette and Bingham, Jeffrey and Wu, Jeffrey and Gao, Jensen and Hu, Jiaheng and Wu, Jiajun and Wu, Jialin and Sun, Jiankai and Luo, Jianlan and Gu, Jiayuan and Tan, Jie and Oh, Jihoon and Wu, Jimmy and Lu, Jingpei and Yang, Jingyun and Malik, Jitendra and Silvério, João and Hejna, Joey and Booher, Jonathan and Tompson, Jonathan and Yang, Jonathan and Salvador, Jordi and Lim, Joseph J. and Han, Junhyek and Wang, Kaiyuan and Rao, Kanishka and Pertsch, Karl and Hausman, Karol and Go, Keegan and Gopalakrishnan, Keerthana and Goldberg, Ken and Byrne, Kendra and Oslund, Kenneth and Kawaharazuka, Kento and Black, Kevin and Lin, Kevin and Zhang, Kevin and Ehsani, Kiana and Lekkala, Kiran and Ellis, Kirsty and Rana, Krishan and Srinivasan, Krishnan and Fang, Kuan and Singh, Kunal Pratap and Zeng, Kuo-Hao and Hatch, Kyle and Hsu, Kyle and Itti, Laurent and Chen, Lawrence Yunliang and Pinto, Lerrel and Fei-Fei, Li and Tan, Liam and Fan, Linxi Jim and Ott, Lionel and Lee, Lisa and Weihs, Luca and Chen, Magnum and Lepert, Marion and Memmel, Marius and Tomizuka, Masayoshi and Itkina, Masha and Castro, Mateo Guaman and Spero, Max and Du, Maximilian and Ahn, Michael and Yip, Michael C. and Zhang, Mingtong and Ding, Mingyu and Heo, Minho and Srirama, Mohan Kumar and Sharma, Mohit and Kim, Moo Jin and Kanazawa, Naoaki and Hansen, Nicklas and Heess, Nicolas and Joshi, Nikhil J and Suenderhauf, Niko and Liu, Ning and Di Palo, Norman and Shafiullah, Nur Muhammad Mahi and Mees, Oier and Kroemer, Oliver and Bastani, Osbert and Sanketi, Pannag R and Miller, Patrick Tree and Yin, Patrick and Wohlhart, Paul and Xu, Peng and Fagan, Peter David and Mitrano, Peter and Sermanet, Pierre and Abbeel, Pieter and Sundaresan, Priya and Chen, Qiuyu and Vuong, Quan and Rafailov, Rafael and Tian, Ran and Doshi, Ria and Martín-Martín, Roberto and Baijal, Rohan and Scalise, Rosario and Hendrix, Rose and Lin, Roy and Qian, Runjia and Zhang, Ruohan and Mendonca, Russell and Shah, Rutav and Hoque, Ryan and Julian, Ryan and Bustamante, Samuel and Kirmani, Sean and Levine, Sergey and Lin, Shan and Moore, Sherry and Bahl, Shikhar and Dass, Shivin and Sonawani, Shubham and Song, Shuran and Xu, Sichun and Haldar, Siddhant and Karamcheti, Siddharth and Adebola, Simeon and Guist, Simon and Nasiriany, Soroush and Schaal, Stefan and Welker, Stefan and Tian, Stephen and Ramamoorthy, Subramanian and Dasari, Sudeep and Belkhale, Suneel and Park, Sungjae and Nair, Suraj and Mirchandani, Suvir and Osa, Takayuki and Gupta, Tanmay and Harada, Tatsuya and Matsushima, Tatsuya and Xiao, Ted and Kollar, Thomas and Yu, Tianhe and Ding, Tianli and Davchev, Todor and Zhao, Tony Z. and Armstrong, Travis and Darrell, Trevor and Chung, Trinity and Jain, Vidhi and Vanhoucke, Vincent and Zhan, Wei and Zhou, Wenxuan and Burgard, Wolfram and Chen, Xi and Wang, Xiaolong and Zhu, Xinghao and Geng, Xinyang and Liu, Xiyuan and Liangwei, Xu and Li, Xuanlin and Lu, Yao and Ma, Yecheng Jason and Kim, Yejin and Chebotar, Yevgen and Zhou, Yifan and Zhu, Yifeng and Wu, Yilin and Xu, Ying and Wang, Yixuan and Bisk, Yonatan and Cho, Yoonyoung and Lee, Youngwoon and Cui, Yuchen and Cao, Yue and Wu, Yueh-Hua and Tang, Yujin and Zhu, Yuke and Zhang, Yunchu and Jiang, Yunfan and Li, Yunshuang and Li, Yunzhu and Iwasawa, Yusuke and Matsuo, Yutaka and Ma, Zehan and Xu, Zhuo and Cui, Zichen Jeff and Zhang, Zichen and Lin, Zipeng},
  booktitle={2024 IEEE International Conference on Robotics and Automation (ICRA)}, 
  title={Open X-Embodiment: Robotic Learning Datasets and RT-X Models : Open X-Embodiment Collaboration0}, 
  year={2024},
  volume={},
  number={},
  pages={6892-6903},
  doi={10.1109/ICRA57147.2024.10611477}}

@INPROCEEDINGS{Ego4d_CVPR_2022,
  author={Grauman, Kristen and Westbury, Andrew and Byrne, Eugene and Chavis, Zachary and Furnari, Antonino and Girdhar, Rohit and Hamburger, Jackson and Jiang, Hao and Liu, Miao and Liu, Xingyu and Martin, Miguel and Nagarajan, Tushar and Radosavovic, Ilija and Ramakrishnan, Santhosh Kumar and Ryan, Fiona and Sharma, Jayant and Wray, Michael and Xu, Mengmeng and Xu, Eric Zhongcong and Zhao, Chen and Bansal, Siddhant and Batra, Dhruv and Cartillier, Vincent and Crane, Sean and Do, Tien and Doulaty, Morrie and Erapalli, Akshay and Feichtenhofer, Christoph and Fragomeni, Adriano and Fu, Qichen and Gebreselasie, Abrham and González, Cristina and Hillis, James and Huang, Xuhua and Huang, Yifei and Jia, Wenqi and Khoo, Weslie and Koláĭ, Jáchym and Kottur, Satwik and Kumar, Anurag and Landini, Federico and Li, Chao and Li, Yanghao and Li, Zhenqiang and Mangalam, Karttikeya and Modhugu, Raghava and Munro, Jonathan and Murrell, Tullie and Nishiyasu, Takumi and Price, Will and Puentes, Paola Ruiz and Ramazanova, Merey and Sari, Leda and Somasundaram, Kiran and Southerland, Audrey and Sugano, Yusuke and Tao, Ruijie and Vo, Minh and Wang, Yuchen and Wu, Xindi and Yagi, Takuma and Zhao, Ziwei and Zhu, Yunyi and Arbeláez, Pablo and Crandall, David and Damen, Dima and Farinella, Giovanni Maria and Fuegen, Christian and Ghanem, Bernard and Ithapu, Vamsi Krishna and Jawahar, C. V. and Joo, Hanbyul and Kitani, Kris and Li, Haizhou and Newcombe, Richard and Oliva, Aude and Park, Hyun Soo and Rehg, James M. and Sato, Yoichi and Shi, Jianbo and Shou, Mike Zheng and Torralba, Antonio and Torresani, Lorenzo and Yan, Mingfei and Malik, Jitendra},
  booktitle={2022 IEEE/CVF Conference on Computer Vision and Pattern Recognition (CVPR)}, 
  title={Ego4D: Around the World in 3,000 Hours of Egocentric Video}, 
  year={2022},
  volume={},
  number={},
  pages={18973-18990},
  doi={10.1109/CVPR52688.2022.01842}}

@misc{Dinov2_2023,
      title={DINOv2: Learning Robust Visual Features without Supervision}, 
      author={Maxime Oquab and Timothée Darcet and Théo Moutakanni and Huy Vo and Marc Szafraniec and Vasil Khalidov and Pierre Fernandez and Daniel Haziza and Francisco Massa and Alaaeldin El-Nouby and Mahmoud Assran and Nicolas Ballas and Wojciech Galuba and Russell Howes and Po-Yao Huang and Shang-Wen Li and Ishan Misra and Michael Rabbat and Vasu Sharma and Gabriel Synnaeve and Hu Xu and Hervé Jegou and Julien Mairal and Patrick Labatut and Armand Joulin and Piotr Bojanowski},
      year={2024},
      eprint={2304.07193},
      archivePrefix={arXiv},
      primaryClass={cs.CV},
      url={https://arxiv.org/abs/2304.07193}, 
}

@article{LIBERO_Liu_2023,
  title={LIBERO: Benchmarking Knowledge Transfer for Lifelong Robot Learning},
  author={Liu, Bo and Zhu, Yifeng and Gao, Chongkai and Feng, Yihao and Liu, Qiang and Zhu, Yuke and Stone, Peter},
  journal={arXiv preprint arXiv:2306.03310},
  year={2023}
}

@INPROCEEDINGS{UniVLA_RSS_25, 
    AUTHOR    = {Qingwen Bu AND Yanting Yang AND Jisong Cai AND Shenyuan Gao AND Guanghui Ren AND Maoqing Yao AND Ping Luo AND Hongyang Li}, 
    TITLE     = {{Learning to Act Anywhere with Task-centric Latent Actions}}, 
    BOOKTITLE = {Proceedings of Robotics: Science and Systems}, 
    YEAR      = {2025}, 
    ADDRESS   = {LosAngeles, CA, USA}, 
    MONTH     = {June}, 
    DOI       = {10.15607/RSS.2025.XXI.014} 
}

@inproceedings{PAD_guo2024_NIPS,
  title     = {Prediction with Action: Visual Policy Learning via Joint Denoising Process},
  author    = {Guo, Yanjiang and Hu, Yucheng and Zhang, Jianke and Wang, Yen-Jen and Chen, Xiaoyu and Lu, Chaochao and Chen, Jianyu},
  booktitle = {Advances in Neural Information Processing Systems},
  volume    = {37},
  year      = {2024},
  doi       = {10.52202/079017-3570},
  url       = {https://proceedings.neurips.cc/paper_files/paper/2024/hash/cbe25fa0e7c7084049276888a09acc8d-Abstract-Conference.html}
}

@inproceedings{UWM_zhu2025_RSS,
  title     = {Unified World Models: Coupling Video and Action Diffusion for Pretraining on Large Robotic Datasets},
  author    = {Zhu, Chuning and Yu, Raymond and Feng, Siyuan and Burchfiel, Benjamin and Shah, Paarth and Gupta, Abhishek},
  booktitle = {Proceedings of Robotics: Science and Systems (RSS)},
  year      = {2025},
  doi       = {10.15607/RSS.2025.XXI.015},
  url       = {https://www.roboticsproceedings.org/rss21/p015.pdf}
}

@misc{ILPO_edwards2019_ICML,
      title={Imitating Latent Policies from Observation}, 
      author={Ashley D. Edwards and Himanshu Sahni and Yannick Schroecker and Charles L. Isbell},
      year={2019},
      eprint={1805.07914},
      archivePrefix={arXiv},
      primaryClass={cs.LG},
      url={https://arxiv.org/abs/1805.07914}, 
}

@misc{LAPO_schmidt2024_ICLR,
      title={Learning to Act without Actions}, 
      author={Dominik Schmidt and Minqi Jiang},
      year={2024},
      eprint={2312.10812},
      archivePrefix={arXiv},
      primaryClass={cs.LG},
      url={https://arxiv.org/abs/2312.10812}, 
}

@inproceedings{LAPA_ye2025,
  title     = {Latent Action Pretraining from Videos},
  author    = {Ye, Seonghyeon and Jang, Joel and Jeon, Byeongguk and Joo, Se June and Yang, Jianwei and Peng, Baolin and Mandlekar, Ajay and Tan, Reuben and Chao, Yu-Wei and Lin, Bill Yuchen and Liden, Lars and Lee, Kimin and Gao, Jianfeng and Zettlemoyer, Luke and Fox, Dieter and Seo, Minjoon},
  booktitle = {International Conference on Learning Representations (ICLR)},
  year      = {2025},
  url       = {https://openreview.net/forum?id=VYOe2eBQeh}
}

@inproceedings{CLASP_ICLR2019,
  title     = {Learning What You Can Do Before Doing Anything},
  author    = {Rybkin, Oleh and Pertsch, Karl and Derpanis, Konstantinos G. and Daniilidis, Kostas and Jaegle, Andrew},
  booktitle = {International Conference on Learning Representations (ICLR)},
  year      = {2019},
  url       = {https://openreview.net/forum?id=SylPMnR9Ym}
}

@misc{InformationBottleneck_2017,
      title={Opening the Black Box of Deep Neural Networks via Information}, 
      author={Ravid Shwartz-Ziv and Naftali Tishby},
      year={2017},
      eprint={1703.00810},
      archivePrefix={arXiv},
      primaryClass={cs.LG},
      url={https://arxiv.org/abs/1703.00810}, 
}

@inproceedings{
FICC_ICLR_2023,
title={Become a Proficient Player with Limited Data through Watching Pure Videos},
author={Weirui Ye and Yunsheng Zhang and Pieter Abbeel and Yang Gao},
booktitle={The Eleventh International Conference on Learning Representations },
year={2023},
url={https://openreview.net/forum?id=Sy-o2N0hF4f},
}

@misc{VQVAE_NIPS_2017,
      title={Neural Discrete Representation Learning}, 
      author={Aaron van den Oord and Oriol Vinyals and Koray Kavukcuoglu},
      year={2018},
      eprint={1711.00937},
      archivePrefix={arXiv},
      primaryClass={cs.LG},
      url={https://arxiv.org/abs/1711.00937}, 
}

@misc{Genie_1_ICML_2024,
      title={Genie: Generative Interactive Environments}, 
      author={Jake Bruce and Michael Dennis and Ashley Edwards and Jack Parker-Holder and Yuge Shi and Edward Hughes and Matthew Lai and Aditi Mavalankar and Richie Steigerwald and Chris Apps and Yusuf Aytar and Sarah Bechtle and Feryal Behbahani and Stephanie Chan and Nicolas Heess and Lucy Gonzalez and Simon Osindero and Sherjil Ozair and Scott Reed and Jingwei Zhang and Konrad Zolna and Jeff Clune and Nando de Freitas and Satinder Singh and Tim Rocktäschel},
      year={2024},
      eprint={2402.15391},
      archivePrefix={arXiv},
      primaryClass={cs.LG},
      url={https://arxiv.org/abs/2402.15391}, 
}

@inproceedings{howto100m_ICCV_2019,
   title={How{T}o100{M}: {L}earning a {T}ext-{V}ideo {E}mbedding by {W}atching {H}undred {M}illion {N}arrated {V}ideo {C}lips},
   author={Miech, Antoine and Zhukov, Dimitri and Alayrac, Jean-Baptiste and Tapaswi, Makarand and Laptev, Ivan and Sivic, Josef},
   booktitle={ICCV},
   year={2019},
}

@misc{somethingSomethingV2_2017,
      title={The "something something" video database for learning and evaluating visual common sense}, 
      author={Raghav Goyal and Samira Ebrahimi Kahou and Vincent Michalski and Joanna Materzyńska and Susanne Westphal and Heuna Kim and Valentin Haenel and Ingo Fruend and Peter Yianilos and Moritz Mueller-Freitag and Florian Hoppe and Christian Thurau and Ingo Bax and Roland Memisevic},
      year={2017},
      eprint={1706.04261},
      archivePrefix={arXiv},
      primaryClass={cs.CV},
      url={https://arxiv.org/abs/1706.04261}, 
}

@ARTICLE{EPIC-KITCHENS-100_IJCV_2022,
           title={Rescaling Egocentric Vision: Collection, Pipeline and Challenges for EPIC-KITCHENS-100},
           author={Damen, Dima and Doughty, Hazel and Farinella, Giovanni Maria and Furnari, Antonino 
           and Ma, Jian and Kazakos, Evangelos and Moltisanti, Davide and Munro, Jonathan 
           and Perrett, Toby and Price, Will and Wray, Michael},
           journal   = {International Journal of Computer Vision (IJCV)},
           year      = {2022},
           volume = {130},
           pages = {33–55},
           Url       = {https://doi.org/10.1007/s11263-021-01531-2}
}

@misc{CALVIN_2022,
      title={CALVIN: A Benchmark for Language-Conditioned Policy Learning for Long-Horizon Robot Manipulation Tasks}, 
      author={Oier Mees and Lukas Hermann and Erick Rosete-Beas and Wolfram Burgard},
      year={2022},
      eprint={2112.03227},
      archivePrefix={arXiv},
      primaryClass={cs.RO},
      url={https://arxiv.org/abs/2112.03227}, 
}

@inproceedings{VidMan_2024_NIPS,
  title     = {VidMan: Exploiting Implicit Dynamics from Video Diffusion Model for Effective Robot Manipulation},
  author    = {Wen, Youpeng and Lin, Junfan and Zhu, Yi and Han, Jianhua and Xu, Hang and Zhao, Shen and Liang, Xiaodan},
  booktitle = {Advances in Neural Information Processing Systems},
  volume    = {37},
  year      = {2024},
  doi       = {10.52202/079017-1298},
  url       = {https://proceedings.neurips.cc/paper_files/paper/2024/hash/481c70828a4ff20d31a646cc6cc95f3d-Abstract-Conference.html}
}

@inproceedings{track2act_ECCV_2024,
  title     = {Track2Act: Predicting Point Tracks from Internet Videos Enables Generalizable Robot Manipulation},
  author    = {Bharadhwaj, Homanga and Mottaghi, Roozbeh and Gupta, Abhinav and Tulsiani, Shubham},
  booktitle = {Computer Vision -- ECCV 2024},
  pages     = {306--324},
  year      = {2024},
  doi       = {10.1007/978-3-031-73116-7_18},
  url       = {https://doi.org/10.1007/978-3-031-73116-7_18}
}

@article{VAEs_Kingma_2019,
   title={An Introduction to Variational Autoencoders},
   volume={12},
   ISSN={1935-8245},
   url={http://dx.doi.org/10.1561/2200000056},
   DOI={10.1561/2200000056},
   number={4},
   journal={Foundations and Trends® in Machine Learning},
   publisher={Emerald},
   author={Kingma, Diederik P. and Welling, Max},
   year={2019},
   pages={307–392} }

@inproceedings{SKIL_RSS_2025,
  title     = {SKIL: Semantic Keypoint Imitation Learning for Generalizable Data-efficient Manipulation},
  author    = {Wang, Shengjie and You, Jiacheng and Hu, Yihang and Li, Jiongye and Gao, Yang},
  booktitle = {Proceedings of Robotics: Science and Systems (RSS)},
  year      = {2025},
  doi       = {10.15607/RSS.2025.XXI.161},
  url       = {https://www.roboticsproceedings.org/rss21/p161.pdf}
}

@inproceedings{
susie_2023,
title={Zero-Shot Robotic Manipulation with Pre-Trained Image-Editing Diffusion Models},
author={Kevin Black and Mitsuhiko Nakamoto and Pranav Atreya and Homer Rich Walke and Chelsea Finn and Aviral Kumar and Sergey Levine},
booktitle={The Twelfth International Conference on Learning Representations},
year={2024},
url={https://openreview.net/forum?id=c0chJTSbci}
}

@article{EPnP_IJCV_2009,
author = {Lepetit, Vincent and Moreno-Noguer, Francesc and Fua, Pascal},
year = {2009},
month = {02},
pages = {},
title = {{EPnP: An accurate O(n) solution to the PnP problem}},
volume = {81},
journal = {International Journal of Computer Vision (IJCV)},
doi = {10.1007/s11263-008-0152-6}
}

@article{RANSAC,
author = {Fischler, Martin A. and Bolles, Robert C.},
title = {Random sample consensus: a paradigm for model fitting with applications to image analysis and automated cartography},
year = {1981},
issue_date = {June 1981},
publisher = {Association for Computing Machinery},
address = {New York, NY, USA},
volume = {24},
number = {6},
issn = {0001-0782},
url = {https://doi.org/10.1145/358669.358692},
doi = {10.1145/358669.358692},
journal = {Commun. ACM},
month = jun,
pages = {381–395},
numpages = {15}
}

@misc{InstructPix2Pix_2023,
      title={InstructPix2Pix: Learning to Follow Image Editing Instructions}, 
      author={Tim Brooks and Aleksander Holynski and Alexei A. Efros},
      year={2023},
      eprint={2211.09800},
      archivePrefix={arXiv},
      primaryClass={cs.CV},
      url={https://arxiv.org/abs/2211.09800}, 
}

@inproceedings{GVFTAPE_CoRL_2025,
  title     = {Generative Visual Foresight Meets Task-Agnostic Pose Estimation in Robotic Table-top Manipulation},
  author    = {Zhang, Chuye and Zhang, Xiaoxiong and Pan, Wei and Zheng, Linfang and Zhang, Wei},
  booktitle = {Proceedings of The 9th Conference on Robot Learning},
  series    = {Proceedings of Machine Learning Research},
  volume    = {305},
  pages     = {2823--2846},
  year      = {2025},
  publisher = {PMLR},
  url       = {https://proceedings.mlr.press/v305/zhang25g.html}
}

@inproceedings{Rt-affordance_ICRA_2025,
  title={Rt-affordance: Affordances are versatile intermediate representations for robot manipulation},
  author={Nasiriany, Soroush and Kirmani, Sean and Ding, Tianli and Smith, Laura and Zhu, Yuke and Driess, Danny and Sadigh, Dorsa and Xiao, Ted},
  booktitle={2025 IEEE International Conference on Robotics and Automation (ICRA)},
  pages={8249--8257},
  year={2025},
  organization={IEEE}
}

@misc{GENIMA_CoRL_2024,
      title={Generative Image as Action Models}, 
      author={Mohit Shridhar and Yat Long Lo and Stephen James},
      year={2024},
      eprint={2407.07875},
      archivePrefix={arXiv},
      primaryClass={cs.RO},
      url={https://arxiv.org/abs/2407.07875}, 
}

@inproceedings{CLOVER_2024,
  title     = {Closed-Loop Visuomotor Control with Generative Expectation for Robotic Manipulation},
  author    = {Bu, Qingwen and Zeng, Jia and Chen, Li and Yang, Yanchao and Zhou, Guyue and Yan, Junchi and Luo, Ping and Cui, Heming and Ma, Yi and Li, Hongyang},
  booktitle = {Advances in Neural Information Processing Systems},
  volume    = {37},
  year      = {2024},
  doi       = {10.52202/079017-4411},
  url       = {https://proceedings.neurips.cc/paper_files/paper/2024/hash/fad8962279154544ed69bb63eb14d677-Abstract-Conference.html}
}

@inproceedings{
AVDC_2023,
title={Learning to Act from Actionless Videos through Dense Correspondences},
author={Po-Chen Ko and Jiayuan Mao and Yilun Du and Shao-Hua Sun and Joshua B. Tenenbaum},
booktitle={The Twelfth International Conference on Learning Representations},
year={2024},
url={https://openreview.net/forum?id=Mhb5fpA1T0}
}

@inproceedings{
V2A_2025,
title={Grounding Video Models to Actions through Goal Conditioned Exploration},
author={Yunhao Luo and Yilun Du},
booktitle={The Thirteenth International Conference on Learning Representations},
year={2025},
url={https://openreview.net/forum?id=G6dMvRuhFr}
}

@article{moka_RSS2024,
  title={Moka: Open-world robotic manipulation through mark-based visual prompting},
  author={Liu, Fangchen and Fang, Kuan and Abbeel, Pieter and Levine, Sergey},
  journal={arXiv preprint arXiv:2403.03174},
  year={2024}
}

@inproceedings{keto_ICRA2020,
  title={Keto: Learning keypoint representations for tool manipulation},
  author={Qin, Zengyi and Fang, Kuan and Zhu, Yuke and Fei-Fei, Li and Savarese, Silvio},
  booktitle={2020 IEEE International Conference on Robotics and Automation (ICRA)},
  pages={7278--7285},
  year={2020},
  organization={IEEE}
}

@inproceedings{tra_moe_CVPR2025,
  title={Tra-moe: Learning trajectory prediction model from multiple domains for adaptive policy conditioning},
  author={Yang, Jiange and Zhu, Haoyi and Wang, Yating and Wu, Gangshan and He, Tong and Wang, Limin},
  booktitle={Proceedings of the Computer Vision and Pattern Recognition Conference},
  pages={6960--6970},
  year={2025}
}

@misc{MegaPose_2022,
      title={MegaPose: 6D Pose Estimation of Novel Objects via Render \& Compare}, 
      author={Yann Labbé and Lucas Manuelli and Arsalan Mousavian and Stephen Tyree and Stan Birchfield and Jonathan Tremblay and Justin Carpentier and Mathieu Aubry and Dieter Fox and Josef Sivic},
      year={2022},
      eprint={2212.06870},
      archivePrefix={arXiv},
      primaryClass={cs.CV},
      url={https://arxiv.org/abs/2212.06870}, 
}

@inproceedings{GeneralFlow_CoRL_2024,
  title     = {General Flow as Foundation Affordance for Scalable Robot Learning},
  author    = {Yuan, Chengbo and Wen, Chuan and Zhang, Tong and Gao, Yang},
  booktitle = {Proceedings of the 8th Conference on Robot Learning},
  series    = {Proceedings of Machine Learning Research},
  volume    = {270},
  pages     = {1541--1566},
  year      = {2024},
  publisher = {PMLR},
  url       = {https://proceedings.mlr.press/v270/yuan25a.html}
}

@article{flowbot3d_RSS_2022,
  title={Flowbot3d: Learning 3d articulation flow to manipulate articulated objects},
  author={Eisner, Ben and Zhang, Harry and Held, David},
  journal={arXiv preprint arXiv:2205.04382},
  year={2022}
}

@inproceedings{APV_2022,
  title     = {Reinforcement Learning with Action-Free Pre-Training from Videos},
  author    = {Seo, Younggyo and Lee, Kimin and James, Stephen and Abbeel, Pieter},
  booktitle = {Proceedings of the 39th International Conference on Machine Learning},
  series    = {Proceedings of Machine Learning Research},
  volume    = {162},
  pages     = {19561--19579},
  year      = {2022},
  publisher = {PMLR},
  url       = {https://proceedings.mlr.press/v162/seo22a.html}
}

@inproceedings{ContextWM_NIPS_2023,
title={Pre-training Contextualized World Models with In-the-wild Videos for Reinforcement Learning},
author={Jialong Wu and Haoyu Ma and Chaoyi Deng and Mingsheng Long},
booktitle={Thirty-seventh Conference on Neural Information Processing Systems},
year={2023},
url={https://openreview.net/forum?id=8GuEVzAUQS}
}

@inproceedings{SWIM_2023,
  title     = {Structured World Models from Human Videos},
  author    = {Mendonca, Russell and Bahl, Shikhar and Pathak, Deepak},
  booktitle = {Proceedings of Robotics: Science and Systems (RSS)},
  year      = {2023},
  doi       = {10.15607/RSS.2023.XIX.012},
  url       = {https://www.roboticsproceedings.org/rss19/p012.pdf}
}

@misc{VIP_ICLR_2023,
      title={VIP: Towards Universal Visual Reward and Representation via Value-Implicit Pre-Training}, 
      author={Yecheng Jason Ma and Shagun Sodhani and Dinesh Jayaraman and Osbert Bastani and Vikash Kumar and Amy Zhang},
      year={2023},
      eprint={2210.00030},
      archivePrefix={arXiv},
      primaryClass={cs.RO},
      url={https://arxiv.org/abs/2210.00030}, 
}

@misc{ManipulateBySeeing_ICCV_2023,
      title={Manipulate by Seeing: Creating Manipulation Controllers from Pre-Trained Representations}, 
      author={Jianren Wang and Sudeep Dasari and Mohan Kumar Srirama and Shubham Tulsiani and Abhinav Gupta},
      year={2023},
      eprint={2303.08135},
      archivePrefix={arXiv},
      primaryClass={cs.RO},
      url={https://arxiv.org/abs/2303.08135}, 
}

@misc{LIV_ma_2023,
      title={LIV: Language-Image Representations and Rewards for Robotic Control}, 
      author={Yecheng Jason Ma and William Liang and Vaidehi Som and Vikash Kumar and Amy Zhang and Osbert Bastani and Dinesh Jayaraman},
      year={2023},
      eprint={2306.00958},
      archivePrefix={arXiv},
      primaryClass={cs.RO},
      url={https://arxiv.org/abs/2306.00958}, 
}

@misc{AVID1_smith2020,
      title={AVID: Learning Multi-Stage Tasks via Pixel-Level Translation of Human Videos}, 
      author={Laura Smith and Nikita Dhawan and Marvin Zhang and Pieter Abbeel and Sergey Levine},
      year={2020},
      eprint={1912.04443},
      archivePrefix={arXiv},
      primaryClass={cs.RO},
      url={https://arxiv.org/abs/1912.04443}, 
}

@article{MoE_2017,
  title={Outrageously large neural networks: The sparsely-gated mixture-of-experts layer},
  author={Shazeer, Noam and Mirhoseini, Azalia and Maziarz, Krzysztof and Davis, Andy and Le, Quoc and Hinton, Geoffrey and Dean, Jeff},
  journal={arXiv preprint arXiv:1701.06538},
  year={2017}
}

@inproceedings{Cotracker_2024,
  title={Cotracker: It is better to track together},
  author={Karaev, Nikita and Rocco, Ignacio and Graham, Benjamin and Neverova, Natalia and Vedaldi, Andrea and Rupprecht, Christian},
  booktitle={European conference on computer vision},
  pages={18--35},
  year={2024},
  organization={Springer}
}

@misc{UniSim_ICRA_2024,
      title={Learning Interactive Real-World Simulators}, 
      author={Sherry Yang and Yilun Du and Kamyar Ghasemipour and Jonathan Tompson and Leslie Kaelbling and Dale Schuurmans and Pieter Abbeel},
      year={2024},
      eprint={2310.06114},
      archivePrefix={arXiv},
      primaryClass={cs.AI},
      url={https://arxiv.org/abs/2310.06114}, 
}

@misc{DreamerV2_2022,
      title={Mastering Atari with Discrete World Models}, 
      author={Danijar Hafner and Timothy Lillicrap and Mohammad Norouzi and Jimmy Ba},
      year={2022},
      eprint={2010.02193},
      archivePrefix={arXiv},
      primaryClass={cs.LG},
      url={https://arxiv.org/abs/2010.02193}, 
}

@inproceedings{VRB_2023,
  title     = {Affordances from Human Videos as a Versatile Representation for Robotics},
  author    = {Bahl, Shikhar and Mendonca, Russell and Chen, Lili and Jain, Unnat and Pathak, Deepak},
  booktitle = {Proceedings of the IEEE/CVF Conference on Computer Vision and Pattern Recognition (CVPR)},
  pages     = {13778--13790},
  year      = {2023},
  url       = {https://openaccess.thecvf.com/content/CVPR2023/html/Bahl_Affordances_From_Human_Videos_as_a_Versatile_Representation_for_Robotics_CVPR_2023_paper.html}
}

@inproceedings{
MimicPlay_CoRL_2023,
title={MimicPlay: Long-Horizon Imitation Learning by Watching Human Play},
author={Chen Wang and Linxi Fan and Jiankai Sun and Ruohan Zhang and Li Fei-Fei and Danfei Xu and Yuke Zhu and Anima Anandkumar},
booktitle={7th Annual Conference on Robot Learning},
year={2023},
url={https://openreview.net/forum?id=hRZ1YjDZmTo}
}

@INPROCEEDINGS{ZeroMimic_ICRA_2025,
  author={Shi, Junyao and Zhao, Zhuolun and Wang, Tianyou and Pedroza, Ian and Luo, Amy and Wang, Jie and Ma, Jason and Jayaraman, Dinesh},
  booktitle={2025 IEEE International Conference on Robotics and Automation (ICRA)}, 
  title={ZeroMimic: Distilling Robotic Manipulation Skills from Web Videos}, 
  year={2025},
  volume={},
  number={},
  pages={16939-16947},
  doi={10.1109/ICRA55743.2025.11128283}}

@inproceedings{HOI4D_CVPR_2022,
  title={HOI4D: A 4D Egocentric Dataset for Category-Level Human-Object Interaction},
  author={Liu, Yunze and Liu, Yun and Jiang, Che and Lyu, Kangbo and Wan, Weikang and Shen, Hao and Liang, Boqiang and Fu, Zhoujie and Wang, He and Yi, Li},
  booktitle={Proceedings of the IEEE/CVF Conference on Computer Vision and Pattern Recognition},
  pages={21013--21022},
  year={2022}
}

@ARTICLE{ICP_1987,
  author={Arun, K. S. and Huang, T. S. and Blostein, S. D.},
  journal={IEEE Transactions on Pattern Analysis and Machine Intelligence}, 
  title={Least-Squares Fitting of Two 3-D Point Sets}, 
  year={1987},
  volume={PAMI-9},
  number={5},
  pages={698-700},
  doi={10.1109/TPAMI.1987.4767965}}

@misc{xu2024surveyroboticsfoundationmodels,
      title={A Survey on Robotics with Foundation Models: toward Embodied AI}, 
      author={Zhiyuan Xu and Kun Wu and Junjie Wen and Jinming Li and Ning Liu and Zhengping Che and Jian Tang},
      year={2024},
      eprint={2402.02385},
      archivePrefix={arXiv},
      primaryClass={cs.RO},
      url={https://arxiv.org/abs/2402.02385}, 
}

@misc{zhong2025surveyvisionlanguageactionmodelsaction,
      title={A Survey on Vision-Language-Action Models: An Action Tokenization Perspective}, 
      author={Yifan Zhong and Fengshuo Bai and Shaofei Cai and Xuchuan Huang and Zhang Chen and Xiaowei Zhang and Yuanfei Wang and Shaoyang Guo and Tianrui Guan and Ka Nam Lui and Zhiquan Qi and Yitao Liang and Yuanpei Chen and Yaodong Yang},
      year={2025},
      eprint={2507.01925},
      archivePrefix={arXiv},
      primaryClass={cs.RO},
      url={https://arxiv.org/abs/2507.01925}, 
}

@misc{eze2025learningwatchingreviewvideobased,
      title={Learning by Watching: A Review of Video-based Learning Approaches for Robot Manipulation}, 
      author={Chrisantus Eze and Christopher Crick},
      year={2025},
      eprint={2402.07127},
      archivePrefix={arXiv},
      primaryClass={cs.RO},
      url={https://arxiv.org/abs/2402.07127}, 
}

@article{james2020rlbench,
  title={Rlbench: The robot learning benchmark \& learning environment},
  author={James, Stephen and Ma, Zicong and Arrojo, David Rovick and Davison, Andrew J},
  journal={IEEE Robotics and Automation Letters},
  volume={5},
  number={2},
  pages={3019--3026},
  year={2020},
  publisher={IEEE}
}

@inproceedings{yu2020meta_world,
  title={Meta-world: A benchmark and evaluation for multi-task and meta reinforcement learning},
  author={Yu, Tianhe and Quillen, Deirdre and He, Zhanpeng and Julian, Ryan and Hausman, Karol and Finn, Chelsea and Levine, Sergey},
  booktitle={Conference on robot learning},
  pages={1094--1100},
  year={2020},
  organization={PMLR}
}

@article{mandlekar2021robomimic,
  title={What matters in learning from offline human demonstrations for robot manipulation},
  author={Mandlekar, Ajay and Xu, Danfei and Wong, Josiah and Nasiriany, Soroush and Wang, Chen and Kulkarni, Rohun and Fei-Fei, Li and Savarese, Silvio and Zhu, Yuke and Mart{\'\i}n-Mart{\'\i}n, Roberto},
  journal={arXiv preprint arXiv:2108.03298},
  year={2021}
}

@inproceedings{gu2023maniskill2,
title={ManiSkill2: A Unified Benchmark for Generalizable Manipulation Skills},
author={Jiayuan Gu and Fanbo Xiang and Xuanlin Li and Zhan Ling and Xiqiang Liu and Tongzhou Mu and Yihe Tang and Stone Tao and Xinyue Wei and Yunchao Yao and Xiaodi Yuan and Pengwei Xie and Zhiao Huang and Rui Chen and Hao Su},
booktitle={The Eleventh International Conference on Learning Representations },
year={2023},
url={https://openreview.net/forum?id=b_CQDy9vrD1}
}

@article{pumacay2024colosseum,
  title={The colosseum: A benchmark for evaluating generalization for robotic manipulation},
  author={Pumacay, Wilbert and Singh, Ishika and Duan, Jiafei and Krishna, Ranjay and Thomason, Jesse and Fox, Dieter},
  journal={arXiv preprint arXiv:2402.08191},
  year={2024}
}

@article{survey_learning_based_dynamics_2025,
author = {Bo Ai  and Stephen Tian  and Haochen Shi  and Yixuan Wang  and Tobias Pfaff  and Cheston Tan  and Henrik I. Christensen  and Hao Su  and Jiajun Wu  and Yunzhu Li },
title = {A review of learning-based dynamics models for robotic manipulation},
journal = {Science Robotics},
volume = {10},
number = {106},
pages = {eadt1497},
year = {2025},
doi = {10.1126/scirobotics.adt1497},
URL = {https://www.science.org/doi/abs/10.1126/scirobotics.adt1497},
eprint = {https://www.science.org/doi/pdf/10.1126/scirobotics.adt1497},
}

@inproceedings{
khazatsky2024droid,
title={{DROID}: A Large-Scale In-The-Wild Robot Manipulation Dataset},
author={Alexander Khazatsky and Karl Pertsch and Suraj Nair and Ashwin Balakrishna and Sudeep Dasari and Siddharth Karamcheti and Soroush Nasiriany and Mohan Kumar Srirama and Lawrence Yunliang Chen and Kirsty Ellis and Peter David Fagan and Joey Hejna and Masha Itkina and Marion Lepert and Yecheng Jason Ma and Patrick Tree Miller and Jimmy Wu and Suneel Belkhale and Shivin Dass and Huy Ha and Arhan Jain and Abraham Lee and Youngwoon Lee and Marius Memmel and Sungjae Park and Ilija Radosavovic and Kaiyuan Wang and Albert Zhan and Kevin Black and Cheng Chi and Kyle Beltran Hatch and Shan Lin and Jingpei Lu and Jean Mercat and Abdul Rehman and Pannag R Sanketi and Archit Sharma and Cody Simpson and Quan Vuong and Homer Rich Walke and Blake Wulfe and Ted Xiao and Jonathan Heewon Yang and Arefeh Yavary and Tony Z. Zhao and Christopher Agia and Rohan Baijal and Mateo Guaman Castro and Daphne Chen and Qiuyu Chen and Trinity Chung and Jaimyn Drake and Ethan Paul Foster and Jensen Gao and David Antonio Herrera and Minho Heo and Kyle Hsu and Jiaheng Hu and Donovon Jackson and Charlotte Le and Yunshuang Li and Xinyu Lin and Zehan Ma and Abhiram Maddukuri and Suvir Mirchandani and Daniel Morton and Tony Khuong Nguyen and Abigail O'Neill and Rosario Scalise and Derick Seale and Victor Son and Stephen Tian and Emi Tran and Andrew E. Wang and Yilin Wu and Annie Xie and Jingyun Yang and Patrick Yin and Yunchu Zhang and Osbert Bastani and Glen Berseth and Jeannette Bohg and Ken Goldberg and Abhinav Gupta and Abhishek Gupta and Dinesh Jayaraman and Joseph J Lim and Jitendra Malik and Roberto Mart{\'\i}n-Mart{\'\i}n and Subramanian Ramamoorthy and Dorsa Sadigh and Shuran Song and Jiajun Wu and Michael C. Yip and Yuke Zhu and Thomas Kollar and Sergey Levine and Chelsea Finn},
booktitle={RSS 2024 Workshop: Data Generation for Robotics},
year={2024},
url={https://openreview.net/forum?id=Ml2pTYLNLi}
}

@misc{zhang2025stepworldmodelssurvey,
      title={A Step Toward World Models: A Survey on Robotic Manipulation}, 
      author={Peng-Fei Zhang and Ying Cheng and Xiaofan Sun and Shijie Wang and Fengling Li and Lei Zhu and Heng Tao Shen},
      year={2025},
      eprint={2511.02097},
      archivePrefix={arXiv},
      primaryClass={cs.RO},
      url={https://arxiv.org/abs/2511.02097}, 
}

@article{fang2023rh20t,
  title={Rh20t: A comprehensive robotic dataset for learning diverse skills in one-shot},
  author={Fang, Hao-Shu and Fang, Hongjie and Tang, Zhenyu and Liu, Jirong and Wang, Chenxi and Wang, Junbo and Zhu, Haoyi and Lu, Cewu},
  journal={arXiv preprint arXiv:2307.00595},
  year={2023}
}

@techreport{kragic2002survey_visual_servoing_for_manipulation,
  title       = {Survey on Visual Servoing for Manipulation},
  author      = {Kragic, Danica and Christensen, Henrik I.},
  institution = {KTH Royal Institute of Technology},
  type        = {Technical Report},
  number      = {ISRN KTH/NA/P-02/01-SE},
  year        = {2002},
  note        = {CVAP259}
}

@article{wang2025robot_embodied_visual_perception_survey,
  title   = {Robot Manipulation Based on Embodied Visual Perception: A Survey},
  author  = {Wang, Sicheng and Nikoli{\'c}, Milutin N. and Lam, Tin Lun and Gao, Qing and Ding, Runwei and Zhang, Tianwei},
  journal = {CAAI Transactions on Intelligence Technology},
  year    = {2025},
  doi     = {10.1049/cit2.70022},
  publisher = {Wiley}
}

@article{shao2025large_VLA_survey,
  title={Large vlm-based vision-language-action models for robotic manipulation: A survey},
  author={Shao, Rui and Li, Wei and Zhang, Lingsen and Zhang, Renshan and Liu, Zhiyang and Chen, Ran and Nie, Liqiang},
  journal={arXiv preprint arXiv:2508.13073},
  year={2025}
}

@article{din2025vision_VLA_survey,
  title={Vision language action models in robotic manipulation: A systematic review},
  author={Din, Muhayy Ud and Akram, Waseem and Saoud, Lyes Saad and Rosell, Jan and Hussain, Irfan},
  journal={arXiv preprint arXiv:2507.10672},
  year={2025}
}

@article{li2024foundation_manipulation_survey,
author = {Dingzhe Li and Yixiang Jin and YuHao Sun and Yong A and Hongze Yu and Jun Shi and Xiaoshuai Hao and Peng Hao and Huaping Liu and Xiang Li and Xinde Li and Fuchun Sun and Jianwei Zhang and Bin Fang},
title ={What foundation models can bring for robot learning in manipulation: A survey},
journal = {The International Journal of Robotics Research},
volume = {0},
number = {0},
pages = {02783649251390579},
year = {2025},
doi = {10.1177/02783649251390579},
URL = { https://doi.org/10.1177/02783649251390579 },
eprint = { https://doi.org/10.1177/02783649251390579 }
}

@article{chen2011active_perception_survey,
  title={Active vision in robotic systems: A survey of recent developments},
  author={Chen, Shengyong and Li, Youfu and Kwok, Ngai Ming},
  journal={The International Journal of Robotics Research},
  volume={30},
  number={11},
  pages={1343--1377},
  year={2011},
  publisher={Sage Publications Sage UK: London, England}
}

@article{cong2021_3D_perception_manipulation_survey,
  title={A comprehensive study of 3-D vision-based robot manipulation},
  author={Cong, Yang and Chen, Ronghan and Ma, Bingtao and Liu, Hongsen and Hou, Dongdong and Yang, Chenguang},
  journal={IEEE Transactions on Cybernetics},
  volume={53},
  number={3},
  pages={1682--1698},
  year={2021},
  publisher={IEEE}
}

@article{shahria2022_vision_manipulation_survey,
  title={A comprehensive review of vision-based robotic applications: Current state, components, approaches, barriers, and potential solutions},
  author={Shahria, Md Tanzil and Sunny, Md Samiul Haque and Zarif, Md Ishrak Islam and Ghommam, Jawhar and Ahamed, Sheikh Iqbal and Rahman, Mohammad H},
  journal={Robotics},
  volume={11},
  number={6},
  pages={139},
  year={2022},
  publisher={MDPI}
}

@article{mccarthy2025_video_manipulation_survye,
  title   = {Towards Generalist Robot Learning from Internet Video: A Survey},
  author  = {McCarthy, Robert and Tan, Daniel C. H. and Schmidt, Dominik and Acero, Fernando and Herr, Nathan and Du, Yilun and Thuruthel, Thomas G. and Li, Zhibin},
  journal = {Journal of Artificial Intelligence Research},
  volume  = {83},
  year    = {2025}
}

@article{motoda2025_VLA_survey,
author = {Tomohiro Motoda  and Koshi Makihara  and Ryoichi Nakajo  and Hanbit Oh  and Keisuke Shirai  and Ryo Hanai  and Masaki Murooka  and Yuma Suzuki  and Hiroki Nishihara  and Mitsuru Takeda  and Takumi Takada  and Takayuki Hori  and Yukiyasu Domae },
title = {Recipe for Vision-Language-Action Models in Robotic Manipulation: A Survey},
journal = {TechRxiv},
volume = {2025},
number = {0826},
pages = {},
year = {2025},
doi = {10.36227/techrxiv.175624610.06665789/v1},
URL = {https://www.techrxiv.org/doi/abs/10.36227/techrxiv.175624610.06665789/v1},
eprint = {https://www.techrxiv.org/doi/pdf/10.36227/techrxiv.175624610.06665789/v1}
}

@article{yamanobe2017_affordance_manipulation_survey,
  title     = {A Brief Review of Affordance in Robotic Manipulation Research},
  author    = {Yamanobe, Natsuki and Wan, Weiwei and Ramirez-Alpizar, Ixchel G. and Petit, Damien and Tsuji, Tokuo and Akizuki, Shuichi and Hashimoto, Manabu and Nagata, Kazuyuki and Harada, Kensuke},
  journal   = {Advanced Robotics},
  volume    = {31},
  number    = {19-20},
  pages     = {1086--1101},
  year      = {2017},
  publisher = {Taylor \& Francis}
}

@article{deng2025_RL_VLA_manipulation_survey,
author = {Haoyuan Deng  and Zhenyu Wu  and Haichao Liu  and Wenkai Guo  and Yuquan Xue  and Ziyu Shan  and Chuanrui Zhang  and Bofang Jia  and Yuan Ling  and Guanxing Lu  and Ziwei Wang },
title = {A Survey on Reinforcement Learning of Vision-Language-Action Models for Robotic Manipulation},
journal = {TechRxiv},
volume = {2025},
number = {1209},
pages = {},
year = {2025},
doi = {10.36227/techrxiv.176531955.54563920/v1},
URL = {https://www.techrxiv.org/doi/abs/10.36227/techrxiv.176531955.54563920/v1},
eprint = {https://www.techrxiv.org/doi/pdf/10.36227/techrxiv.176531955.54563920/v1}
}

@InProceedings{H2O_Kwon_2021_ICCV,
author = {Kwon, Taein and Tekin, Bugra and St\"uhmer, Jan and Bogo, Federica and Pollefeys, Marc},
title = {H2O: Two Hands Manipulating Objects for First Person Interaction Recognition},
booktitle = {Proceedings of the IEEE/CVF International Conference on Computer Vision (ICCV)},
month = {October},
year = {2021},
pages = {10138-10148}
}

@misc{DexYCB_chao2021_CVPR,
      title={DexYCB: A Benchmark for Capturing Hand Grasping of Objects}, 
      author={Yu-Wei Chao and Wei Yang and Yu Xiang and Pavlo Molchanov and Ankur Handa and Jonathan Tremblay and Yashraj S. Narang and Karl Van Wyk and Umar Iqbal and Stan Birchfield and Jan Kautz and Dieter Fox},
      year={2021},
      eprint={2104.04631},
      archivePrefix={arXiv},
      primaryClass={cs.CV},
      url={https://arxiv.org/abs/2104.04631}, 
}

@misc{OpenSora_zheng2024,
      title={Open-Sora: Democratizing Efficient Video Production for All}, 
      author={Zangwei Zheng and Xiangyu Peng and Tianji Yang and Chenhui Shen and Shenggui Li and Hongxin Liu and Yukun Zhou and Tianyi Li and Yang You},
      year={2024},
      eprint={2412.20404},
      archivePrefix={arXiv},
      primaryClass={cs.CV},
      url={https://arxiv.org/abs/2412.20404}, 
}

@misc{RSSM_hafner2019,
      title={Learning Latent Dynamics for Planning from Pixels}, 
      author={Danijar Hafner and Timothy Lillicrap and Ian Fischer and Ruben Villegas and David Ha and Honglak Lee and James Davidson},
      year={2019},
      eprint={1811.04551},
      archivePrefix={arXiv},
      primaryClass={cs.LG},
      url={https://arxiv.org/abs/1811.04551}, 
}

@inproceedings{dream2flow_dharmarajan2025,
  title     = {Dream2Flow: Bridging Video Generation and Open-World Manipulation with 3D Object Flow},
  author    = {Dharmarajan, Karthik and Huang, Wenlong and Wu, Jiajun and Fei-Fei, Li and Zhang, Ruohan},
  booktitle = {ICRA 2026 Workshop on Spatial Reasoning and Robot Autonomy},
  year      = {2026},
  url       = {https://openreview.net/forum?id=NMQ9Qw5j3i}
}

@misc{pointworld_huang2026,
      title={PointWorld: Scaling 3D World Models for In-The-Wild Robotic Manipulation}, 
      author={Wenlong Huang and Yu-Wei Chao and Arsalan Mousavian and Ming-Yu Liu and Dieter Fox and Kaichun Mo and Li Fei-Fei},
      year={2026},
      eprint={2601.03782},
      archivePrefix={arXiv},
      primaryClass={cs.RO},
      url={https://arxiv.org/abs/2601.03782}, 
}

@inproceedings{RIGVid_patel2025,
  title     = {Robotic Manipulation by Imitating Generated Videos Without Physical Demonstrations},
  author    = {Patel, Shivansh and Mohan, Shraddhaa and Mai, Hanlin and Jain, Unnat and Lazebnik, Svetlana and Li, Yunzhu},
  booktitle = {International Conference on Learning Representations (ICLR)},
  year      = {2026},
  url       = {https://openreview.net/forum?id=tv0Sz8A9Tc}
}

@inproceedings{Gen2Act_bharadhwaj2024,
  title     = {Gen2Act: Human Video Generation in Novel Scenarios enables Generalizable Robot Manipulation},
  author    = {Bharadhwaj, Homanga and Dwibedi, Debidatta and Gupta, Abhinav and Tulsiani, Shubham and Doersch, Carl and Xiao, Ted and Shah, Dhruv and Xia, Fei and Sadigh, Dorsa and Kirmani, Sean},
  booktitle = {Proceedings of the 9th Conference on Robot Learning},
  series    = {Proceedings of Machine Learning Research},
  volume    = {305},
  pages     = {3936--3951},
  year      = {2025},
  publisher = {PMLR},
  url       = {https://proceedings.mlr.press/v305/bharadhwaj25a.html}
}

@misc{RSRD_kerr2024,
      title={Robot See Robot Do: Imitating Articulated Object Manipulation with Monocular 4D Reconstruction}, 
      author={Justin Kerr and Chung Min Kim and Mingxuan Wu and Brent Yi and Qianqian Wang and Ken Goldberg and Angjoo Kanazawa},
      year={2024},
      eprint={2409.18121},
      archivePrefix={arXiv},
      primaryClass={cs.RO},
      url={https://arxiv.org/abs/2409.18121}, 
}

@article{video_depth_anything_chen2025,
  title={Video Depth Anything: Consistent Depth Estimation for Super-Long Videos},
  author={Chen, Sili and Guo, Hengkai and Zhu, Shengnan and Zhang, Feihu and Huang, Zilong and Feng, Jiashi and Kang, Bingyi},
  journal={arXiv:2501.12375},
  year={2025}
}

@article{household_robots_fiorini2000cleaning,
  title={Cleaning and household robots: A technology survey},
  author={Fiorini, Paolo and Prassler, Erwin},
  journal={Autonomous robots},
  volume={9},
  number={3},
  pages={227--235},
  year={2000},
  publisher={Springer}
}

@article{household_robots_soni2024advancing,
  title={Advancing household robotics: Deep interactive reinforcement learning for efficient training and enhanced performance},
  author={Soni, Arpita and Alla, Sujatha and Dodda, Suresh and Volikatla, Hemanth},
  journal={arXiv preprint arXiv:2405.18687},
  year={2024}
}

@article{human_robot_collection_review_baratta2023human,
  title={Human robot collaboration in industry 4.0: a literature review},
  author={Baratta, Alessio and Cimino, Antonio and Gnoni, Maria Grazia and Longo, Francesco},
  journal={Procedia Computer Science},
  volume={217},
  pages={1887--1895},
  year={2023},
  publisher={Elsevier}
}

@inproceedings{Robotics_logistics_echelmeyer2008robotics,
  title={Robotics-logistics: Challenges for automation of logistic processes},
  author={Echelmeyer, Wolfgang and Kirchheim, Alice and Wellbrock, Eckhard},
  booktitle={2008 IEEE International Conference on Automation and Logistics},
  pages={2099--2103},
  year={2008},
  organization={IEEE}
}

@article{robot_industrial_automation_faheem2024ai,
  title={AI and robotic: About the transformation of construction industry automation as well as labor productivity},
  author={Faheem, Muhammad Ashraf and Zafar, Nabeel and Kumar, Parkash and Melon, MMH and Prince, NU and Al Mamun, Mohd Abdullah},
  journal={Remittances Review},
  volume={9},
  pages={871--888},
  year={2024},
}

@article{
Feng_human_video_survey_TechRxiv_2026,
author = {Zhiyuan Feng  and Qixiu Li  and Huizhi Liang  and Rushuai Yang  and Yichao Shen  and Zhiying Du  and Zhaowei Zhang  and Yu Deng  and Li Zhao  and Hao Zhao  and Zongqing Lu  and Oier Mees  and Marc Pollefeys  and Jiaolong Yang  and Baining Guo },
title = {From Human Videos to Robot Manipulation: A Survey on Scalable Vision-Language-Action Learning with Human-Centric Data},
journal = {TechRxiv},
volume = {2026},
number = {0216},
pages = {},
year = {2026},
doi = {10.36227/techrxiv.177126525.54038135/v1},
URL = {https://www.techrxiv.org/doi/abs/10.36227/techrxiv.177126525.54038135/v1},
eprint = {https://www.techrxiv.org/doi/pdf/10.36227/techrxiv.177126525.54038135/v1},
        }

@article{Tsuji2026_immitation_learning_survey,
  author = {Toshiaki Tsuji and Yasuhiro Kato and Gokhan Solak and Heng Zhang and Tadej Petri{\v{c}} and Francesco Nori and Arash Ajoudani},
  title = {A survey on imitation learning for contact-rich tasks in robotics},
  journal = {The International Journal of Robotics Research},
  year = {2026},
  doi = {10.1177/02783649261417694},
  url = {https://doi.org/10.1177/02783649261417694}
}

@inproceedings{ye2026dreamzero,
title={World Action Models are Zero-shot Policies},
author={Seonghyeon Ye and Yunhao Ge and Kaiyuan Zheng and Shenyuan Gao and Sihyun Yu and George Kurian and Suneel Indupuru and You Liang Tan and Chuning Zhu and Jiannan Xiang and Ayaan Naveed Malik and Kyungmin Lee and William Liang and Nadun Ranawaka Arachchige and Jiasheng Gu and Yinzhen Xu and Guanzhi Wang and Fengyuan Hu and Avnish Narayan and Johan Bjorck and Jing Wang and Gwanghyun Kim and Dantong Niu and Ruijie Zheng and Yuqi Xie and Jimmy Wu and Qi Wang and Danfei Xu and Yilun Du and Ryan Julian and Yevgen Chebotar and Scott Reed and Jan Kautz and Yuke Zhu and Linxi Fan and Joel Jang},
booktitle={ICLR 2026 the 2nd Workshop on World Models: Understanding, Modelling and Scaling},
year={2026},
url={https://openreview.net/forum?id=cd33uUB609}
}

@inproceedings{lyu2026lda,
  title     = {LDA-1B: Scaling Latent Dynamics Action Model via Universal Embodied Data Ingestion},
  author    = {Jiangran Lyu and Kai Liu and Xuheng Zhang and Haoran Liao and Yusen Feng and Wenxuan Zhu and Tingrui Shen and Jiayi Chen and Jiazhao Zhang and Yifei Dong and Wenbo Cui and Senmao Qi and Shuo Wang and Yixin Zheng and Mi Yan and Xuesong Shi and Haoran Li and Dongbin Zhao and Ming-Yu Liu and Zhizheng Zhang and Li Yi and Yizhou Wang and He Wang},
  booktitle = {Proceedings of Robotics: Science and Systems},
  year      = {2026},
  note      = {To appear},
  url       = {https://arxiv.org/abs/2602.12215}
}

@inproceedings{kim2026cosmos,
  title     = {Cosmos Policy: Fine-Tuning Video Models for Visuomotor Control and Planning},
  author    = {Kim, Moo Jin and Gao, Yihuai and Lin, Tsung-Yi and Lin, Yen-Chen and Ge, Yunhao and Lam, Grace and Liang, Percy and Song, Shuran and Liu, Ming-Yu and Finn, Chelsea and Gu, Jinwei},
  booktitle = {International Conference on Learning Representations (ICLR)},
  year      = {2026},
  url       = {https://openreview.net/forum?id=wPEIStHxYH}
}

@article{yuan2026fast_WAM,
  title={Fast-WAM: Do World Action Models Need Test-time Future Imagination?},
  author={Yuan, Tianyuan and Dong, Zibin and Liu, Yicheng and Zhao, Hang},
  journal={arXiv preprint arXiv:2603.16666},
  year={2026}
}

@article{guo2026X_WAM,
  title={Unified 4D World Action Modeling from Video Priors with Asynchronous Denoising},
  author={Guo, Jun and Li, Qiwei and Li, Peiyan and Chen, Zilong and Sun, Nan and Su, Yifei and Wang, Heyun and Zhang, Yuan and Li, Xinghang and Liu, Huaping},
  journal={arXiv preprint arXiv:2604.26694},
  year={2026}
}

@misc{CLAM_liang2025,
      title={CLAM: Continuous Latent Action Models for Robot Learning from Unlabeled Demonstrations}, 
      author={Anthony Liang and Pavel Czempin and Matthew Hong and Yutai Zhou and Erdem Biyik and Stephen Tu},
      year={2025},
      eprint={2505.04999},
      archivePrefix={arXiv},
      primaryClass={cs.RO},
      url={https://arxiv.org/abs/2505.04999}, 
}

@inproceedings{Moto_chen2025,
  title     = {Moto: Latent Motion Token as the Bridging Language for Learning Robot Manipulation from Videos},
  author    = {Chen, Yi and Ge, Yuying and Tang, Weiliang and Li, Yizhuo and Ge, Yixiao and Ding, Mingyu and Shan, Ying and Liu, Xihui},
  booktitle = {Proceedings of the IEEE/CVF International Conference on Computer Vision (ICCV)},
  pages     = {19752--19763},
  year      = {2025},
  url       = {https://openaccess.thecvf.com/content/ICCV2025/html/Chen_Moto_Latent_Motion_Token_as_the_Bridging_Language_for_Learning_ICCV_2025_paper.html}
}

@misc{ConLA_dai2026,
      title={ConLA: Contrastive Latent Action Learning from Human Videos for Robotic Manipulation}, 
      author={Weisheng Dai and Kai Lan and Jianyi Zhou and Bo Zhao and Xiu Su and Junwen Tong and Weili Guan and Shuo Yang},
      year={2026},
      eprint={2602.00557},
      archivePrefix={arXiv},
      primaryClass={cs.RO},
      url={https://arxiv.org/abs/2602.00557}, 
}

@inproceedings{
villaX_chen2026,
title={villa-X: Enhancing Latent Action Modeling in Vision-Language-Action Models},
author={Xiaoyu Chen and Hangxing Wei and Pushi Zhang and Chuheng Zhang and Kaixin Wang and Yanjiang Guo and Rushuai Yang and Yucen Wang and Xinquan Xiao and Li Zhao and Jianyu Chen and Jiang Bian},
booktitle={The Fourteenth International Conference on Learning Representations},
year={2026},
url={https://openreview.net/forum?id=y5CaJb17Fn}
}

@INPROCEEDINGS{GO1_Bu_IROS_2025,
  author={AgiBot-World-Contributors and Qingwen Bu and Jisong Cai and Li Chen and Xiuqi Cui and Yan Ding and Siyuan Feng and Shenyuan Gao and Xindong He and Xuan Hu and Xu Huang and Shu Jiang and Yuxin Jiang and Cheng Jing and Hongyang Li and Jialu Li and Chiming Liu and Yi Liu and Yuxiang Lu and Jianlan Luo and Ping Luo and Yao Mu and Yuehan Niu and Yixuan Pan and Jiangmiao Pang and Yu Qiao and Guanghui Ren and Cheng Ruan and Jiaqi Shan and Yongjian Shen and Chengshi Shi and Mingkang Shi and Modi Shi and Chonghao Sima and Jianheng Song and Huijie Wang and Wenhao Wang and Dafeng Wei and Chengen Xie and Guo Xu and Junchi Yan and Cunbiao Yang and Lei Yang and Shukai Yang and Maoqing Yao and Jia Zeng and Chi Zhang and Qinglin Zhang and Bin Zhao and Chengyue Zhao and Jiaqi Zhao and Jianchao Zhu},
  booktitle={2025 IEEE/RSJ International Conference on Intelligent Robots and Systems (IROS)}, 
  title={AgiBot World Colosseo: A Large-Scale Manipulation Platform for Scalable and Intelligent Embodied Systems}, 
  year={2025},
  volume={},
  number={},
  pages={3549-3556},
  doi={10.1109/IROS60139.2025.11247088}}

@inproceedings{
DynaMo_cui2024,
title={DynaMo: In-Domain Dynamics Pretraining for Visuo-Motor Control},
author={Zichen Jeff Cui and Hengkai Pan and Aadhithya Iyer and Siddhant Haldar and Lerrel Pinto},
booktitle={The Thirty-eighth Annual Conference on Neural Information Processing Systems},
year={2024},
url={https://openreview.net/forum?id=vUrOuc6NR3}
}

@inproceedings{
LAOM_2025latent,
title={Latent Action Learning Requires Supervision in the Presence of Distractors},
author={Alexander Nikulin and Ilya Zisman and Denis Tarasov and Lyubaykin Nikita and Andrei Polubarov and Igor Kiselev and Vladislav Kurenkov},
booktitle={Forty-second International Conference on Machine Learning},
year={2025},
url={https://openreview.net/forum?id=2gcEQCT7QW}
}

@inproceedings{HiLAM_kim2026,
  title     = {Hierarchical Latent Action Model},
  author    = {Kim, Hanjung and Pinto, Lerrel and Kim, Seon Joo},
  booktitle = {ICLR 2026 Workshop on World Models: Understanding, Modelling and Scaling},
  year      = {2026},
  url       = {https://openreview.net/forum?id=IUaiouMYXp}
}

@misc{RotVLA_li2026,
      title={RotVLA: Rotational Latent Action for Vision-Language-Action Model}, 
      author={Qiwei Li and Xicheng Gong and Xinghang Li and Peiyan Li and Quanyun Zhou and Hangjun Ye and Jiahuan Zhou and Yadong Mu},
      year={2026},
      eprint={2605.13403},
      archivePrefix={arXiv},
      primaryClass={cs.RO},
      url={https://arxiv.org/abs/2605.13403}, 
}

@misc{SoftVQVAE_chen2025,
      title={SoftVQ-VAE: Efficient 1-Dimensional Continuous Tokenizer}, 
      author={Hao Chen and Ze Wang and Xiang Li and Ximeng Sun and Fangyi Chen and Jiang Liu and Jindong Wang and Bhiksha Raj and Zicheng Liu and Emad Barsoum},
      year={2025},
      eprint={2412.10958},
      archivePrefix={arXiv},
      primaryClass={cs.CV},
      url={https://arxiv.org/abs/2412.10958}, 
}

@misc{ALAM_tang2026,
      title={ALAM: Algebraically Consistent Latent Action Model for Vision-Language-Action Models}, 
      author={Zuojin Tang and Haoyun Liu and Xinyuan Chang and Changjie Wu and Dongjie Huo and Yandan Yang and Bin Liu and Zhejia Cai and Feng Xiong and Mu Xu and jiachen Luo and De Ma and Zhiheng Ma and Gang Pan},
      year={2026},
      eprint={2605.10819},
      archivePrefix={arXiv},
      primaryClass={cs.RO},
      url={https://arxiv.org/abs/2605.10819}, 
}

@misc{CoMo_yang2026,
      title={CoMo: Learning Continuous Latent Motion from Internet Videos for Scalable Robot Learning}, 
      author={Jiange Yang and Yansong Shi and Haoyi Zhu and Mingyu Liu and Kaijing Ma and Yating Wang and Gangshan Wu and Tong He and Limin Wang},
      year={2026},
      eprint={2505.17006},
      archivePrefix={arXiv},
      primaryClass={cs.CV},
      url={https://arxiv.org/abs/2505.17006}, 
}

@inproceedings{
SIMPLER_li_CORL_2024,
title={Evaluating Real-World Robot Manipulation Policies in Simulation},
author={Xuanlin Li and Kyle Hsu and Jiayuan Gu and Oier Mees and Karl Pertsch and Homer Rich Walke and Chuyuan Fu and Ishikaa Lunawat and Isabel Sieh and Sean Kirmani and Sergey Levine and Jiajun Wu and Chelsea Finn and Hao Su and Quan Vuong and Ted Xiao},
booktitle={8th Annual Conference on Robot Learning},
year={2024},
url={https://openreview.net/forum?id=LZh48DTg71}
}

\end{document}